\documentclass{article}

 \usepackage[eandd, final]{neurips_2026}
\usepackage[utf8]{inputenc} % allow utf-8 input
\usepackage[T1]{fontenc}    % use 8-bit T1 fonts
\usepackage{hyperref}       % hyperlinks
\usepackage{url}            % simple URL typesetting
\usepackage{booktabs}       % professional-quality tables
\usepackage{amsfonts}       % blackboard math symbols
\usepackage{nicefrac}       % compact symbols for 1/2, etc.
\usepackage{microtype}      % microtypography
\usepackage{xcolor}         % colors

\newcommand{\ie}{\textit{i.e.}}
\newcommand{\eg}{\textit{e.g.}}
\newcommand{\etc}{\textit{etc.}}
\newcommand{\methodname}{\textbf{\textsc{VisFactor}}}
\newcommand{\nummodel}{39}
\newcommand{\bestmodel}{GPT-5.5}
\newcommand{\best}{55.9\%}

\usepackage{colortbl}
\usepackage{nicefrac}
\usepackage{arydshln}
\usepackage{graphicx}
\usepackage{subfig}
\usepackage{enumitem}
\usepackage{multirow}
\usepackage[most]{tcolorbox}
\usepackage{siunitx}
\usepackage{pifont}

\definecolor{myred}{RGB}{251, 180, 174}
\definecolor{myorange}{RGB}{254, 217, 166}
\definecolor{mybrown}{RGB}{229, 216, 189}

\definecolor{myblue}{RGB}{179, 205, 227}
\definecolor{mypurple}{RGB}{222, 203, 228}
\definecolor{mypink}{RGB}{252, 218, 236}

\definecolor{mygreen}{RGB}{203, 235, 197}
\definecolor{myyellow}{RGB}{255, 254, 204}
\definecolor{mygray}{RGB}{232, 232, 232}

\definecolor{crossred}{RGB}{139, 0, 0}
\definecolor{checkgreen}{RGB}{0, 100, 0}

\newcommand{\cmark}{\textcolor{checkgreen}{\ding{51}}}
\newcommand{\xmark}{~}

\tcbset{
  promptbox/.style={
    enhanced,
    float,
    floatplacement=h,
    top=10pt,
    bottom=6pt,
    left=6pt,
    right=6pt,
    boxsep=2pt,
    colback=white,
    coltitle=white,
    colbacktitle=black,
    colframe=black,
    fontupper=\scriptsize,
    attach boxed title to top left={yshift=-0.1in,xshift=0.15in},
    boxed title style={boxrule=0pt,colframe=white,},
  }
}
\newtcolorbox{promptbox}[2][]{promptbox, title=#2,#1}

\renewcommand{\cite}{\citep}

\title{Human Cognitive Benchmarks Reveal Foundational Visual Gaps in MLLMs}

\author{Jen-tse Huang$^{1}$ \quad Dasen Dai$^{1}$ \quad Jen-yuan Huang$^{2}$ \quad Youliang Yuan$^{3}$ \quad Xiaoyuan Liu$^{3}$ \\
\bf Wenxuan Wang$^{4*}$ \quad\ Wenxiang Jiao$^{5}$ \quad\ Pinjia He$^{3}$ \quad\ Zhaopeng Tu$^{5}$ \quad\ Haodong Duan$^{6*}$ \\
$^{1}$CUHK \quad $^{2}$PKU \quad $^{3}$CUHKSZ \quad $^{4}$RUC \quad $^{5}$Tencent \quad $^{6}$SHLab \\
$^{*}$Corresponding authors: \quad \texttt{jwxwang@gmail.com} \quad \texttt{duanhaodong@pjlab.org.cn}
}

\begin{document}

\maketitle

\begin{center}
\vspace{-20pt}

\href{https://cuhk-arise.github.io/VisFactor/}{\includegraphics[width=10pt]{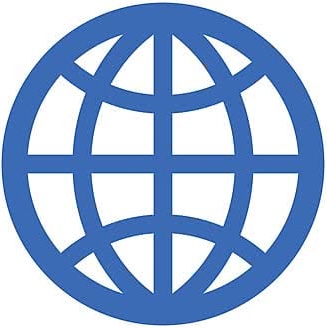}\ \textbf{Homepage Leaderboard}} \quad \quad
\href{https://github.com/CUHK-ARISE/VisFactor}{\includegraphics[width=10pt]{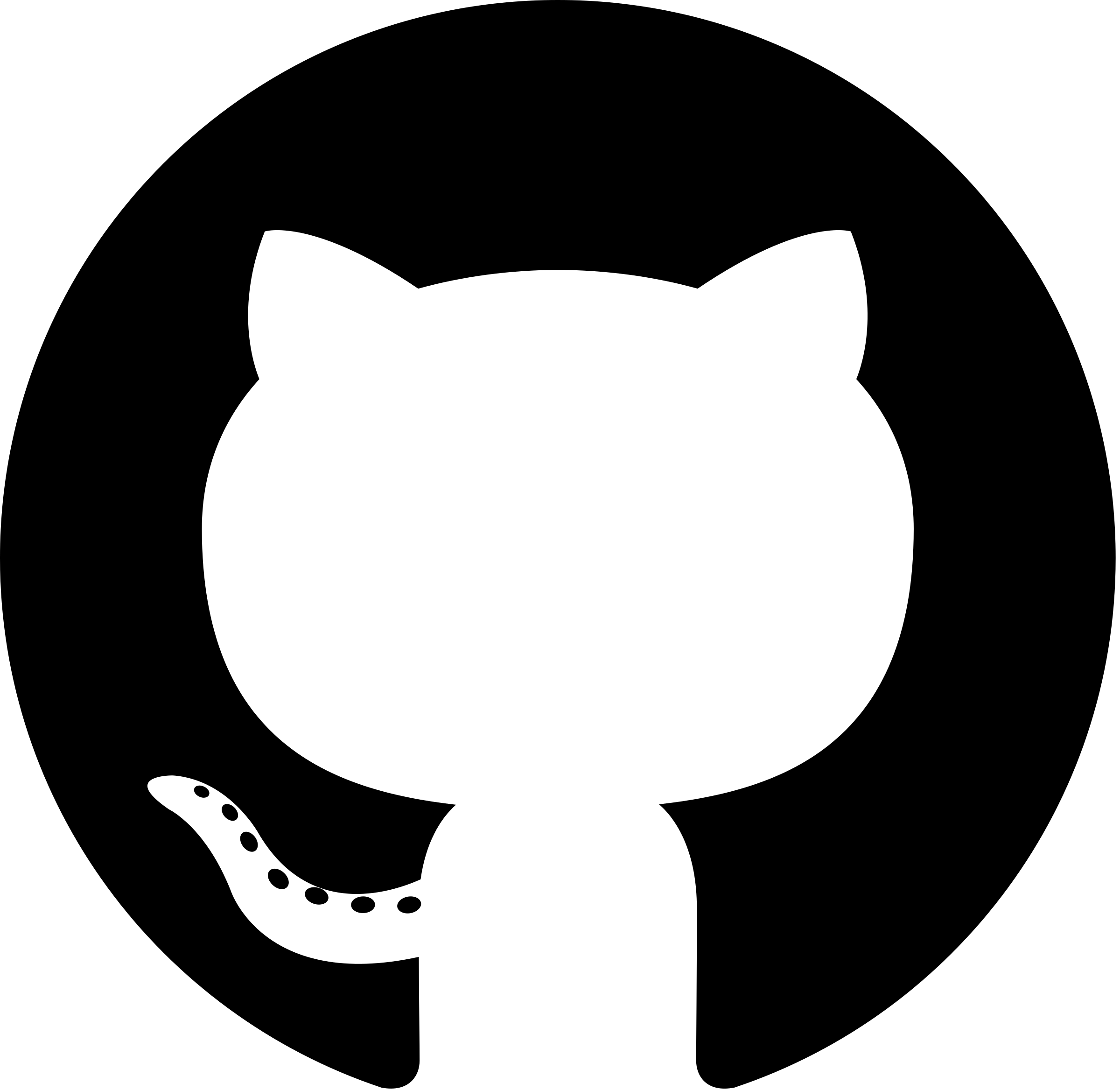}\ \textbf{GitHub Code}} \quad \quad
\href{https://huggingface.co/datasets/penguin-G/VisFactor}{\includegraphics[width=10pt]{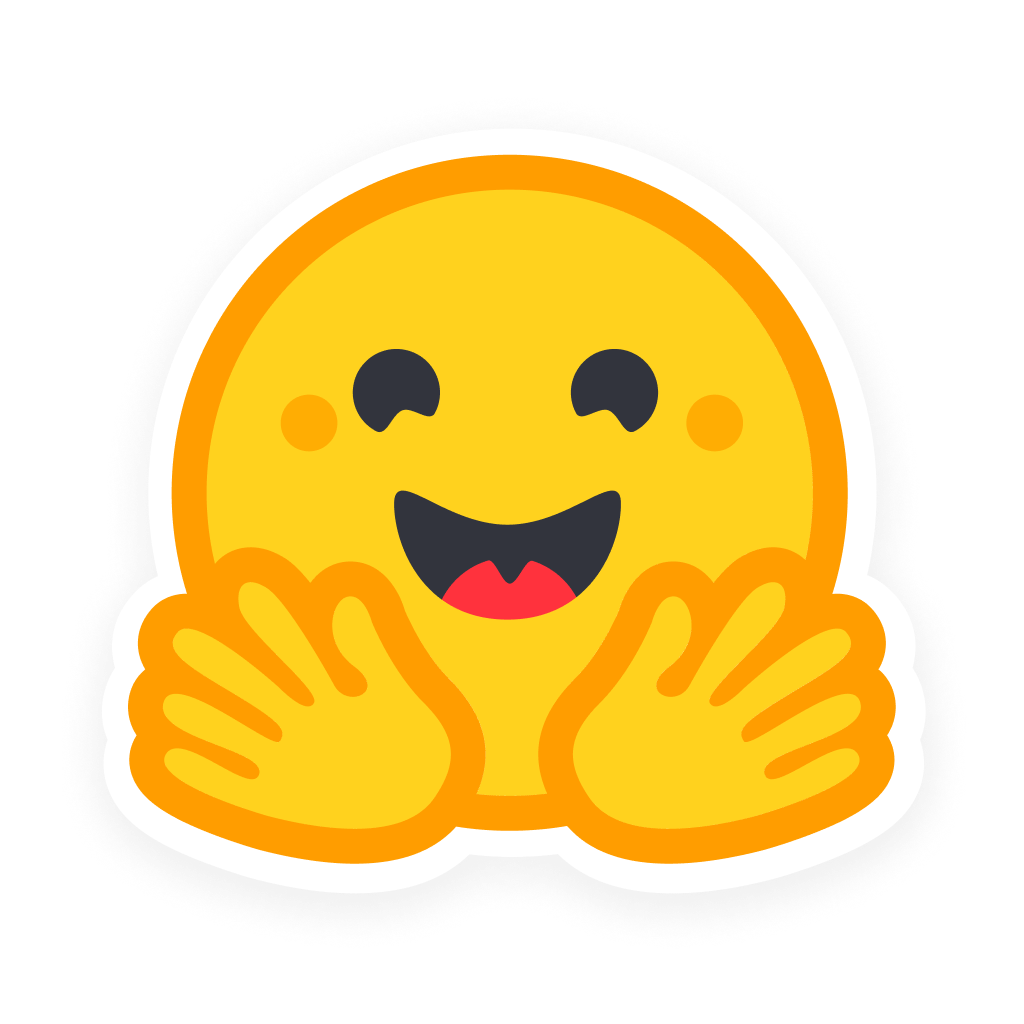}\ \textbf{HuggingFace Dataset}}

\vspace{10pt}
\end{center}

\begin{abstract}
Humans develop perception through a bottom-up hierarchy: from basic primitives and Gestalt principles to high-level semantics.
Current Multimodal Large Language Models (MLLMs) are trained directly on complex downstream tasks.
Does it guarantee their performance on these foundational visual capabilities?
To systematically investigate this gap, we introduce {\methodname}, a benchmark that digitizes 20 vision-centric subtests from FRCT, a well-established cognitive psychology assessment spanning four domains of human visual cognition.
Furthermore, we design algorithms to automatically construct and validate unlimited test cases with controllable difficulty.
Using {\methodname}, we evaluate {\nummodel} frontier MLLMs, including both proprietary (\eg, GPT, Gemini) and open-source (\eg, LLaMA, Qwen) models.
The best model achieves a score of only {\best}.
Analysis reveals good internal consistency (Cronbach's $\alpha = 0.94$) and construct validity (compared to existing vision benchmarks).
Models consistently fail on tasks such as mental rotation, spatial relation inference, and figure--ground discrimination, regardless of model size or prompting strategy.
These findings suggest that performance improvements on existing general benchmarks might represent \textit{castles in the air} instead of a genuine mastery of human-like visual cognition.
\end{abstract}

\begin{figure}[h]
    \centering
    \includegraphics[width=1.0\linewidth]{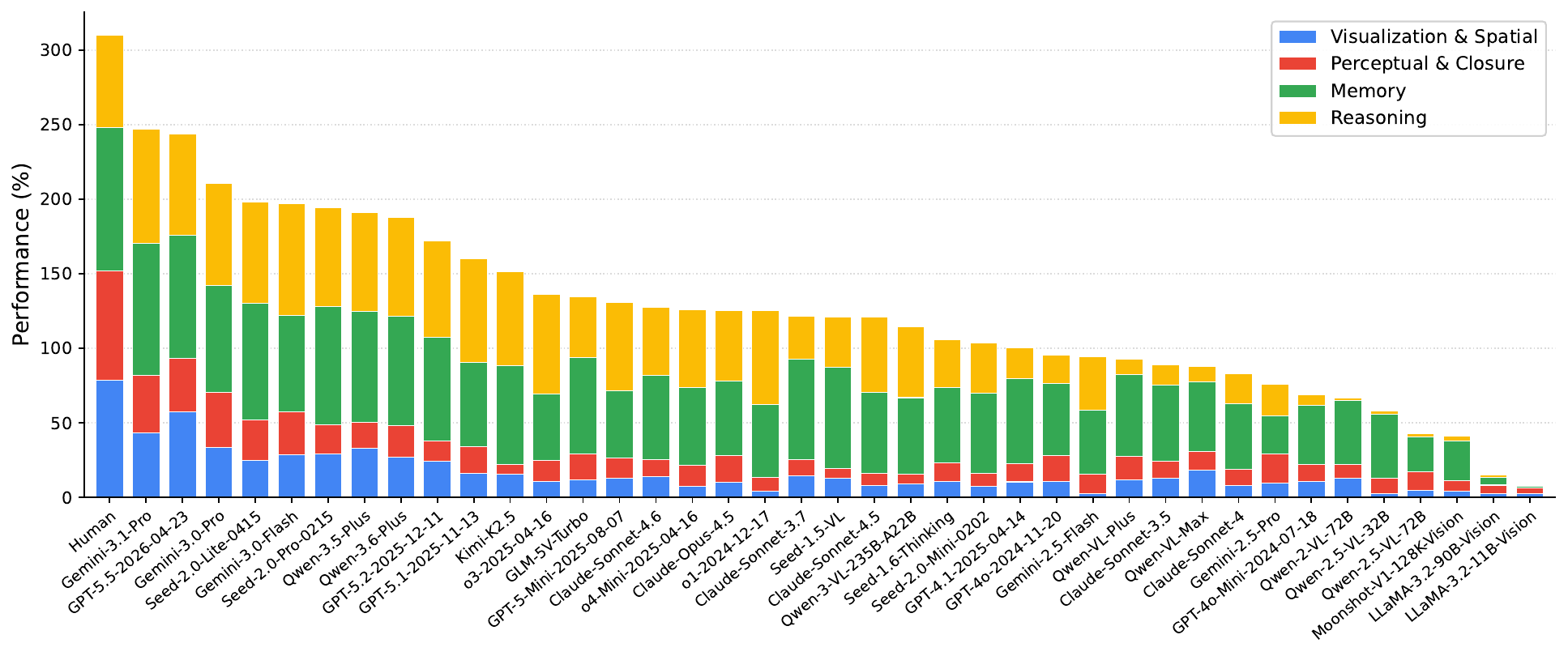}
    \caption{A stacked bar chart ranks all models by overall score, with each bar decomposed into the four cognitive domains. There is still a gap between the best model and our human baseline (average performance of undergraduate students). Figure~\ref{fig:timeline} further shows model scores vs. their release dates.}
    \label{fig:placeholder}
\end{figure}

\addtocontents{toc}{\protect\setcounter{tocdepth}{0}}

\section{Introduction}
\label{sec:introduction}

Multimodal Large Language Models (MLLMs) have rapidly advanced the state of multimodal artificial intelligence, delivering impressive results in text recognition \cite{liu2024ocrbench, chen2025ocean}, mathematical reasoning \cite{yang2024mathglm, peng2024multimath}, and clinical decision support \cite{azad2023foundational, buckley2026multimodal, ye2024gmai}.
On holistic benchmarks such as MMBench \cite{liu2024mmbench}, frontier models like Gemini-2.5-Pro have reached nearly \textit{90\%} accuracy.
These results have fueled optimism that direct large‑scale pretraining on complex tasks may already confer near‑human visual cognition, boosting downstream applications like embodied AI.

However, human vision develops hierarchically: low-level primitives—such as edge, line, and orientation detection—form the basis for Gestalt principles like closure and grouping, which ultimately scaffold higher-order semantic reasoning.
This raises a critical question: do MLLMs truly possess the fundamental cognitive visual abilities beyond the complex tasks they are trained on?
Closer inspection reveals a significant gap; targeted studies show MLLMs still fail on visual reasoning problems that human novices solve effortlessly~\cite{fu2024blink}.
For example, \citet{ramakrishnan2025does} reports near-random accuracy on the mental rotation test and maze completion test.
Why do models that \textit{see} so effectively in benchmarks fail to \textit{perceive}?
This paradox highlights a key limitation in current evaluation paradigms: most benchmarks primarily prioritize downstream task performance while neglecting the foundational visual faculties that underpin human reasoning.

In cognitive psychology, researchers decompose human cognition into latent factors that can be measured independently.
The \textit{Factor-Referenced Cognitive Test} (FRCT) battery~\cite{ekstrom1976manual} operationalizes this by mapping psychometric factors to narrowly defined subtests.
In contrast to omnibus IQ scales, the FRCT delivers a fine-grained cognitive profile, making it ideal for diagnosing the precise visual capacities an MLLM truly possesses.

With this inspiration, we introduce {\methodname}, which for the first time, adapts 20 vision-centric FRCT subtests into an automated, multimodal benchmark specifically designed for MLLMs.
{\methodname} spans four critical cognitive domains: (1) visualization and spatial reasoning, (2) perceptual and closure, (3) memory, and (4) reasoning.
Prior multimodal benchmarks~\cite{ramakrishnan2025does, fu2024blink} often rely on 
multiple-choice (\nicefrac{1}{N} chance with N choices) or Yes-No (\nicefrac{1}{2} chance) question formats.
Such formats introduce significant random guessing opportunities, often failing to reveal true performance disparities between models.
For a more rigorous evaluation, we generate at least four rule-based variants for every question and deliberately diversify the correct-answer distribution.\footnote{For instance, multiple-choice answers are not always ``A'', nor are ``Yes/No'' items disproportionately ``Yes''.}
This design reduces the overall chance-level accuracy to \textit{2.9\%}, ensuring that any success on {\methodname} reflects genuine visual reasoning rather than lucky guesses.

We evaluated {\nummodel} frontier MLLMs spanning major families, including GPT-5.5~\cite{gpt55}, Gemini-3.1~\cite{gemini31pro}, Claude-4.6~\cite{claudes46}, LLaMA~\cite{llama32}, Qwen3.6~\cite{qwen36}, and SEED-2.0~\cite{seed20}.
Despite advanced prompting strategies such as Chain-of-Thought (CoT)~\cite{wei2022chain}, the best model scores only \textit{{\best}}.
Failures consistently manifested in core visual tasks such as mental rotation, spatial relation inference, and figure–ground discrimination, irrespective of model size or architecture.

The original FRCT has a finite item set, posing a risk of direct overfitting by future models.
To future-proof {\methodname}, we implement a parametric generator for twelve subtests.
This system automatically produces and validates an infinite supply of difficulty-controlled instances that faithfully adhere to the FRCT framework.
By precisely modulating key parameters (\eg, rotation angle, occlusion level, number of steps, and grid size), we can create test suites with controlled difficulties, enabling robust performance tracking without benchmark saturation.
Furthermore, specific tests (\eg, Paper Folding) can generate intermediate visual solution steps, facilitating supervised fine-tuning for future architectural improvements.
Our contributions are as follows:
\begin{enumerate}[nosep]
    \item We introduce the first benchmark to ground MLLM vision assessment to FRCT, providing a human cognitive aspect for multimodal evaluation.
    \item Implementing with VLMEvalKit~\cite{duan2024vlmevalkit}, we digitize FRCT vision items with rigorous evaluation protocol and develop parametric methods to generate and validate controllable-difficulty, infinite-item test sets.
    \item We benchmark {\nummodel} MLLMs and compare the results to 13 existing vision benchmarks, providing a comprehensive analysis of current capabilities to guide future research.
\end{enumerate}
\section{{\methodname}: introduction and implementation}

\subsection{Introduction to twenty subtests}

\paragraph{(1) Visualization \& spatial reasoning}
This category evaluates a subject's capacity to mentally represent, manipulate, and navigate spatial structures.
This includes Spatial Relations \textbf{(S)}, which tests the fundamental ability to perceive and mentally rotate intact objects—such as 2D cards or 3D cubes—while maintaining their global orientation; Spatial Scanning \textbf{(SS)}, which measures the efficiency of exploring complex, cluttered visual fields to identify viable paths or routes, akin to navigating a maze or planning a trajectory on a map; and Visualization \textbf{(VZ)}, which assesses a more advanced form of spatial manipulation that requires the mental restructuring, folding, or transformation of distinct components into entirely new arrangements, as demonstrated in form board, paper folding, and 3D surface development tasks.
These subtests together measure models' ability to track object motions and relative positions in 2D and 3D space.

\paragraph{(2) Perceptual \& closure}
This category evaluates a subject's fundamental ability to process, organize, and interpret visual stimuli under distracting or ambiguous conditions.
It encompasses Closure Flexibility \textbf{(CF)}, which tests the capacity to hold a visual percept in mind in order to disembed a target figure from a complex, distracting background; Closure Speed \textbf{(CS)}, which measures the ability to rapidly unify incomplete, obscured, or disparate visual fields into a recognizable, single concept (\eg, Gestalt completion); and Perceptual Speed \textbf{(P)}, which assesses the fluency and accuracy of scanning or comparing simple visual symbols to quickly identify identical figures.
Together, these subtests isolate models' essential low-level visual processing and pattern recognition capabilities.

\begin{figure*}[t]
    \centering
    \includegraphics[width=1.0\linewidth]{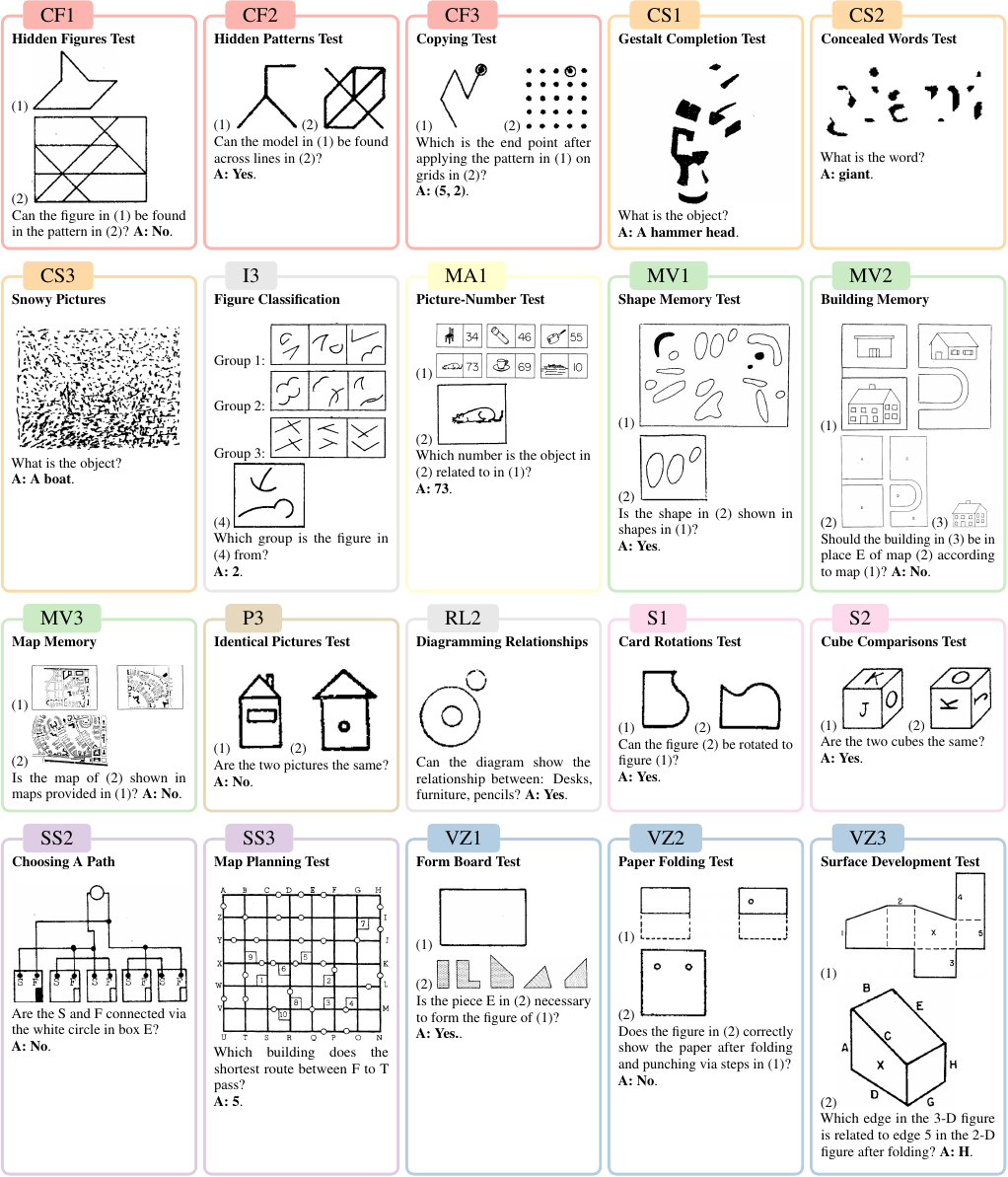}
    \caption{{\methodname} comprises 20 vision-centric cognitive subtests. Each task is designed to isolate core factors of human visual cognition, covering 10 distinct factors in total. Example stimuli, questions, and ground-truth answers are shown for each task. The subtests are converted into either yes/no questions or fill-in-the-blank questions according to \S\ref{sec:variants}. Prompts are shown in \S\ref{sec:description}.}
    \label{fig:all-tests}
    \vspace{-10pt}
\end{figure*}

\paragraph{(3) Memory}
This category evaluates a subject's capacity to encode, retain, and retrieve visual and relational information over short to intermediate durations.
This comprises two factors: Visual Memory \textbf{(MV)}, which assesses the ability to remember the exact configuration, location, and orientation of spatial figures—probing iconic memory and short-term visual retention through tasks that require recalling specific shapes or complex map layouts; and Associative Memory \textbf{(MA)}, which measures the ability to recall specific pairings of previously learned, unrelated stimuli (\eg, picture-number combinations), evaluating intermediate-term associative retrieval and the utilization of mnemonic mediators.
Although in the context of MLLMs, they test the ability to attend to correct content instead of their ``memory,'' we decided to keep the original names for taxonomic fidelity.

\paragraph{(4) Reasoning}
This category evaluates a subject's ability for high-level logical synthesis, abstraction, and deduction.
It consists of: Induction \textbf{(I)}, which assesses the ability to form and test hypotheses to discover underlying rules within visual data, such as identifying shared geometric patterns to classify complex figures; and Logical Reasoning \textbf{(RL)}, which measures deductive abilities, specifically the capacity to evaluate the correctness of conclusions derived from given premises, such as diagramming structured set relationships.
Together, they probe beyond mere visual recognition, requiring the model to abstract rules, retrieve logical algorithms, and apply structured reasoning.

\subsection{Implementation}

\paragraph{Test selection}
The original FRCT battery comprises 72 subtests.
We exclude those that cannot be assessed with a vision–language interface whose output is text only:
(1) \textbf{Image-production tasks} (4 subtests):
Figural Fluency~(FF1–FF3) and Spatial Scanning~(SS1) ask participants to draw or trace; this is incompatible with text-only output.
(2) \textbf{Speech-dependent tasks} (3 subtests):
Memory Span~(MS1–MS3) require subjects to write down what they hear and therefore probe speech-to-text rather than visual cognition.
In the remaining 65 subtests, 45 of them can be completed with pure text input.
GPT-5.5 already scores 82.9 on these text subtests, so we mainly focus on those demanding visual reasoning but accept text answers, forming our benchmark, {\methodname}.
The 20 subtests cover 10 FRCT factors: CF, CS, I, MA, MV, P, RL, S, SS, and VZ (as introduced above).
Figure~\ref{fig:all-tests} shows example questions and answers of each subtest.
Dataset statistics are included in Figure~\ref{fig:overview} and Table~\ref{tab:statistics} in \S\ref{sec:dataset}.
Detailed introduction to each subtest is included in \S\ref{sec:description}.

\paragraph{Digitization}
\textbf{(1) Instructions:}
Directly feeding the human-oriented FRCT instructions to MLLMs proves verbose and occasionally ambiguous.
We therefore ask GPT-4o and Gemini-2.5-Flash to summarize each instruction set to its minimal, AI-friendly form.
A human annotator reconciles the two summaries with the originals, producing a final prompt for every subtest.
\textbf{(2) Questions and Answers:}
All images are captured at \SI{300}{dpi} and cropped to the region containing only the task stimuli (no additional texts).
Ground-truth answers are extracted verbatim from the FRCT manuals.
\textbf{(3) Evaluation Protocol:}
\S\ref{sec:variants} introduces how we reduce chance-level baseline to 2.89\% and how to process MLLM outputs and decide correctness.

\paragraph{Parametric generation}
We implement automatic generation for a subset---CF1--CF3, CS1--CS3, MA1, S1--S2, SS3, VZ1--VZ2.
Figure~\ref{fig:generated-sample} illustrates sample questions generated by our algorithms.
To guarantee correctness, we carefully design algorithms that produce valid question–answer pairs.
For example, in the S2 Cube Comparison Test, we design an algorithm to determine whether two cubes---represented by three characters denoting the upper, front, and right faces, along with their rotation angles---are identical.
In the VZ2 Paper Folding Test, our method randomly selects symmetry axes, folds the figure, punches a hole, and then unfolds it in reverse order to obtain the final answer.
The details of our algorithms are included in \S\ref{sec:implementation}.
\section{Experiments}

\subsection{Settings}
\label{sec:exp-settings}

\paragraph{Models}
We evaluate {\nummodel} models: GPT models \cite{gpt4o-gpt4omini, gpt41, gpt5-gpt5mini-gpt5nano, gpt51, gpt52, gpt55}, Gemini models \cite{gemini25, gemini30pro, gemini30flash, gemini31pro}, Claude models \cite{claudes35, claudes37, claudeo4-claudes4, claudes45, claudes46, claudeo45}, Qwen models \cite{qwen2vl, qwen25vl, qwen3vl, qwenvlmax-qwenvlplus, qwen35, qwen36}, Seed models \cite{seed15vl, seed16, seed20}, Kimi models \cite{moonshotv1vision, kimik25}, LLaMA models \cite{llama32}, o-series \cite{o1, o3-o4mini}, and GLM \cite{glm5v}.

\paragraph{Hyper-parameters}
We set the temperature to 0 or minimal for all models.
The thinking budget is configured as \emph{high} for reasoning models (\eg, Claude-4.5, Gemini-3.0, and GPT-5.5).
Greedy decoding is used as the default sampling strategy.
All models are accessed via their official APIs or Model-as-a-Service platform.\footnote{\url{https://deepinfra.com/}}
In our implementation, the retry count is set to 3, allowing each case up to three retries before being marked as a failure.
All test cases are conducted in a zero-shot setting.

\paragraph{Human results}
To set a baseline to interpret model performance, we conduct a human evaluation using the identical {\methodname} digital protocol administered to the models.
We sample 20 items per subtest, including all associated variants, yielding 1,540 questions in total.
We use the same task instructions and scoring rules as for the MLLMs.
We recruit 31 university students, ensuring each question is completed by three independent participants.
Table~\ref{tab:main-results} shows that humans achieve an accuracy of 78.8\% on average, confirming a substantial performance gap between the strongest model we test, {\bestmodel}, which achieves {\best}, and university participants.
Humans outperform MLLMs on nearly all subtests except RL2 (Diagramming Relationships), where success relies more on textual object knowledge, a known strength of MLLMs rather than visual reasoning.

\begin{table*}[t]
    \centering
    \caption{The performance of {\nummodel} models on {\methodname}. The rightmost column shows the total score. Darker scores show higher scores. The best model is {\bestmodel}, achieving {\best}.}
    \label{tab:main-results}
    \includegraphics[width=1.0\linewidth]{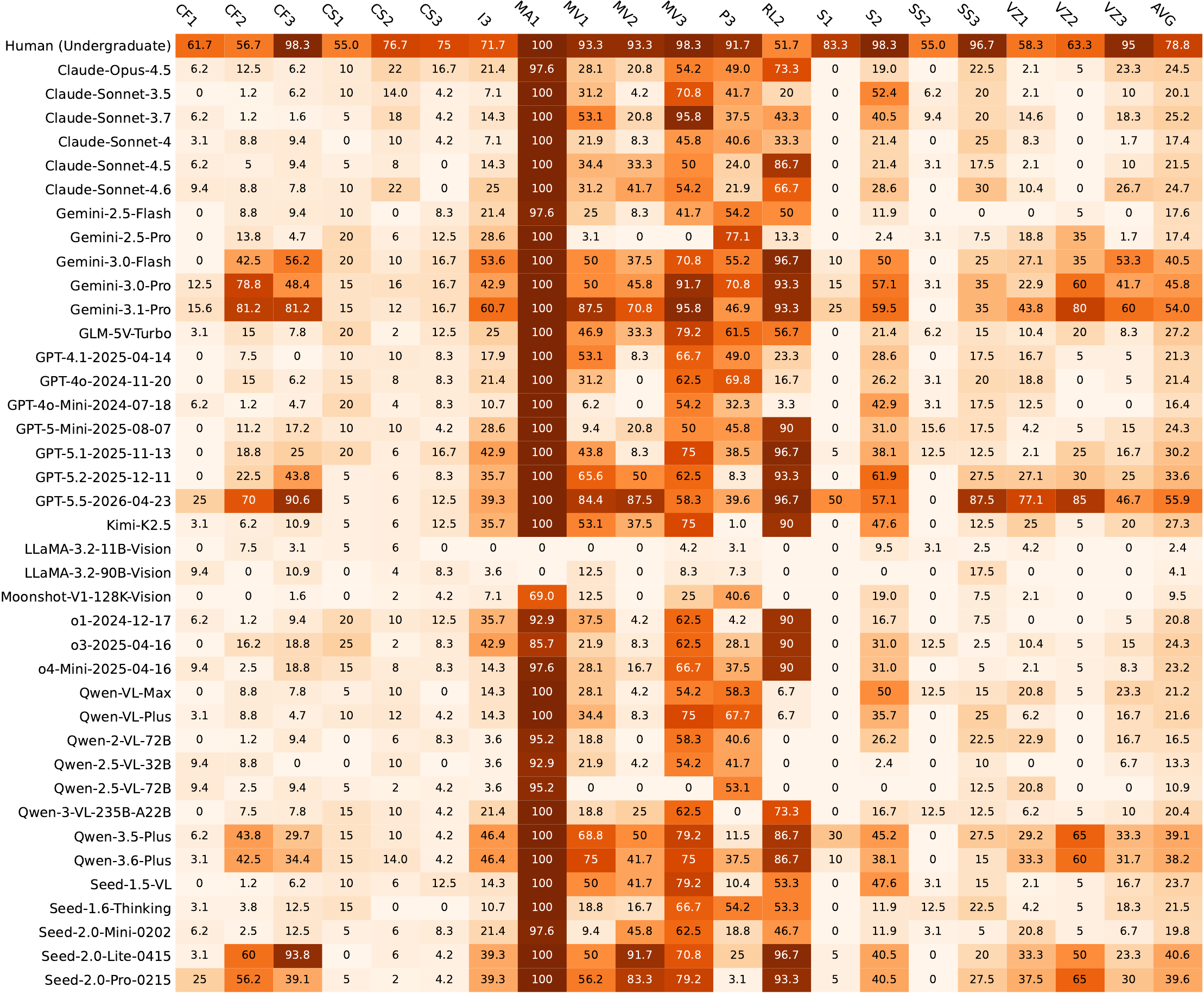}
\end{table*}

\subsection{Results}
\label{sec:results}

\paragraph{Performance by subtests}
Among the {\nummodel} evaluated frontier models, \textit{{\bestmodel}} achieves the highest overall score (\textit{\best}), yet still lags significantly behind human performance.
Models generally exhibit stronger performance on memorization tasks (MA1, MV1--MV3), indicating robust capabilities in attending to relevant input context.
Many models also perform well on RL2, as it primarily requires identifying commonalities among text-based concepts and demands minimal visual reasoning.
Furthermore, models succeed on Bongard problems (I3), likely due to the prevalence of existing benchmarks and extensive prior research focus in this area.
Conversely, models struggle across visualization and spatial reasoning subtests, with the VZ series proving particularly hard.

\paragraph{The ``Middle Score Anomaly''~\cite{babaiee2025visual}}
This phenomenon refers to models unexpectedly achieving intermediate performance—neither random nor near-perfect—on tasks that are extremely easy for humans.
For instance, the P3 (Identical Pictures Test) simply requires determining whether two images depict the same object.
Humans can either solve this task almost perfectly or fail entirely (\ie, perform at chance level if they lack the necessary perceptual ability).
It would be highly unusual for a human to achieve, say, 70\% accuracy on this task—suggesting partial understanding but inexplicable failures.
However, as shown in Figure~\ref{fig:middle-score-anomaly}, we observe that most models obtain 30--60\% accuracy on P3, while random guessing yields only 3.13\%.
We interpret this as evidence that current models lack genuine reasoning capabilities, at least in the context of the tasks presented in {\methodname}.
Our further failure analysis (\S\ref{sec:failure}) reveals that the apparent strengths of current MLLMs often stem from concept-level recognition rather than genuine cognitive visual processing.

\paragraph{Performance by models}
Among earlier models, performance does not strictly correlate with parameter size or release iteration.
For instance, Qwen-2.5-72B is outperformed by both its smaller counterpart, Qwen-2.5-32B, and its predecessor, Qwen-2-72B.
Similar trend inversions are observed across other model families: Claude-3.7 outperforms Claude-4, GPT-4o surpasses GPT-4.1, and Seed-1.5 exceeds Seed-1.6.
However, extending the temporal scale reveals a distinct paradigm shift.
While earlier models plateaued at lower performance levels (\eg, 30.2 for GPT-5.1), models released from late 2025 onward demonstrate a substantial breakthrough.
Beginning with Gemini-3.0 and continuing through 2026 with models like GPT-5.5, Qwen-3.5, and Seed-2.0, scores have surpassed 40, with Gemini and GPT reaching peaks of 54.0 and 55.9.
We attribute this rapid progression to a growing industry focus on foundational visual capabilities in 2026.
As models increasingly advance, our parametric generation approach becomes critical to future-proof the benchmark.

\paragraph{Impact of reasoning efforts}
We evaluate the effect of CoT prompting across three GPT-4 models (GPT-4.1-2025-04-14, GPT-4o-2024-11-20, and GPT-4o-Mini-2024-07-18) alongside varying reasoning efforts (none, low, high) for GPT-5.1-2025-11-13.
As shown in Table~\ref{tab:reasoning}, CoT yields only marginal performance improvements for the GPT-4 family.
Furthermore, Pearson correlation coefficients between CoT token count and accuracy are negative ($-0.18$, $-0.28$, and $-0.35$, respectively), indicating that generation length is not a reliable proxy for reasoning quality on {\methodname}.
Conversely, GPT-5.1 exhibits substantial performance gains as reasoning effort increases.
This indicates that reasoning models benefit from extended CoT, whereas non-reasoning models which lack specialized training gain little or no improvement from longer chains.

\paragraph{Verbal thinking hurts performance~\cite{liu2025mind}}
We observe a pattern that longer CoT hurts performance: for example, in CF3, our participants reported to rely on holistic spatial judgment directly from the start to the end point (``jumping steps'') instead of tracing paths sequentially, whereas MLLMs' text-based reasoning forces step-by-step traversal, leading to errors.
These behaviors highlight a structural mismatch between human intuitive spatial reasoning and text-mediated reasoning in current MLLMs.
This aligns with recent findings showing that CoT does not universally enhance model performance; in fact, certain cognitive tasks may exhibit degraded performance with CoT.
Studies in cognitive psychology have shown that more verbalization can hurt human performance in certain visual or holistic reasoning tasks~\cite{schooler1990verbal, dijksterhuis2004think, van2008intentional}.
Specifically, we observe declines in performance on perceptual and closure tasks (P3, CS2) and spatial visualization tasks (SS3, VZ1).
Conversely, CoT consistently improves performance on reasoning tasks such as I3 and RL2, consistent with prior results that CoT is particularly effective on objective tasks like math and coding~\cite{sprague2025cot}.

\paragraph{Influence of temperatures}
To assess model robustness against temperatures, we evaluate temperatures 0.5 and 1.0 for three models: GPT-4.1-2025-04-14, GPT-4o-2024-11-20, and GPT-4o-Mini-2024-07-18.
As shown in Table~\ref{tab:temperature}, the overall performance fluctuates only marginally across temperature settings, and the total score remains stable, indicating that our conclusions are not sensitive to the choice of decoding temperature.

\begin{table*}[t]
    \centering
    \caption{The performance of the GPT-5.5 model on the generated subsets in {\methodname}. The ``Original'' row reports performance on the original FRCT questions. The ``Normal'' row uses the same configuration as the original questions. The ``Easy'' and ``Hard'' rows correspond to questions that are modified to be easier and more difficult, respectively.}
    \label{tab:generated-results-5}
    \includegraphics[width=1.0\linewidth]{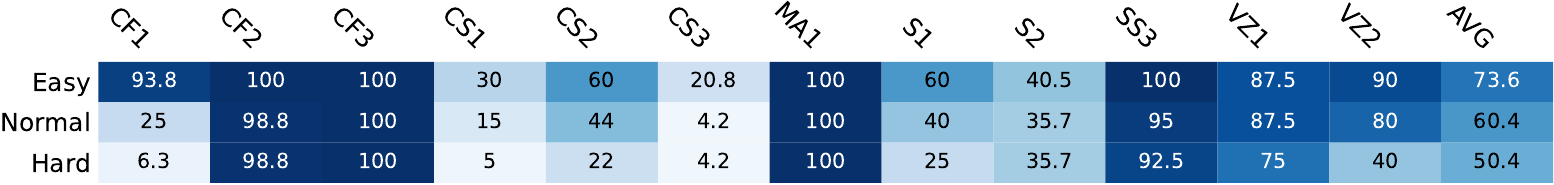}
\end{table*}

\subsection{Performance on generated tests}

Using our generation algorithms, we first construct a ``Normal'' subset in which each configuration closely mirrors the original FRCT questions.
We then create ``Easy'' and ``Hard'' subsets by systematically adjusting parameters that modulate task difficulty.
For instance, we vary the grid size for CF1, CF2, CF3, SS3, and VZ1; the noise severity for CS1, CS2, and CS3; the number of item pairs to be memorized in MA1; and the number of folds in VZ2.

We evaluate two models: GPT-5.5-2026-04-23 and GPT-4.1-2025-04-14.
The results are presented in Table~\ref{tab:generated-results-5} and Table~\ref{tab:generated-results-4}, respectively.
Overall, model performance decreases progressively across the easy, normal, and hard subsets.
However, for GPT-4.1 the gap is relatively smaller (28.9, 23.2, and 22.0), and the performance can even increase on some hard subsets (CF2, CS3, MA1, SS3, and VZ1).
We believe that there is a floor-effect: GPT-4.1 is simply too weak on the generated subsets to reveal the intended gradient.
Using a substantially stronger model, GPT-5.5, the intended monotonous decrease emerges clearly, with no subsets increase, confirming that our parameters (grid size, noise severity, number of folds, number of pairs, \etc) genuinely modulate difficulty.

Increasing the grid size, GPT-5.5 still achieves almost 100\% on CF2 and CF3, while it cannot handle CF1 well yet.
For CS1--CS3, GPT-4.1 surprisingly outperforms GPT-5.5.
Finally, on the domain of visualization \& spatial reasoning (from S1 to VZ2), GPT-5.5 performs better than expected, even when our version expands VZ2 (Paper Folding Test) to include up to five folds.
Note that we can change the parameters to further increase the difficulty, our {\methodname} is not saturated yet.
\subsection{A case study: how do models master MA1?}
\label{sec:failure-MA1}

\begin{figure*}[h]
  \centering
  \subfloat[MA1 generated with CF2 figures]{
    \includegraphics[width=0.48\linewidth]{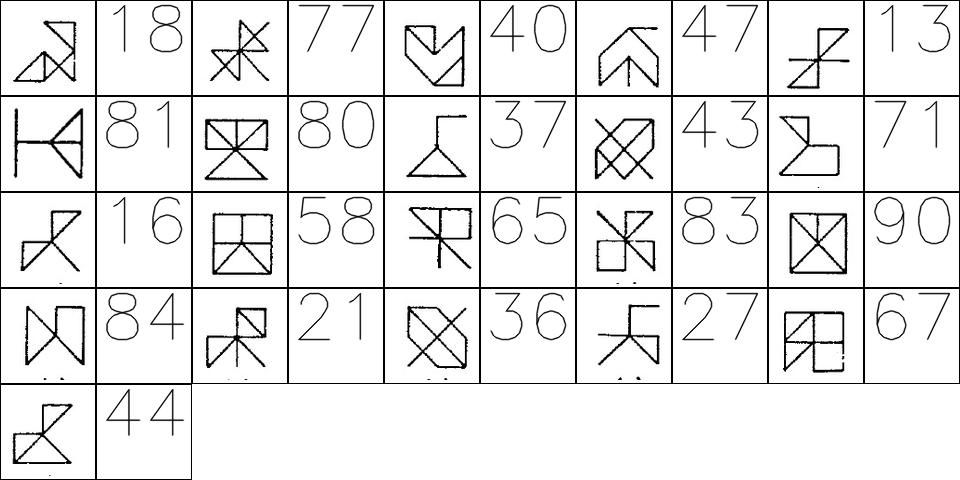}
    \label{fig:CF2-generated-MA1}
  } \hfill
  \subfloat[MA1 generated with MV1 figures.]{
    \includegraphics[width=0.48\linewidth]{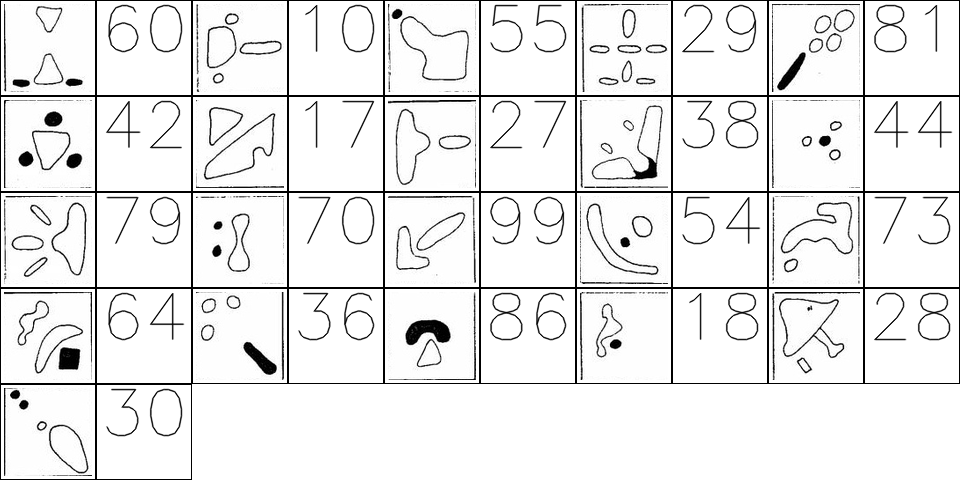}
    \label{fig:MV1-generated-MA1}
  }
  \caption{An example of our generated MA1 image-number pairs using CF2 and MV1 figures.}
\end{figure*}

Given that models achieve very high accuracy on this memory test, we further investigate the mechanisms underlying their performance.
An intuitive hypothesis is that models translate visual cues into high-level, human-interpretable concepts (\eg, ``soccer,'' ``chair,'' ``fish'') and memorize the concept–number pairs, rather than the raw image patterns.
To test this hypothesis, we use CF2-generated images, which consist only of lines arranged in a $3 \times 3$ grid, to create MA1 test cases via our automatic generation algorithm (see Figure~\ref{fig:CF2-generated-MA1} for an example).
We generate datasets with varying numbers of image–number pairs, ranging from 10 to 80, and evaluate GPT-4.1, Claude-3.7, and Qwen-VL-Max (results in Table~\ref{tab:MA1-generated}).
For semantically rich images, all three models maintain strong performance across different pair counts.
In contrast, accuracy declines sharply with abstract CF2 images.
As the number of pairs increases, GPT-4.1 demonstrates the greatest robustness, retaining 33.3\% accuracy at 80 pairs.
Claude-3.7 performs moderately, while Qwen-VL-Max fails at 40 pairs.
We further construct test cases using abstract figures from MV1 (Figure~\ref{fig:MV1-generated-MA1}) at 20 pairs.
The three models achieve accuracies of 81.0\%, 42.9\%, and 54.8\%, respectively, consistent with their performance on CF2-generated tests.
These results together suggest that models rely heavily on interpretable, concept-level representations rather than low-level visual patterns.

To ensure that the performance drop is not simply due to distributional shift, we generate extreme yet valid visual combinations using diffusion models (\eg, ``a horse on the moon'').
In these cases, the model maintains high accuracy, further supporting our hypothesis: the model performs well as long as the visual input can be mapped to familiar, conceptual categories.
This hypothesis is further supported by the analysis of P3, where high-performing examples typically involve easily verbalizable content, while failures are associated with visually complex and linguistically demanding patterns.
These results also suggest that models struggle to interpret abstract visual patterns such as the line-based CF2 stimuli, reinforcing the idea that their success depends on concept recognition rather than low-level perception.

\begin{table*}[t]
    \centering
    \caption{MA1 performance of three models using different image sources and pair numbers.}
    \label{tab:MA1-generated}
    \begin{tabular}{lcccccccc}
    \toprule
    \multirow{2}{*}{\bf Number of Pairs} & \multicolumn{2}{c}{\bf 10} & \multicolumn{2}{c}{\bf 20} & \multicolumn{2}{c}{\bf 40} & \multicolumn{2}{c}{\bf 80} \\
    \cmidrule(lr){2-3} \cmidrule(lr){4-5} \cmidrule(lr){6-7} \cmidrule(lr){8-9}
    & MA1 & CF2 & MA1 & CF2 & MA1 & CF2 & MA1 & CF2 \\
    \midrule
    GPT-4.1-2025-04-14 & 90.48 & 78.57 & 83.33 & 57.14 & 88.10 & 52.38 & 92.86 & 33.33 \\
    Claude-3.7-Sonnet  & 97.62 & 73.81 & 73.81 & 45.24 & 85.71 & 38.10 & 85.71 & 9.52 \\
    Qwen-VL-Max-250408 & 97.62 & 83.33 & 88.10 & 47.62 & 90.48 & 2.38  & 73.81 & 7.14 \\
    \bottomrule
    \end{tabular}
\end{table*}
\section{Related work}

\paragraph{Evaluation with natural images}
Natural images are commonly used to evaluate the visual capabilities of MLLMs, as they more closely reflect real-world scenarios~\cite{zhao2024lova3, liu2024mmbench, chow2025physbench, wadhawan2024contextual}.
Recent research has emphasized MLLMs' spatial reasoning abilities~\cite{kamath2023s, liu2023visual, yang2025thinking}, including tasks such as top-view map interpretation~\cite{li2024topviewrs} and region-level depth reasoning~\cite{cheng2024spatialrgpt}.
However, we argue that natural images often introduce additional noise and variability, making them less suitable for assessing core visual competencies.
While benchmarks such as Blink~\cite{fu2024blink}, MMT-Bench~\cite{ying2024mmt}, HallusionBench~\cite{guan2024hallusionbench}, and CoreCognition~\cite{li2025core} incorporate synthetic images for tasks like IQ tests, visual hallucination detection, and physical reasoning, their overall focus remains primarily on natural image settings.

\begin{table*}[t]
    \centering
    \caption{Comparison of {\methodname} and other vision benchmarks. \#T: number of tasks; \#Q: number of queries; P: psychological grounding; G: generation of new test cases; D: different difficulties for a task; M: rigorous measurement; I: \textbf{N}atural or \textbf{S}ynthetic images; PR: pattern recognition; BM: Bongard/matrix reasoning; MM: memory; RT: rotation; MZ: maze; PZ: puzzle.}
    \label{tab:review}
    \resizebox{1.0\linewidth}{!}{
    \setlength{\tabcolsep}{2pt}
    \rowcolors{2}{blue!10}{white}
    \begin{tabular}{lcclcccccccccc}
        \toprule
        \bf Benchmarks & \bf \#T & \bf \#Q & \multicolumn{1}{c}{\bf P} & \bf G & \bf D & \bf M & \bf I & \bf PR & \bf BM & \bf MM & \bf RT & \bf MZ & \bf PZ \\
        \midrule
        ARC-AGI-2 \cite{chollet2025arc} & 1 & 240 & \cmark & \xmark & \cmark & N/A & S & \cmark & \cmark & \xmark & \cmark & \xmark & \cmark \\
        BlindTest \cite{rahmanzadehgervi2024vision} & 7 & 9,720 & \cmark & \cmark & \cmark & \cmark & S & \xmark & \xmark & \xmark & \xmark & \xmark & \xmark \\
        BLINK \cite{fu2024blink} & 14 & 3,807 & \cmark~IQ & \xmark & \xmark & \xmark & NS & \cmark & \cmark & \xmark & \cmark & \xmark & \cmark \\
        Bongard-HOI \cite{jiang2022bongard} & 4 & 53K & \cmark~BP & \xmark & \xmark & \xmark & N & \xmark & \cmark & \xmark & \xmark & \xmark & \xmark \\
        Bongard-LOGO \cite{nie2020bongard} & 3 & 12K & \cmark~BP & \cmark & \cmark & \xmark & S & \xmark & \cmark & \xmark & \xmark & \xmark & \xmark \\
        Bongard-OpenWorld \cite{wu2024bongard} & 10 & 1,010 & \cmark~BP & \xmark & \cmark & \xmark & N & \xmark & \cmark & \xmark & \xmark & \xmark & \xmark \\
        ConceptARC \cite{moskvichev2023conceptarc} & 16 & 480 & \cmark & \xmark & \cmark & N/A & S & \cmark & \cmark & \xmark & \xmark & \xmark & \xmark \\
        ConTextual \cite{wadhawan2024contextual} & 8 & 506 & \xmark & \xmark & \cmark & N/A & NS & \xmark & \xmark & \xmark & \xmark & \xmark & \xmark \\
        CoreCognition \cite{li2025core} & 12 & 1,503 & \cmark & \xmark & \xmark & \cmark & NS & \xmark & \xmark & \cmark & \cmark & \xmark & \xmark \\
        EvalQABench \cite{zhao2024lova3} & 9 & 69K & \xmark & \xmark & \xmark & \xmark & N & \xmark & \xmark & \xmark & \xmark & \xmark & \xmark \\
        GamiBench \cite{spencer2025gamibench} & 3 & 744 & \xmark & \xmark & \cmark & \cmark & S & \xmark & \xmark & \xmark & \xmark & \xmark & \cmark \\
        GVIL \cite{zhang2023grounding} & 4 & 1,600 & \cmark & \xmark & \xmark & \cmark & NS & \xmark & \xmark & \xmark & \xmark & \xmark & \xmark \\
        HallusionBench \cite{guan2024hallusionbench} & 9 & 1,129 & \cmark & \xmark & \cmark & \cmark & NS & \xmark & \xmark & \xmark & \xmark & \xmark & \xmark \\
        IllusionVQA \cite{shahgir2024illusionvqa} & 13 & 1,435 & \cmark & \xmark & \xmark & \xmark & NS & \cmark & \xmark & \xmark & \xmark & \xmark & \xmark \\
        M3GIA \cite{song2024m3gia} & 18 & 1,800 & \cmark~RPM & \xmark & \cmark & \xmark & NS & \cmark & \cmark & \xmark & \cmark & \xmark & \cmark \\
        MaRs-VQA \cite{cao2025visual} & 18 & 1,440 & \cmark~RPM & \xmark & \xmark & \xmark & S & \xmark & \cmark & \xmark & \xmark & \xmark & \xmark \\
        MathVista \cite{lu2024mathvista} & 7 & 6,141 & \cmark~IQ & \xmark & \cmark & \xmark & NS & \xmark & \cmark & \xmark & \xmark & \xmark & \xmark \\
        Mensa \& IT \cite{zhang2024far} & 2 & 101 & \cmark~RPM & \xmark & \cmark & \xmark & S & \cmark & \cmark & \xmark & \xmark & \xmark & \xmark \\
        MMBench \cite{liu2024mmbench} & 20 & 3,217 & \xmark & \xmark & \xmark & \cmark & NS & \xmark & \xmark & \xmark & \xmark & \xmark & \xmark \\
        MMT-Bench \cite{ying2024mmt} & 162 & 31,325 & \cmark~RPM & \xmark & \xmark & \xmark & NS & \cmark & \cmark & \xmark & \xmark & \xmark & \cmark \\
        Mv-Math \cite{wang2025mv} & 11 & 2,009 & \xmark & \xmark & \cmark & N/A & S & \cmark & \xmark & \xmark & \cmark & \cmark & \cmark \\
        NLVR \cite{wu2024surprising} & 2 & 92K & \xmark & \xmark & \xmark & \xmark & S & \xmark & \xmark & \xmark & \xmark & \xmark & \xmark \\
        PhysBench \cite{chow2025physbench} & 19 & 10,002 & \xmark & \xmark & \xmark & \xmark & N & \xmark & \xmark & \xmark & \xmark & \xmark & \xmark \\
        RAVEN \cite{zhang2019raven} & 7 & 70K & \cmark~RPM & \cmark & \xmark & \xmark & S & \xmark & \cmark & \xmark & \xmark & \xmark & \xmark \\
        Shape-Blind \cite{rudman2025forgotten} & 7 & 7,590 & \xmark & \cmark & \xmark & N/A & S & \xmark & \xmark & \xmark & \xmark & \xmark & \xmark \\
        SPACE \cite{ramakrishnan2025does} & 15 & 2,377 & \cmark~IQ & \cmark & \cmark & \xmark & NS & \cmark & \xmark & \cmark & \cmark & \cmark & \cmark \\
        SpatialRGPT-Bench \cite{cheng2024spatialrgpt} & 12 & 1,406 & \xmark & \xmark & \xmark & N/A & N & \xmark & \xmark & \xmark & \xmark & \xmark & \xmark \\
        TopViewRS \cite{li2024topviewrs} & 9 & 11,384 & \xmark & \xmark & \xmark & \xmark & N & \xmark & \xmark & \xmark & \xmark & \xmark & \xmark \\
        VisualGraphArena \cite{babaiee2025visual} & 6 & 578K & \xmark & \cmark & \cmark & \xmark & S & \xmark & \xmark & \xmark & \xmark & \xmark & \xmark \\
        VisualPuzzles \cite{song2026visualpuzzles} & 5 & 1,168 & \cmark~IQ & \xmark & \cmark & \xmark & NS & \cmark & \cmark & \xmark & \xmark & \xmark & \cmark \\
        VisualSphinx \cite{feng2025visualsphinx} & 8 & 663K & \cmark~IQ & \cmark & \cmark & \cmark & S & \xmark & \cmark & \xmark & \cmark & \xmark & \cmark \\
        VSI-Bench \cite{yang2025thinking} & 8 & 5,131 & \cmark & \xmark & \xmark & N/A & N & \xmark & \xmark & \cmark & \xmark & \xmark & \xmark \\
        VSR \cite{liu2023visual} & 7 & 10,972 & \cmark & \xmark & \xmark & \xmark & N & \xmark & \xmark & \xmark & \xmark & \xmark & \xmark \\
        What'sUp \cite{kamath2023s} & 3 & 4,958 & \xmark & \xmark & \xmark & \cmark & N & \xmark & \xmark & \xmark & \xmark & \xmark & \xmark \\
        
        {\methodname} (Ours) & 20 & 3046 & \cmark~FRCT & \cmark & \cmark & \cmark & S & \cmark & \cmark & \cmark & \cmark & \cmark & \cmark \\
        
        \bottomrule
    \end{tabular}
    }
\end{table*}

\paragraph{Evaluation with synthetic images}
Synthetic images have been widely employed to evaluate the fundamental visual reasoning capabilities of MLLMs~\cite{rahmanzadehgervi2024vision, wu2024surprising, chollet2025arc, moskvichev2023conceptarc}.
Prior work has leveraged tasks such as Raven's Progressive Matrices~\cite{zhang2024far, song2024m3gia, cao2025visual, zhang2019raven} and the Logic Test from the Chinese Civil Service Examination~\cite{song2026visualpuzzles}, which include puzzles conceptually related to our I3 task.
VisualSphinx~\cite{feng2025visualsphinx} further extends this line of work by generating puzzles structurally similar to RPMs.
Mental Rotation Tests have also been frequently used~\cite{ramakrishnan2025does, song2024m3gia}, aligning with the design of our S1 and S2 tasks.
In addition, synthetic images have supported evaluations of MLLMs on mathematical reasoning problems~\cite{lu2024mathvista, wang2025mv}, including polygons~\cite{rudman2025forgotten} and graph-based challenges~\cite{babaiee2025visual}.
A comprehensive comparison is shown in Table~\ref{tab:review}.
Our proposed {\methodname} advances this direction by providing a more comprehensive evaluation framework for core visual abilities, including 20 tests, systematically grounded in factor analysis from cognitive science.
Furthermore, we implement automatic generation for 12 tests, enabling unlimited training data and ensuring the long-term scalability of the benchmark through higher difficulty levels.

\paragraph{Enhancing MLLMs' visual ability}
A range of strategies have been proposed to strengthen spatial reasoning in MLLMs, including generating intermediate steps~\cite{li2025imagine, wu2024mind}, drawing auxiliary lines~\cite{meng2023chain, hu2024visual}, incorporating coordinates or depth cues~\cite{liu2025spatialcot, cai2025spatialbot}, and augmenting training sets with reasoning data~\cite{shao2024visual}.
Our approach enables automatic generation of high-quality, difficulty-controlled test cases, offering unlimited training data to enhance MLLMs' visual reasoning.

\paragraph{Using psychological tests on AI}
Recent studies have evaluated AI models from psychological perspectives, including behavioral analysis~\cite{coda2024cogbench}, personality~\cite{huang2024apathetic, huang2024reliability}, emotion~\cite{huang2024apathetic}, and mental disorder~\cite{coda2023inducing}.
Research has found advanced human-like abilities in AI models, including Theory-of-Mind abilities~\cite{liu2024interintent, liang2023leveraging, huang2025competing} and role-playing abilities~\cite{ng2024well, wang2024incharacter, wang2025coser}.
Inspired by cognitive science, our work provides a comprehensive framework for evaluating foundational visual abilities.
\section{Discussion}
\label{sec:discussion}

\paragraph{Directions for mitigation}

(1) Training data via our generator: The parametric generator produces unlimited, difficulty-controlled instances (12 subtests) with code-guaranteed correctness.
This supports bottom-up or curriculum training, mastering simple visual factors before complex reasoning.
And, for tests like Paper Folding (VZ2) and Copying (CF3), this can emit intermediate solution steps for process-level SFT.
(2) Targeting the failure modes we identify: For example, the reliance on verbalizable concepts suggests strengthening low-level visual representations; the 45° diagonal bias suggests augmenting with continuous geometric-perception data; and ``longer CoT hurts'' on spatial tasks suggests visual or latent intermediate representations over verbal chains.

\paragraph{Limitations}

(1) Applicability \& Scope: Our framework is restricted to MLLMs that support multi-image inputs and produce text-only outputs. Consequently, it cannot evaluate MLLMs restricted to single-image inputs, nor does it include the four FRCT visual tasks that require image generation.
(2) Incomplete Parametric Generation: While our automatic test-case generator guarantees label correctness via strict code-based rendering—obviating the need for human verification—it currently supports only 12 of the 20 subtests. The remaining subtests are challenging to parameterize programmatically. We deliberately exclude text-to-image models for generation to avoid introducing unverifiable artifacts into the dataset.
(3) Focus on Models without External Tools: We evaluate the baseline capabilities of foundation models directly. We do not assess tool-augmented MLLMs or agentic frameworks, as our primary objective is to isolate and measure the inherent abilities of the foundation models themselves.
(4) Single-Pass Evaluation: Due to the prohibitive computational and API costs associated with comprehensively evaluating {\nummodel} MLLMs—particularly frontier models such as o1—we constrain our evaluation to a single run per model. To mitigate the variance typically addressed by multiple runs, our benchmark design (\S\ref{sec:variants}) inherently restricts randomness, while the extensive number of queries per subtest ensures the statistical robustness of our reported metrics.
(5) Zero-Shot Evaluation: The evaluation of all {\nummodel} models is conducted strictly in a zero-shot setting. We omit few-shot prompting to prevent potential model misdirection since a single test example already contains a substantial number of input images.

\paragraph{Conclusion}

We present {\methodname}, the first benchmark to transpose twenty vision‐centric subtests from the \textit{Factor‑Referenced Cognitive Test} battery into an automated image–text setting.
A systematic evaluation of {\nummodel} MLLMs uncovers a striking gap: despite their prowess on holistic leaderboards, the best model attains only \textit{\best}, often performing near chance on tasks that human novices solve with ease.
CoT improves only marginally and sometimes even decreases the overall performance.
Exposing a missing substrate for genuine visual reasoning like the ``Middle Score Anomaly,'' we demonstrate a fundamental difference between human and AI in how they construct vision capabilities.
Hallucinated perception in safety‑critical applications, brittle spatial reasoning in robotics, and misaligned multimodal feedback loops may trace back to weak {\methodname} performance.
By releasing {\methodname} and its controllable‑difficulty generator, we aim to facilitate the fields of MLLM visual reasoning, cognitive interpretability, and synthetic data.

\begin{ack}
We thank the reviewers and area chairs of our submissions to NeurIPS 2026, ICML 2026, and ICLR 2026 who provided invaluable feedback that has substantially improved our work.
The work was mainly done when Jen-tse Huang, Youliang Yuan, Xiaoyuan Liu, and Wenxuan Wang were interns at Tencent AI Lab, where they received advise from two mentors, Wenxiang Jiao and Zhaopeng Tu.
The authors declare no funding that supports this work and no competing interests.
\end{ack}

\bibliography{References/benchmark, References/frct, References/model, References/reference}

@article{liu2024ocrbench,
  title={OCRBench: on the hidden mystery of OCR in large multimodal models},
  author={Liu, Yuliang and Li, Zhang and Huang, Mingxin and Yang, Biao and Yu, Wenwen and Li, Chunyuan and Yin, Xu-Cheng and Liu, Cheng-Lin and Jin, Lianwen and Bai, Xiang},
  journal={Science China Information Sciences},
  volume={67},
  number={12},
  pages={220102},
  year={2024},
  publisher={Springer}
}

@article{fu2025ocrbench,
  title={OCRBench v2: An Improved Benchmark for Evaluating Large Multimodal Models on Visual Text Localization and Reasoning},
  author={Fu, Ling and Kuang, Zhebin and Song, Jiajun and Huang, Mingxin and Yang, Biao and Li, Yuzhe and Zhu, Linghao and Luo, Qidi and Wang, Xinyu and Lu, Hao and others},
  journal={Advances in Neural Information Processing Systems},
  volume={38},
  year={2025}
}

@inproceedings{yue2024mmmu,
  title={Mmmu: A massive multi-discipline multimodal understanding and reasoning benchmark for expert agi},
  author={Yue, Xiang and Ni, Yuansheng and Zhang, Kai and Zheng, Tianyu and Liu, Ruoqi and Zhang, Ge and Stevens, Samuel and Jiang, Dongfu and Ren, Weiming and Sun, Yuxuan and others},
  booktitle={Proceedings of the IEEE/CVF conference on computer vision and pattern recognition},
  pages={9556--9567},
  year={2024}
}

@inproceedings{yue2025mmmu,
  title={Mmmu-pro: A more robust multi-discipline multimodal understanding benchmark},
  author={Yue, Xiang and Zheng, Tianyu and Ni, Yuansheng and Wang, Yubo and Zhang, Kai and Tong, Shengbang and Sun, Yuxuan and Yu, Botao and Zhang, Ge and Sun, Huan and others},
  booktitle={Proceedings of the 63rd Annual Meeting of the Association for Computational Linguistics (Volume 1: Long Papers)},
  pages={15134--15186},
  year={2025}
}

@article{ye2024gmai,
  title={Gmai-mmbench: A comprehensive multimodal evaluation benchmark towards general medical ai},
  author={Ye, Jin and Wang, Guoan and Li, Yanjun and Deng, Zhongying and Li, Wei and Li, Tianbin and Duan, Haodong and Huang, Ziyan and Su, Yanzhou and Wang, Benyou and others},
  journal={Advances in Neural Information Processing Systems},
  volume={37},
  pages={94327--94427},
  year={2024}
}

@inproceedings{zuo2025medxpertqa,
  title={MedXpertQA: Benchmarking Expert-Level Medical Reasoning and Understanding},
  author={Zuo, Yuxin and Qu, Shang and Li, Yifei and Chen, Zhang-Ren and Zhu, Xuekai and Hua, Ermo and Zhang, Kaiyan and Ding, Ning and Zhou, Bowen},
  booktitle={International Conference on Machine Learning},
  pages={80961--80990},
  year={2025},
  organization={PMLR}
}

@inproceedings{yang2025embodiedbench,
  title={EmbodiedBench: Comprehensive Benchmarking Multi-modal Large Language Models for Vision-Driven Embodied Agents},
  author={Yang, Rui and Chen, Hanyang and Zhang, Junyu and Zhao, Mark and Qian, Cheng and Wang, Kangrui and Wang, Qineng and Koripella, Teja Venkat and Movahedi, Marziyeh and Li, Manling and others},
  booktitle={International Conference on Machine Learning},
  pages={70576--70631},
  year={2025},
  organization={PMLR}
}

@inproceedings{liu2024mmbench,
  title={Mmbench: Is your multi-modal model an all-around player?},
  author={Liu, Yuan and Duan, Haodong and Zhang, Yuanhan and Li, Bo and Zhang, Songyang and Zhao, Wangbo and Yuan, Yike and Wang, Jiaqi and He, Conghui and Liu, Ziwei and others},
  booktitle={European conference on computer vision},
  pages={216--233},
  year={2024},
  organization={Springer}
}

@article{zhao2024lova3,
  title={Lova3: Learning to visual question answering, asking and assessment},
  author={Zhao, Henry Hengyuan and Zhou, Pan and Gao, Difei and Bai, Zechen and Shou, Mike Zheng},
  journal={Advances in Neural Information Processing Systems},
  volume={37},
  pages={115146--115175},
  year={2024}
}

@inproceedings{fu2024blink,
  title={Blink: Multimodal large language models can see but not perceive},
  author={Fu, Xingyu and Hu, Yushi and Li, Bangzheng and Feng, Yu and Wang, Haoyu and Lin, Xudong and Roth, Dan and Smith, Noah A and Ma, Wei-Chiu and Krishna, Ranjay},
  booktitle={European Conference on Computer Vision},
  pages={148--166},
  year={2024},
  organization={Springer}
}

@inproceedings{ying2024mmt,
  title={MMT-Bench: A Comprehensive Multimodal Benchmark for Evaluating Large Vision-Language Models Towards Multitask AGI},
  author={Ying, Kaining and Meng, Fanqing and Wang, Jin and Li, Zhiqian and Lin, Han and Yang, Yue and Zhang, Hao and Zhang, Wenbo and Lin, Yuqi and Liu, Shuo and others},
  booktitle={International Conference on Machine Learning},
  pages={57116--57198},
  year={2024},
  organization={PMLR}
}

@inproceedings{guan2024hallusionbench,
  title={Hallusionbench: an advanced diagnostic suite for entangled language hallucination and visual illusion in large vision-language models},
  author={Guan, Tianrui and Liu, Fuxiao and Wu, Xiyang and Xian, Ruiqi and Li, Zongxia and Liu, Xiaoyu and Wang, Xijun and Chen, Lichang and Huang, Furong and Yacoob, Yaser and others},
  booktitle={Proceedings of the IEEE/CVF Conference on Computer Vision and Pattern Recognition},
  pages={14375--14385},
  year={2024}
}

@inproceedings{li2025core,
  title={Core Knowledge Deficits in Multi-Modal Language Models},
  author={Li, Yijiang and Gao, Qingying and Zhao, Tianwei and Wang, Bingyang and Sun, Haoran and Lyu, Haiyun and Hawkins, Robert D and Vasconcelos, Nuno and Golan, Tal and Luo, Dezhi and others},
  booktitle={International Conference on Machine Learning},
  pages={34379--34409},
  year={2025},
  organization={PMLR}
}

@inproceedings{chow2025physbench,
  title={Physbench: Benchmarking and enhancing vision-language models for physical world understanding},
  author={Chow, Wei and Mao, Jiageng and Li, Boyi and Seita, Daniel and Campagnolo Guizilini, Vitor and Wang, Yue},
  booktitle={International Conference on Learning Representations},
  volume={2025},
  pages={97959--98108},
  year={2025}
}

@inproceedings{wadhawan2024contextual,
  title={ConTextual: Evaluating Context-Sensitive Text-Rich Visual Reasoning in Large Multimodal Models},
  author={Wadhawan, Rohan and Bansal, Hritik and Chang, Kai-Wei and Peng, Nanyun},
  booktitle={International Conference on Machine Learning},
  pages={49733--49787},
  year={2024},
  organization={PMLR}
}

@article{cheng2024spatialrgpt,
  title={Spatialrgpt: Grounded spatial reasoning in vision-language models},
  author={Cheng, An-Chieh and Yin, Hongxu and Fu, Yang and Guo, Qiushan and Yang, Ruihan and Kautz, Jan and Wang, Xiaolong and Liu, Sifei},
  journal={Advances in Neural Information Processing Systems},
  volume={37},
  pages={135062--135093},
  year={2024}
}

@inproceedings{kamath2023s,
  title={What’s “up” with vision-language models? Investigating their struggle with spatial reasoning},
  author={Kamath, Amita and Hessel, Jack and Chang, Kai-Wei},
  booktitle={Proceedings of the 2023 Conference on Empirical Methods in Natural Language Processing},
  pages={9161--9175},
  year={2023}
}

@inproceedings{yang2025thinking,
  title={Thinking in space: How multimodal large language models see, remember, and recall spaces},
  author={Yang, Jihan and Yang, Shusheng and Gupta, Anjali W and Han, Rilyn and Fei-Fei, Li and Xie, Saining},
  booktitle={Proceedings of the Computer Vision and Pattern Recognition Conference},
  pages={10632--10643},
  year={2025}
}

@inproceedings{li2024topviewrs,
  title={TopViewRS: Vision-Language Models as Top-View Spatial Reasoners},
  author={Li, Chengzu and Zhang, Caiqi and Zhou, Han and Collier, Nigel and Korhonen, Anna and Vuli{\'c}, Ivan},
  booktitle={Proceedings of the 2024 Conference on Empirical Methods in Natural Language Processing},
  pages={1786--1807},
  year={2024}
}

@article{liu2023visual,
  title={Visual Spatial Reasoning},
  author={Liu, Fangyu and Emerson, Guy and Collier, Nigel},
  journal={Transactions of the Association for Computational Linguistics},
  volume={11},
  pages={635--651},
  year={2023}
}

@inproceedings{ramakrishnan2025does,
  title={Does Spatial Cognition Emerge in Frontier Models?},
  author={Ramakrishnan, Santhosh Kumar and Wijmans, Erik and Kraehenbuehl, Philipp and Koltun, Vladlen},
  booktitle={The Thirteenth International Conference on Learning Representations},
  year={2025}
}

@inproceedings{rahmanzadehgervi2024vision,
  title={Vision language models are blind},
  author={Rahmanzadehgervi, Pooyan and Bolton, Logan and Taesiri, Mohammad Reza and Nguyen, Anh Totti},
  booktitle={Proceedings of the Asian Conference on Computer Vision},
  pages={18--34},
  year={2024}
}

@inproceedings{zhang2019raven,
  title={Raven: A dataset for relational and analogical visual reasoning},
  author={Zhang, Chi and Gao, Feng and Jia, Baoxiong and Zhu, Yixin and Zhu, Song-Chun},
  booktitle={Proceedings of the IEEE/CVF conference on computer vision and pattern recognition},
  pages={5317--5327},
  year={2019}
}

@inproceedings{cao2025visual,
  title={What is the visual cognition gap between humans and multimodal llms?},
  author={Cao, Xu and Shen, Yifan and Lai, Bolin and Ye, Wenqian and Ma, Yunsheng and Heintz, Joerg and Chen, Jintai and Huang, Meihuan and Cao, Jianguo and Zhang, Aidong and others},
  booktitle={The Second Conference on Language Modeling},
  year={2025}
}

@article{song2024m3gia,
  title={M3GIA: A Cognition Inspired Multilingual and Multimodal General Intelligence Ability Benchmark},
  author={Song, Wei and Li, Yadong and Xu, Jianhua and Wu, Guowei and Ming, Lingfeng and Yi, Kexin and Luo, Weihua and Li, Houyi and Du, Yi and Guo, Fangda and others},
  journal={arXiv preprint arXiv:2406.05343},
  year={2024}
}

@inproceedings{zhang2024far,
  title={How Far Are We from Intelligent Visual Deductive Reasoning?},
  author={Zhang, Yizhe and Bai, He and Zhang, Ruixiang and Gu, Jiatao and Zhai, Shuangfei and Susskind, Josh and Jaitly, Navdeep},
  booktitle={The First Conference on Language Modeling},
  year={2024}
}

@article{wu2024surprising,
  title={A Surprising Failure? Multimodal LLMs and the NLVR Challenge},
  author={Wu, Anne and Brantley, Kiant{\'e} and Artzi, Yoav},
  journal={arXiv preprint arXiv:2402.17793},
  year={2024}
}

@article{feng2025visualsphinx,
  title={VisualSphinx: Large-Scale Synthetic Vision Logic Puzzles for RL},
  author={Feng, Yichen and Xu, Zhangchen and Jiang, Fengqing and Li, Yuetai and Ramasubramanian, Bhaskar and Niu, Luyao and Lin, Bill Yuchen and Poovendran, Radha},
  journal={arXiv preprint arXiv:2505.23977},
  year={2025}
}

@article{chollet2025arc,
  title={Arc-agi-2: A new challenge for frontier ai reasoning systems},
  author={Chollet, Francois and Knoop, Mike and Kamradt, Gregory and Landers, Bryan and Pinkard, Henry},
  journal={arXiv preprint arXiv:2505.11831},
  year={2025}
}

@article{moskvichev2023conceptarc,
  title={The ConceptARC Benchmark: Evaluating Understanding and Generalization in the ARC Domain},
  author={Moskvichev, Arsenii Kirillovich and Odouard, Victor Vikram and Mitchell, Melanie},
  journal={Transactions on machine learning research},
  year={2023}
}

@inproceedings{song2026visualpuzzles,
  title={VisualPuzzles: Decoupling Multimodal Reasoning Evaluation from Domain Knowledge},
  author={Song, Yueqi and Ou, Tianyue and Kong, Yibo and Li, Zecheng and Neubig, Graham and Yue, Xiang},
  booktitle={Proceedings of the 43rd International Conference on Machine Learning},
  pages={114860--114932},
  volume={306},
  year={2026}
}

@article{spencer2025gamibench,
  title={GamiBench: Evaluating Spatial Reasoning and 2D-to-3D Planning Capabilities of MLLMs with Origami Folding Tasks},
  author={Spencer, Ryan and Yaari, Roey and Vemavarapu, Ritvik and Yang, Joyce and Ngo, Steven and Sharma, Utkarsh},
  journal={arXiv preprint arXiv:2512.22207},
  year={2025}
}

@inproceedings{zhang2023grounding,
  title={Grounding visual illusions in language: Do vision-language models perceive illusions like humans?},
  author={Zhang, Yichi and Pan, Jiayi and Zhou, Yuchen and Pan, Rui and Chai, Joyce},
  booktitle={Proceedings of the 2023 conference on empirical methods in natural language processing},
  pages={5718--5728},
  year={2023}
}

@inproceedings{shahgir2024illusionvqa,
  title={IllusionVQA: A Challenging Optical Illusion Dataset for Vision Language Models},
  author={Shahgir, Haz Sameen and Sayeed, Khondker Salman and Bhattacharjee, Abhik and Ahmad, Wasi Uddin and Dong, Yue and Shahriyar, Rifat},
  booktitle={First Conference on Language Modeling},
  year={2024}
}

@article{nie2020bongard,
  title={Bongard-logo: A new benchmark for human-level concept learning and reasoning},
  author={Nie, Weili and Yu, Zhiding and Mao, Lei and Patel, Ankit B and Zhu, Yuke and Anandkumar, Anima},
  journal={Advances in neural information processing systems},
  volume={33},
  pages={16468--16480},
  year={2020}
}

@inproceedings{jiang2022bongard,
  title={Bongard-hoi: Benchmarking few-shot visual reasoning for human-object interactions},
  author={Jiang, Huaizu and Ma, Xiaojian and Nie, Weili and Yu, Zhiding and Zhu, Yuke and Anandkumar, Anima},
  booktitle={Proceedings of the IEEE/CVF conference on computer vision and pattern recognition},
  pages={19056--19065},
  year={2022}
}

@inproceedings{wu2024bongard,
  title={Bongard-OpenWorld: Few-Shot Reasoning for Free-form Visual Concepts in the Real World},
  author={Wu, Rujie and Ma, Xiaojian and Zhang, Zhenliang and Wang, Wei and Li, Qing and Zhu, Song-Chun and Wang, Yizhou},
  booktitle={The Twelfth International Conference on Learning Representations},
  year={2024}
}

@inproceedings{rudman2025forgotten,
  title={Forgotten polygons: Multimodal large language models are shape-blind},
  author={Rudman, William and Golovanevsky, Michal and Bar, Amir and Palit, Vedant and LeCun, Yann and Eickhoff, Carsten and Singh, Ritambhara},
  booktitle={Findings of the Association for Computational Linguistics: ACL 2025},
  pages={11983--11998},
  year={2025}
}

@inproceedings{babaiee2025visual,
  title={Visual Graph Arena: Evaluating Visual Conceptualization of Vision and Multimodal Large Language Models},
  author={Babaiee, Zahra and Kiasari, Peyman and Rus, Daniela and Grosu, Radu},
  booktitle={International Conference on Machine Learning},
  pages={2081--2113},
  year={2025},
  organization={PMLR}
}

@inproceedings{lu2024mathvista,
  title={MathVista: Evaluating Mathematical Reasoning of Foundation Models in Visual Contexts},
  author={Lu, Pan and Bansal, Hritik and Xia, Tony and Liu, Jiacheng and Li, Chunyuan and Hajishirzi, Hannaneh and Cheng, Hao and Chang, Kai-Wei and Galley, Michel and Gao, Jianfeng},
  booktitle={The Twelfth International Conference on Learning Representations},
  year={2024}
}

@inproceedings{wang2025mv,
  title={Mv-math: Evaluating multimodal math reasoning in multi-visual contexts},
  author={Wang, Peijie and Li, Zhong-Zhi and Yin, Fei and Ran, Dekang and Liu, Cheng-Lin},
  booktitle={Proceedings of the Computer Vision and Pattern Recognition Conference},
  pages={19541--19551},
  year={2025}
}

@book{ekstrom1976manual,
  title={Manual for kit of factor-referenced cognitive tests, 1976},
  author={Ekstrom, Ruth B and Harman, Harry Horace},
  year={1976},
  publisher={Educational testing service}
}

@article{thurstone1938primary,
  title={Primary mental abilities:},
  author={Thurstone, Louis Leon},
  journal={Psychology Monographs},
  volume={1},
  year={1938},
  publisher={Springer}
}

@book{ekstrom1973cognitive,
  title={Cognitive Factors: Some Recent Literature.},
  author={Ekstrom, Ruth B},
  year={1973},
  publisher={ERIC}
}

@book{guilford1967nature,
  title={The nature of human intelligence.},
  author={Guilford, Joy Paul},
  year={1967},
  publisher={McGraw-Hill}
}

@article{guilford1966sixteen,
  title={Sixteen divergent-production abilities at the ninth-grade level},
  author={Guilford, Joy Paul and Hoepfner, Ralph},
  journal={Multivariate Behavioral Research},
  volume={1},
  number={1},
  pages={43--66},
  year={1966},
  publisher={Taylor \& Francis}
}

@book{witkin1971manual,
  title={A manual for the embedded figures tests},
  author={Witkin, Herman A},
  year={1971},
  publisher={Consulting Psychologists Press}
}

@book{royce1973conceptual,
  title={The conceptual framework for a multifactor theory of individual ity. In" Multivariate Analysis and Psychological Theory"(JR Royce, ed.)},
  author={Royce, JR},
  year={1973},
  publisher={Academic Press, London and New York}
}

@article{hettema1968cognitive,
  title={Cognitive abilities as process variables.},
  author={Hettema, Joop},
  journal={Journal of personality and social psychology},
  volume={10},
  number={4},
  pages={461},
  year={1968},
  publisher={American Psychological Association}
}

@article{wardell1973possible,
  title={Possible changes in the taxonomies in Royce},
  author={Wardell, D},
  journal={Center for Advanced Study in Theoretical Psychology},
  pages={252--261},
  year={1973}
}

@book{cattell1971abilities,
  title={Abilities: Their structure, growth, and action.},
  author={Cattell, Raymond B},
  year={1971},
  publisher={Houghton Mifflin}
}

@book{carroll1974psychometric,
  title={Psychometric Tests As Cognitive Tasks: A New" Structure of Intellect." Technical Report No. 4.},
  author={Carroll, John B},
  year={1974},
  publisher={ERIC}
}

@article{frederiksen1967cognitive,
  title={Cognitive factors in the recognition of ambiguous auditory and visual stimuli.},
  author={Frederiksen, John R},
  journal={Journal of Personality and Social Psychology},
  volume={7},
  number={1p2},
  pages={1},
  year={1967},
  publisher={American Psychological Association}
}

@book{thurstone1944factorial,
  title={A factorial study of perception.},
  author={Thurstone, Louis Leon},
  year={1944},
  publisher={The University of Chicago Press}
}

@article{adcock1971flexibility,
  title={Flexibility and creativity},
  author={Adcock, Cyril J and Martin, William A},
  journal={The Journal of General Psychology},
  volume={85},
  number={1},
  pages={71--76},
  year={1971},
  publisher={Taylor \& Francis}
}

@article{messick1975dimensions,
  title={Dimensions of cognitive closure},
  author={Messick, Samuel and French, John W},
  journal={Multivariate behavioral research},
  volume={10},
  number={1},
  pages={3--16},
  year={1975},
  publisher={Taylor \& Francis}
}

@article{harris1971factor,
  title={A factor analytic interpretation strategy},
  author={Harris, Margaret L and Harris, Chester W},
  journal={Educational and Psychological Measurement},
  volume={31},
  number={3},
  pages={589--606},
  year={1971},
  publisher={Sage Publications Sage CA: Thousand Oaks, CA}
}

@article{dye1968growth,
  title={Growth changes in factorial structure by age and sex.},
  author={Dye, Neal W and Very, Philip S},
  journal={Genetic Psychology Monographs},
  year={1968},
  publisher={Heldref Publications}
}

@article{pawlick1966concepts,
  title={Concepts and calculations in human cognitive abilities},
  author={Pawlick, K},
  journal={Cattell, RB (Ed.), Handbook of multivariate experimental psychology},
  year={1966}
}

@article{thurstone1946theories,
  title={Theories of intelligence},
  author={Thurstone, Louis Leon},
  journal={The scientific monthly},
  volume={62},
  number={2},
  pages={101--112},
  year={1946},
  publisher={JSTOR}
}

@article{roff1952factorial,
  title={A factorial study of tests in the perceptual area},
  author={Roff, Merrill},
  journal={Psychometric Monograph},
  volume={8},
  year={1953}
}

@article{petrov1970memory,
  title={Memory structure as a psychic function},
  author={Petrov, Y. I.},
  journal={Voprosi Psikhologii},
  volume={16},
  pages={132--136},
  year={1970}
}

@article{shepard1971mental,
  title={Mental rotation of three-dimensional objects},
  author={Shepard, Roger N and Metzler, Jacqueline},
  journal={Science},
  volume={171},
  number={3972},
  pages={701--703},
  year={1971},
  publisher={American Association for the Advancement of Science}
}

@article{zimmerman1954influence,
  title={The influence of item complexity upon the factor composition of a spatial visualization test},
  author={Zimmerman, Wayne S},
  journal={Educational and Psychological Measurement},
  volume={14},
  number={1},
  pages={106--119},
  year={1954},
  publisher={Sage Publications Sage CA: Los Angeles, CA}
}

@article{werdelin1971relationship,
  title={The Relationship between Difficulty and Factor Loadings of some Visual-Perceptual Tests},
  author={Werdelin, Ingvar and Stjernberg, Gunnel},
  journal={Scandinavian Journal of Psychology},
  volume={12},
  number={1},
  pages={21--28},
  year={1971},
  publisher={Wiley Online Library}
}

@article{guilford1971analysis,
  title={The analysis of intelligence},
  author={Guilford, Joy Paul and Hoepfner, Ralph},
  journal={(No Title)},
  year={1971}
}

@article{kunnapas1969figural,
  title={Figural reversal rate and personal tempo},
  author={K{\"u}nnapas, Teodor},
  journal={Scandinavian journal of psychology},
  volume={10},
  number={1},
  pages={27--32},
  year={1969},
  publisher={Wiley Online Library}
}

@article{shepard1972chronometric,
  title={A chronometric study of mental paper folding},
  author={Shepard, Roger N and Feng, Christine},
  journal={Cognitive psychology},
  volume={3},
  number={2},
  pages={228--243},
  year={1972},
  publisher={Elsevier}
}

@book{ekstrom1974problems,
  title={Problems of Replication of Seven Divergent Production Factors. Technical Report No. 5.},
  author={Ekstrom, Ruth B and others},
  year={1974},
  publisher={ERIC}
}

@book{ekstrom1975attempt,
  title={An Attempt to Confirm Five Recently Identified Cognitive Factors. Technical Report No. 8.},
  author={Ekstrom, Ruth B and others},
  year={1975},
  publisher={ERIC}
}

@article{gpt4o-gpt4omini,
  title={{GPT-4o} system card},
  author={Hurst, Aaron and Lerer, Adam and Goucher, Adam P and Perelman, Adam and Ramesh, Aditya and Clark, Aidan and Ostrow, AJ and Welihinda, Akila and Hayes, Alan and Radford, Alec and others},
  journal={arXiv preprint arXiv:2410.21276},
  year={2024}
}

@misc{gpt41,
  title={Introducing {GPT-4.1} in the {API}},
  author={{OpenAI}},
  year={2025},
  howpublished={OpenAI Blog, Apr 14, 2025. \url{https://openai.com/index/gpt-4-1/}}
}

@article{gpt5-gpt5mini-gpt5nano,
  title={{OpenAI GPT-5} system card},
  author={Singh, Aaditya and Fry, Adam and Perelman, Adam and Tart, Adam and Ganesh, Adi and El-Kishky, Ahmed and McLaughlin, Aidan and Low, Aiden and Ostrow, AJ and Ananthram, Akhila and others},
  journal={arXiv preprint arXiv:2601.03267},
  year={2025}
}

@misc{gpt51,
  title={{GPT-5.1}: A smarter, more conversational {ChatGPT}},
  author={{OpenAI}},
  year={2025},
  howpublished={OpenAI Blog, Nov 12, 2025. \url{https://openai.com/index/gpt-5-1/}}
}

@misc{gpt52,
  title={Introducing {GPT-5.2}},
  author={{OpenAI}},
  year={2025},
  howpublished={OpenAI Blog, Dec 11, 2025. \url{https://openai.com/index/introducing-gpt-5-2/}}
}

@misc{gpt55,
  title={Introducing {GPT-5.5}},
  author={{OpenAI}},
  year={2026},
  howpublished={OpenAI Blog, Apr 23, 2026. \url{https://openai.com/index/introducing-gpt-5-5/}}
}

@article{o1,
  title={{OpenAI o1} system card},
  author={Jaech, Aaron and Kalai, Adam and Lerer, Adam and Richardson, Adam and El-Kishky, Ahmed and Low, Aiden and Helyar, Alec and Madry, Aleksander and Beutel, Alex and Carney, Alex and others},
  journal={arXiv preprint arXiv:2412.16720},
  year={2024}
}

@misc{o3-o4mini,
  title={Introducing {OpenAI o3} and {o4-mini}},
  author={{OpenAI}},
  year={2025},
  howpublished={OpenAI Blog, Apr 16, 2025. \url{https://openai.com/index/introducing-o3-and-o4-mini/}}
}

@article{gemini25,
  title={Gemini 2.5: Pushing the frontier with advanced reasoning, multimodality, long context, and next generation agentic capabilities},
  author={Comanici, Gheorghe and Bieber, Eric and Schaekermann, Mike and Pasupat, Ice and Sachdeva, Noveen and Dhillon, Inderjit and Blistein, Marcel and Ram, Ori and Zhang, Dan and Rosen, Evan and others},
  journal={arXiv preprint arXiv:2507.06261},
  year={2025}
}

@misc{gemini30pro,
  title={A new era of intelligence with {Gemini 3}},
  author={Pichai, Sundar and Hassabis, Demis and Kavukcuoglu, Koray},
  year={2025},
  howpublished={Google Blog, Nov 18, 2025. \url{https://blog.google/products-and-platforms/products/gemini/gemini-3/}}
}

@misc{gemini30flash,
  title={{Gemini 3 Flash}: frontier intelligence built for speed},
  author={Doshi, Tulsee},
  year={2025},
  howpublished={Google Blog, Dec 17, 2025. \url{https://blog.google/products-and-platforms/products/gemini/gemini-3-flash/}}
}

@misc{gemini31pro,
  title={{Gemini 3.1 Pro}: A smarter model for your most complex tasks},
  author={{Gemini Team, Google}},
  year={2026},
  howpublished={Google Blog, Feb 19, 2026. \url{https://blog.google/innovation-and-ai/models-and-research/gemini-models/gemini-3-1-pro/}}
}

@misc{claudes35,
  title={{Claude 3.5 Sonnet}},
  author={{Anthropic}},
  year={2024},
  howpublished={Anthropic Blog, Jun 21, 2024. \url{https://www.anthropic.com/news/claude-3-5-sonnet}}
}

@misc{claudes37,
  title={{Claude 3.7 Sonnet} and {Claude Code}},
  author={{Anthropic}},
  year={2025},
  howpublished={Anthropic Blog, Feb 24, 2025. \url{https://www.anthropic.com/news/claude-3-7-sonnet}}
}

@misc{claudeo4-claudes4,
  title={Introducing {Claude 4}},
  author={{Anthropic}},
  year={2025},
  howpublished={Anthropic Blog, May 22, 2025. \url{https://www.anthropic.com/news/claude-4}}
}

@misc{claudes45,
  title={Introducing {Claude Sonnet 4.5}},
  author={{Anthropic}},
  year={2025},
  howpublished={Anthropic Blog, Sep 29, 2025. \url{https://www.anthropic.com/news/claude-sonnet-4-5}}
}

@misc{claudeo45,
  title={Introducing {Claude Opus 4.5}},
  author={{Anthropic}},
  year={2025},
  howpublished={Anthropic Blog, Nov 24, 2025. \url{https://www.anthropic.com/news/claude-opus-4-5}}
}

@misc{claudes46,
  title={Introducing {Claude Sonnet 4.6}},
  author={{Anthropic}},
  year={2026},
  howpublished={Anthropic Blog, Feb 17, 2026. \url{https://www.anthropic.com/news/claude-sonnet-4-6}}
}

@misc{llama32,
  title={{Llama 3.2}: Revolutionizing edge {AI} and vision with open, customizable models},
  author={{Meta AI}},
  year={2024},
  howpublished={Meta AI Blog, Sep 25, 2024. \url{https://ai.meta.com/blog/llama-3-2-connect-2024-vision-edge-mobile-devices/}}
}

@article{seed15vl,
  title={{Seed1.5-VL} technical report},
  author={Guo, Dong and Wu, Faming and Zhu, Feida and Leng, Fuxing and Shi, Guang and Chen, Haobin and Fan, Haoqi and Wang, Jian and Jiang, Jianyu and Wang, Jiawei and others},
  journal={arXiv preprint arXiv:2505.07062},
  year={2025}
}

@misc{seed16,
  title={Introduction to techniques used in {Seed1.6}},
  author={{ByteDance Seed}},
  year={2025},
  howpublished={ByteDance Seed Blog, Jun 25, 2025. \url{https://seed.bytedance.com/en/blog/introduction-to-techniques-used-in-seed1-6}}
}

@article{seed20,
  title={{Seed2.0} model card: Towards intelligence frontier for real-world complexity},
  author={{ByteDance Seed}},
  journal={arXiv preprint arXiv:2607.00248},
  year={2026}
}

@misc{qwen35,
  title={{Qwen3.5}: Towards native multimodal agents},
  author={{Qwen Team}},
  year={2026},
  howpublished={Qwen Blog, Feb 16, 2026. \url{https://qwen.ai/blog?id=qwen3.5}}
}

@misc{qwen36,
  title={{Qwen3.6-Plus}: Towards real world agents},
  author={{Qwen Team}},
  year={2026},
  howpublished={Qwen Blog, Apr 1, 2026. \url{https://qwen.ai/blog?id=qwen3.6}}
}

@article{qwen2vl,
  title={Qwen2-vl: Enhancing vision-language model's perception of the world at any resolution},
  author={Wang, Peng and Bai, Shuai and Tan, Sinan and Wang, Shijie and Fan, Zhihao and Bai, Jinze and Chen, Keqin and Liu, Xuejing and Wang, Jialin and Ge, Wenbin and others},
  journal={arXiv preprint arXiv:2409.12191},
  year={2024}
}

@article{qwen25vl,
  title={Qwen2.5-vl technical report},
  author={Bai, Shuai and Chen, Keqin and Liu, Xuejing and Wang, Jialin and Ge, Wenbin and Song, Sibo and Dang, Kai and Wang, Peng and Wang, Shijie and Tang, Jun and others},
  journal={arXiv preprint arXiv:2502.13923},
  year={2025}
}

@article{qwen3vl,
  title={Qwen3-vl technical report},
  author={Bai, Shuai and Cai, Yuxuan and Chen, Ruizhe and Chen, Keqin and Chen, Xionghui and Cheng, Zesen and Deng, Lianghao and Ding, Wei and Gao, Chang and Ge, Chunjiang and others},
  journal={arXiv preprint arXiv:2511.21631},
  year={2025}
}

@misc{qwenvlmax-qwenvlplus,
  title={Introducing {Qwen-VL}},
  author={{Qwen Team}},
  year={2024},
  howpublished={Qwen Blog, Jan 25, 2024. \url{https://qwen.ai/blog?id=qwen-vl}}
}

@article{kimik25,
  title={{Kimi K2.5}: Visual agentic intelligence},
  author={Bai, Tongtong and Bai, Yifan and Bao, Yiping and Cai, SH and Cao, Yuan and Charles, Y and Che, HS and Chen, Cheng and Chen, Guanduo and others},
  journal={arXiv preprint arXiv:2602.02276},
  year={2026}
}

@misc{moonshotv1vision,
  title={Configure {Kimi} vision models ({moonshot-v1-8k/32k/128k-vision-preview})},
  author={{Moonshot AI}},
  year={2025},
  howpublished={Kimi API Platform documentation. Accessed 2026-08-29. \url{https://platform.moonshot.ai/docs/guide/use-kimi-vision-model}}
}

@misc{glm5v,
  title={{GLM-5V-Turbo}},
  author={{Z.ai}},
  year={2026},
  howpublished={Z.ai Developer Documentation. Accessed 2026-08-29. \url{https://docs.z.ai/guides/vlm/glm-5v-turbo}}
}

@article{hu2024visual,
  title={Visual sketchpad: Sketching as a visual chain of thought for multimodal language models},
  author={Hu, Yushi and Shi, Weijia and Fu, Xingyu and Roth, Dan and Ostendorf, Mari and Zettlemoyer, Luke and Smith, Noah A and Krishna, Ranjay},
  journal={Advances in Neural Information Processing Systems},
  volume={37},
  pages={139348--139379},
  year={2024}
}

@article{meng2023chain,
  title={Chain of Images for Intuitively Reasoning},
  author={Meng, Fanxu and Yang, Haotong and Wang, Yiding and Zhang, Muhan},
  journal={arXiv preprint arXiv:2311.09241},
  year={2023}
}

@article{shao2024visual,
  title={Visual cot: Advancing multi-modal language models with a comprehensive dataset and benchmark for chain-of-thought reasoning},
  author={Shao, Hao and Qian, Shengju and Xiao, Han and Song, Guanglu and Zong, Zhuofan and Wang, Letian and Liu, Yu and Li, Hongsheng},
  journal={Advances in Neural Information Processing Systems},
  volume={37},
  pages={8612--8642},
  year={2024}
}

@article{wu2024mind,
  title={Mind's eye of LLMs: visualization-of-thought elicits spatial reasoning in large language models},
  author={Wu, Wenshan and Mao, Shaoguang and Zhang, Yadong and Xia, Yan and Dong, Li and Cui, Lei and Wei, Furu},
  journal={Advances in Neural Information Processing Systems},
  volume={37},
  pages={90277--90317},
  year={2024}
}

@inproceedings{li2025imagine,
  title={Imagine while Reasoning in Space: Multimodal Visualization-of-Thought},
  author={Li, Chengzu and Wu, Wenshan and Zhang, Huanyu and Xia, Yan and Mao, Shaoguang and Dong, Li and Vuli\'{c}, Ivan and Wei, Furu},
  booktitle={Proceedings of the 42nd International Conference on Machine Learning},
  pages={36340--36364},
  volume={267},
  year={2025}
}

@article{liu2025spatialcot,
  title={SpatialCoT: Advancing Spatial Reasoning through Coordinate Alignment and Chain-of-Thought for Embodied Task Planning},
  author={Liu, Yuecheng and Chi, Dafeng and Wu, Shiguang and Zhang, Zhanguang and Hu, Yaochen and Zhang, Lingfeng and Zhang, Yingxue and Wu, Shuang and Cao, Tongtong and Huang, Guowei and others},
  journal={arXiv preprint arXiv:2501.10074},
  year={2025}
}

@inproceedings{cai2025spatialbot,
  title={Spatialbot: Precise spatial understanding with vision language models},
  author={Cai, Wenxiao and Ponomarenko, Iaroslav and Yuan, Jianhao and Li, Xiaoqi and Yang, Wankou and Dong, Hao and Zhao, Bo},
  booktitle={2025 IEEE International Conference on Robotics and Automation (ICRA)},
  pages={9490--9498},
  year={2025},
  organization={IEEE}
}

@inproceedings{duan2024vlmevalkit,
  title={Vlmevalkit: An open-source toolkit for evaluating large multi-modality models},
  author={Duan, Haodong and Yang, Junming and Qiao, Yuxuan and Fang, Xinyu and Chen, Lin and Liu, Yuan and Dong, Xiaoyi and Zang, Yuhang and Zhang, Pan and Wang, Jiaqi and others},
  booktitle={Proceedings of the 32nd ACM international conference on multimedia},
  pages={11198--11201},
  year={2024}
}

@article{chen2025ocean,
  title={Ocean-OCR: Towards General OCR Application via a Vision-Language Model},
  author={Chen, Song and Guo, Xinyu and Li, Yadong and Zhang, Tao and Lin, Mingan and Kuang, Dongdong and Zhang, Youwei and Ming, Lingfeng and Zhang, Fengyu and Wang, Yuran and others},
  journal={arXiv preprint arXiv:2501.15558},
  year={2025}
}

@article{yang2024mathglm,
  title={MathGLM-Vision: Solving Mathematical Problems with Multi-Modal Large Language Model},
  author={Yang, Zhen and Chen, Jinhao and Du, Zhengxiao and Yu, Wenmeng and Wang, Weihan and Hong, Wenyi and Jiang, Zhihuan and Xu, Bin and Dong, Yuxiao and Tang, Jie},
  journal={arXiv preprint arXiv:2409.13729},
  year={2024}
}

@article{peng2024multimath,
  title={Multimath: Bridging visual and mathematical reasoning for large language models},
  author={Peng, Shuai and Fu, Di and Gao, Liangcai and Zhong, Xiuqin and Fu, Hongguang and Tang, Zhi},
  journal={arXiv preprint arXiv:2409.00147},
  year={2024}
}

@article{azad2023foundational,
  title={Foundational models in medical imaging: A comprehensive survey and future vision},
  author={Azad, Bobby and Azad, Reza and Eskandari, Sania and Bozorgpour, Afshin and Kazerouni, Amirhossein and Rekik, Islem and Merhof, Dorit},
  journal={arXiv preprint arXiv:2310.18689},
  year={2023}
}

@article{buckley2026multimodal,
  title={Multimodal foundation models exploit text to make medical image predictions},
  author={Buckley, Thomas A and Diao, James A and Srivastava, Cam N and Brodeur, Peter G and Rajpurkar, Pranav and Rodman, Adam and Manrai, Arjun K},
  journal={Nature Communications},
  year={2026},
  publisher={Nature Publishing Group}
}

@article{wei2022chain,
  title={Chain-of-thought prompting elicits reasoning in large language models},
  author={Wei, Jason and Wang, Xuezhi and Schuurmans, Dale and Bosma, Maarten and Xia, Fei and Chi, Ed and Le, Quoc V and Zhou, Denny and others},
  journal={Advances in Neural Information Processing Systems},
  volume={35},
  pages={24824--24837},
  year={2022}
}

@inproceedings{sprague2025cot,
  title={To cot or not to cot? chain-of-thought helps mainly on math and symbolic reasoning},
  author={Sprague, Zayne and Yin, Fangcong and Rodriguez, Juan and Jiang, Dongwei and Wadhwa, Manya and Singhal, Prasann and Zhao, Xinyu and Ye, Xi and Mahowald, Kyle and Durrett, Greg},
  booktitle={International Conference on Learning Representations},
  volume={2025},
  pages={94118--94162},
  year={2025}
}

@inproceedings{liu2025mind,
  title={Mind Your Step (by Step): Chain-of-Thought can Reduce Performance on Tasks where Thinking Makes Humans Worse},
  author={Liu, Ryan and Geng, Jiayi and Wu, Addison J. and Sucholutsky, Ilia and Lombrozo, Tania and Griffiths, Thomas L.},
  booktitle={Proceedings of the 42nd International Conference on Machine Learning},
  pages={38489--38517},
  volume={267},
  year={2025}
}

@article{coda2023inducing,
  title={Inducing anxiety in large language models can induce bias},
  author={Coda-Forno, Julian and Witte, Kristin and Jagadish, Akshay K. and Binz, Marcel and Akata, Zeynep and Schulz, Eric},
  journal={arXiv preprint arXiv:2304.11111},
  year={2023}
}

@inproceedings{coda2024cogbench,
  title={CogBench: a large language model walks into a psychology lab},
  author={Coda-Forno, Julian and Binz, Marcel and Wang, Jane X and Schulz, Eric},
  booktitle={Proceedings of the 41st International Conference on Machine Learning},
  pages={9076--9108},
  volume={235},
  year={2024}
}

@article{huang2024apathetic,
  title={Apathetic or Empathetic? Evaluating {LLM}s' Emotional Alignments with Humans},
  author={Huang, Jen-tse and Lam, Man Ho and Li, Eric John and Ren, Shujie and Wang, Wenxuan and Jiao, Wenxiang and Tu, Zhaopeng and Lyu, Michael R},
  journal={Advances in Neural Information Processing Systems},
  volume={37},
  year={2024}
}

@inproceedings{huang2024reliability,
  title={On the Reliability of Psychological Scales on Large Language Models},
  author={Huang, Jen-tse and Jiao, Wenxiang and Lam, Man Ho and Li, Eric John and Wang, Wenxuan and Lyu, Michael R.},
  booktitle={Proceedings of The 2024 Conference on Empirical Methods in Natural Language Processing},
  year={2024}
}

@inproceedings{liu2024interintent,
  title={InterIntent: Investigating Social Intelligence of LLMs via Intention Understanding in an Interactive Game Context},
  author={Liu, Ziyi and Anand, Abhishek and Zhou, Pei and Huang, Jen-tse and Zhao, Jieyu},
  booktitle={Proceedings of the 2024 Conference on Empirical Methods in Natural Language Processing},
  year={2024}
}

@article{liang2023leveraging,
  title={Leveraging Word Guessing Games to Assess the Intelligence of Large Language Models},
  author={Liang, Tian and He, Zhiwei and Huang, Jen-tse and Wang, Wenxuan and Jiao, Wenxiang and Wang, Rui and Yang, Yujiu and Tu, Zhaopeng and Shi, Shuming and Wang, Xing},
  journal={arXiv preprint arXiv:2310.20499},
  year={2023}
}

@inproceedings{huang2025competing,
  title={Competing large language models in multi-agent gaming environments},
  author={Huang, Jen-tse and Li, Eric John and Lam, Man Ho and Liang, Tian and Wang, Wenxuan and Yuan, Youliang and Jiao, Wenxiang and Wang, Xing and Tu, Zhaopeng and Lyu, Michael},
  booktitle={International Conference on Learning Representations},
  volume={2025},
  pages={65639--65681},
  year={2025}
}

@inproceedings{wang2024incharacter,
  title={Incharacter: Evaluating personality fidelity in role-playing agents through psychological interviews},
  author={Wang, Xintao and Xiao, Yunze and Huang, Jen-tse and Yuan, Siyu and Xu, Rui and Guo, Haoran and Tu, Quan and Fei, Yaying and Leng, Ziang and Wang, Wei and others},
  booktitle={Proceedings of the 62nd Annual Meeting of the Association for Computational Linguistics (Volume 1: Long Papers)},
  pages={1840--1873},
  year={2024}
}

@inproceedings{wang2025coser,
  title={CoSER: Coordinating LLM-Based Persona Simulation of Established Roles.},
  author={Wang, Xintao and Wang, Heng and Zhang, Yifei and Yuan, Xinfeng and Xu, Rui and Huang, Jen-tse and Yuan, Siyu and Guo, Haoran and Chen, Jiangjie and Zhou, Shuchang and others},
  booktitle={Proceedings of the 42nd International Conference on Machine Learning},
  page={64822--64858},
  volume={267},
  year={2025}
}

@article{ng2024well,
  title={How Well Can LLMs Echo Us? Evaluating AI Chatbots' Role-Play Ability with ECHO},
  author={Ng, Man Tik and Tse, Hui Tung and Huang, Jen-tse and Li, Jingjing and Wang, Wenxuan and Lyu, Michael R},
  journal={arXiv preprint arXiv:2404.13957},
  year={2024}
}

@article{schooler1990verbal,
  title={Verbal overshadowing of visual memories: Some things are better left unsaid},
  author={Schooler, Jonathan W and Engstler-Schooler, Tonya Y},
  journal={Cognitive psychology},
  volume={22},
  number={1},
  pages={36--71},
  year={1990},
  publisher={Elsevier}
}

@article{dijksterhuis2004think,
  title={Think different: the merits of unconscious thought in preference development and decision making.},
  author={Dijksterhuis, Ap},
  journal={Journal of personality and social psychology},
  volume={87},
  number={5},
  pages={586},
  year={2004},
  publisher={American Psychological Association}
}

@article{van2008intentional,
  title={Intentional artificial grammar learning: When does it work?},
  author={Van den Bos, Esther and Poletiek, Fenna H},
  journal={European Journal of Cognitive Psychology},
  volume={20},
  number={4},
  pages={793--806},
  year={2008},
  publisher={Taylor \& Francis}
}
\bibliographystyle{mybib}

\addtocontents{toc}{\protect\setcounter{tocdepth}{2}}
\clearpage
\appendix

\tableofcontents

\section{{\methodname} dataset details}

\subsection{Statistics}
\label{sec:dataset}

\begin{figure}[h]
    \centering
    \includegraphics[width=0.86\linewidth]{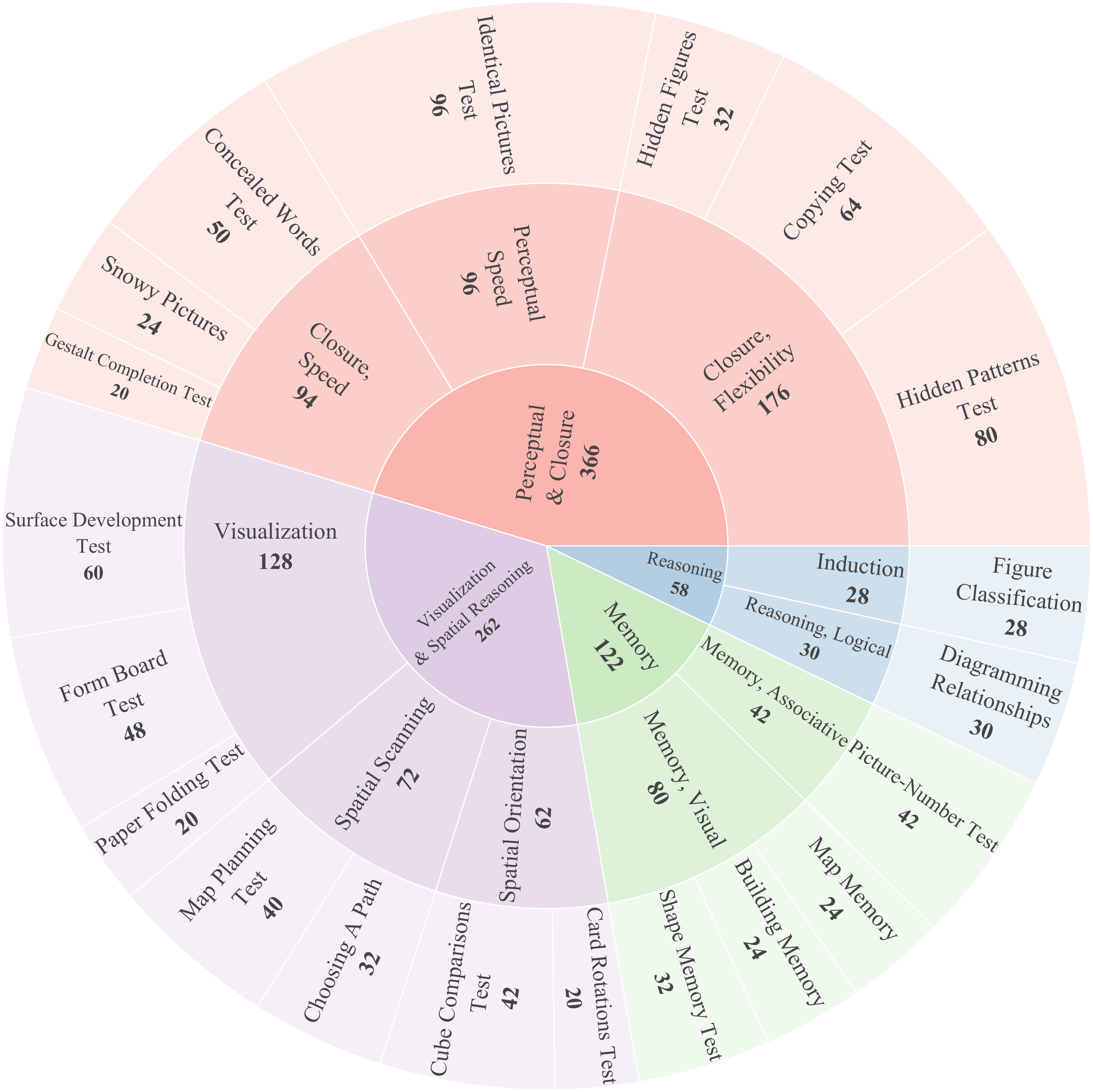}
    \caption{{\methodname} integrates 20 subtests across 10 factors from four domains.}
    \label{fig:overview}
\end{figure}

\clearpage

\begin{table}[h]
    \centering
    \caption{{\methodname} includes 3,046 queries (808 questions), providing sufficient statistical power.}
    \label{tab:statistics}
    \resizebox{0.86\linewidth}{!}{
    \begin{tabular}{llcccc}
    \toprule
    \bf Category & \bf Name & \bf ID & \bf \#Questions & \bf \#Queries & \bf Guess Acc. \\
    \midrule
    \multirow{6}{*}{Perceptual} & Hidden Figures Test & CF1 & 32 & 160 & 3.13\% \\
    \multirow{6}{*}{\& Closure} & Hidden Patterns Test & CF2 & 80 & 400 & 3.13\% \\
    & Copying Test & CF3 & 64 & 64 & 4.00\% \\
    & Gestalt Completion Test & CS1 & 20 & 20 & 0.00\% \\
    & Concealed Words Test & CS2 & 50 & 50 & 0.00\% \\
    & Snowy Pictures & CS3 & 24 & 24 & 0.00\% \\
    & Identical Pictures Test & P3 & 96 & 480 & 3.13\% \\
    \midrule
    \multirow{2}{*}{Reasoning} & Figure Classification & I3 & 28 & 224 & 0.23\% \\
    & Diagramming Relationships & RL2 & 30 & 150 & 3.13\% \\
    \midrule
    \multirow{4}{*}{Memory} & Picture-Number Test & MA1 & 42 & 42 & 4.76\% \\
    & Shape Memory Test & MV1 & 32 & 128 & 6.25\% \\
    & Building Memory & MV2 & 24 & 120 & 3.13\% \\
    & Map Memory & MV3 & 24 & 96 & 6.25\% \\
    \midrule
    & Card Rotations Test & S1 & 20 & 160 & 0.39\% \\
    & Cube Comparisons Test & S2 & 42 & 168 & 6.25\% \\
    Visualization & Choosing A Path & SS2 & 32 & 160 & 3.13\% \\
    \& Spatial & Map Planning Test & SS3 & 40 & 80 & 1.00\% \\
    Reasoning & Form Board Test & VZ1 & 48 & 240 & 3.13\% \\
    & Paper Folding Test & VZ2 & 20 & 100 & 3.13\% \\
    & Surface Development Test & VZ3 & 60 & 180 & 3.65\% \\
    \midrule
    & & All & 808 & 3,046 & 2.89\% \\
    \bottomrule
    \end{tabular}
    }
\end{table}

\subsection{Instructions for models}

The original FRCT prompts are verbose and full of human-oriented instructions, which are sometimes unnecessary and hard for older models with weaker instruction-following.
Here, we show an example of the original instruction for the VZ1 (Form Board Test):

\begin{quote}
    \textit{This is a test of your ability to tell what pieces can be put together to make a certain figure. Each test page is divided into two columns. At the top of each column is a geometrical figure. Beneath each figure are several problems. Each problem consists of a row of five shaded pieces. Your task is to decide which of the five shaded pieces will make the complete figure when put together. Any number of shaded pieces, from two to five, may be used to make the complete figure. Each piece may be turned around to any position but it cannot be turned over. It may help you to sketch the way the pieces fit together. You may use any blank space for doing this. When you know which pieces make the complete figure, mark a plus (+) in the box under ones that are used and a minus (-) in the box under ones that are not used.<IMAGE\_0><IMAGE\_1>}
\end{quote}

We pasted the originals to GPT-4o and Gemini-2.5 and asked them to output the minimal, essential instructions, which a human annotator then reconciled against the original manual.
The goal of the rewrite is precisely to remove construct-irrelevant instruction-following difficulty.
Below is an example of the rewritten instruction for the VZ1 shown above:

\begin{quote}
    \textit{Look at the two images. Below is the first image, which is the figure you must make:<IMAGE\_0>Below is the second image, which are the five pieces you can use:<IMAGE\_1>Rules:<br>1. Use 2–5 of the pieces to fill the figure exactly.<br>2. You may rotate pieces but do not flip them.<br><br>Task: Decide whether the <ADDITIONAL\_0> piece is in the set of pieces that makes the figure.<br><br>Output: Respond with only one word: “TRUE” if it is or “FALSE” if it is not, in JSON format as follows: {"answer": YOUR\_ANSWER\_HERE}.}
\end{quote}

\clearpage

\subsection{Evaluation}
\label{sec:variants}

To prevent inflated scores from lucky guesses, we modify test formats as follows, except CF3 (25‑way), MA1 (21‑way) and all fill‑in‑the‑blank subtests (CS1--CS3) that already exhibit $\le 5\%$ random success.
The average random guessing performance is reduced from $22.47\%$ to $2.89\%$, with no single test exceeding $6.25\%$.

\begin{enumerate}[leftmargin=*]

\item \textbf{Decomposed multiple choice}:
For seven subtests with five options (CF1, MV2, P3, RL2, SS2, VZ1, VZ2), we pose \emph{one yes/no query per option} and require the model to answer \emph{all} correctly for credit.
Chance accuracy thus drops from 20\% to $(0.5)^5 \approx 3.13\%$.

\item \textbf{Grouped‑consistency items}:
These subtests repeatedly probe the same latent feature across a small set of items.
We aggregate each cluster and award credit only if \emph{all} constituent items are correct.
This applies to:
\textit{(i)} CF2 Hidden Patterns Test: 400 binary items grouped into 80 sets of five; chance $(0.5)^5 \approx 3.13\%$.
\textit{(ii)} I3 Figure Classification: 8 figures to be classified into two or three groups; chance $\approx 0.23\%$.
\textit{(iii)} S1 Card Rotation Test: 8 judgments of the same card; chance $(0.5)^8 \approx 0.39\%$.

\item \textbf{Symmetry variants}:
MV1, MV3 and S2 originally ask whether figure~A matches figure~B.
We generate three more variants per item—``A differs from B'', ``B matches A'', ``B differs from A''—so that ``yes'' and ``no'' answers are balanced, preventing easy success by models that consistently answer yes or no.
The probability of guessing all four correctly by chance is $(0.5)^4 = 6.25\%$.

\item \textbf{Specialized rewrites}:
\textit{(i)} SS3 (Map Planning Test).
Each item asks participants to find the building number that the shortest path between a \emph{start} and an \emph{end} point passes in a map.
Exchanging start and end leaves the correct answer unchanged.
We therefore require the model to answer \emph{both} directions correctly, lowering chance from 10\% to 1\%.
\textit{(ii)} VZ3 (Surface Development Test).
Each item asks: which 3-D edge corresponds to the marked 2-D edge after folding?
Since multiple 2-D edges may map to the same 3-D edge, simply swapping the query direction (asking which 2-D edge matches a given 3-D edge) would introduce one-to-many ambiguity and ill-defined ground truth.
Therefore, we add additional questions asking whether a pair of 2-D and 3-D edges are the same, resulting in all ``yes'' ground truth.
To create ``no'' pairs, we generate questions with cyclic-permuted 3-D edge labels (\eg, $A \to B \to C \to D \to E \to A$).
MLLMs receive credit only if they correctly answer the fill-in-the-blank question and both yes/no questions.
The fill-in-the-blank question is not open-ended: the answer must be one of the lettered edges of the 3-D object in that item, and the number of candidate edges $N_i$ varies across items.
Guessing uniformly among them succeeds with probability $14.6\%$, so the chance level of a full VZ3 item is $14.6\% \times (0.5)^2 = 3.65\%$.

\end{enumerate}

We then adopt a unified and fully specified evaluation protocol across all task types.
For yes-no questions, model outputs are normalized and matched to the gold label using the sets \texttt{\{t, y, 1, true, yes\}} for the True class and \texttt{\{f, n, 0, false, no\}} for the False class.
Multiple-choice questions are evaluated by directly matching the predicted option letter (A/B/C/D).
For numeric fill-in-the-blank problems, we require exact matching.
For CS1 and CS3, multiple acceptable ground-truth answers are provided; a prediction is counted as correct if its normalized form matches any valid answer variant.
CS2 uses strict exact-match evaluation.
The total score is the average of all 20 tests.

\clearpage

\section{Further analysis}

\subsection{Performance over time}

\begin{figure}[h]
    \centering
    \includegraphics[width=0.86\linewidth]{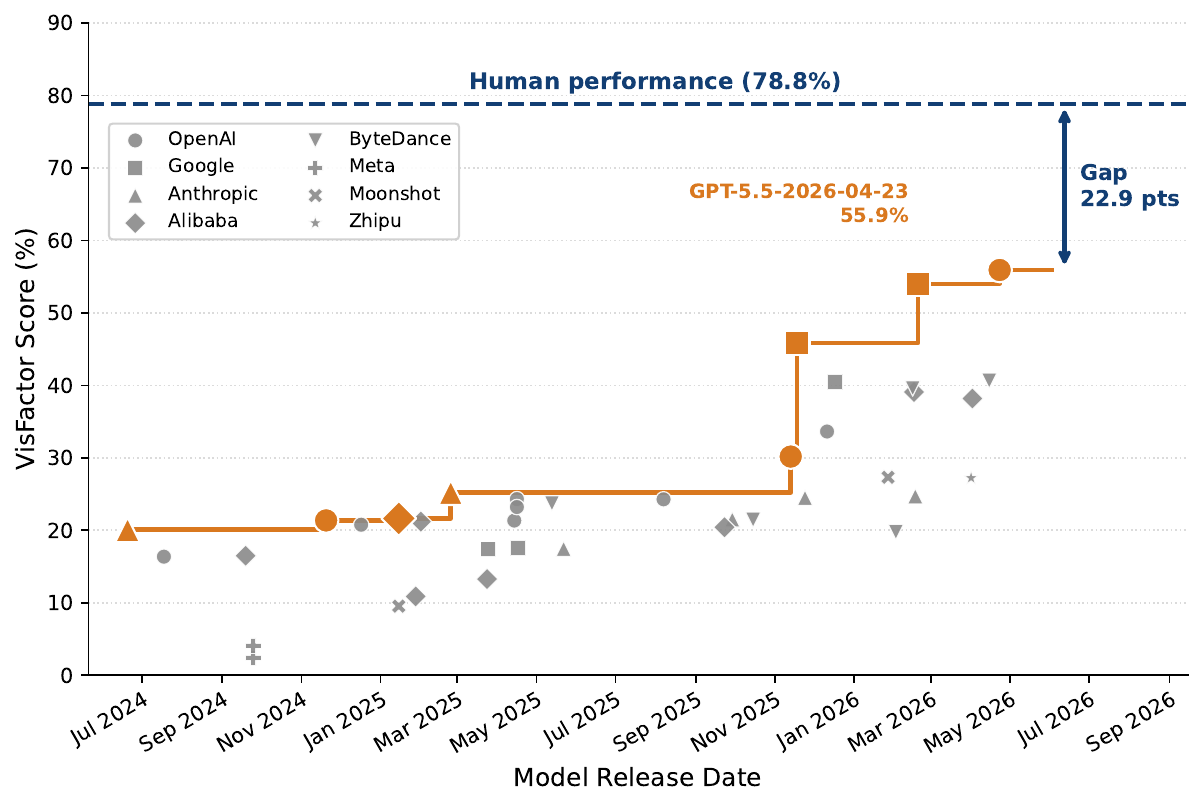}
    \caption{\textbf{Performance vs. release date.} Plotting each model's overall score against its release date shows scores plateauing near 25 through most of 2025, then rising sharply from Gemini-3.0 onward to exceed 40, with GPT-5.5 peaking at 55.9.}
    \label{fig:timeline}
\end{figure}

\subsection{The ``Middle Score Anomaly''}

\begin{figure}[h]
    \centering
    \includegraphics[width=0.86\linewidth]{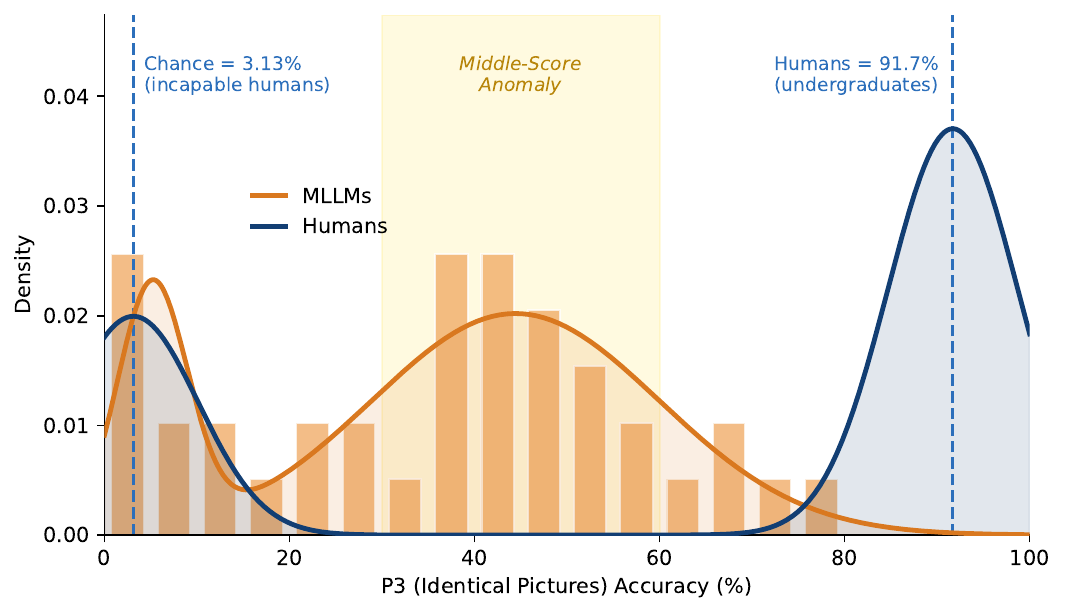}
    \caption{\textbf{The ``Middle Score Anomaly'' on the P3 (Identical Pictures Test).} We plot the distribution of model scores on P3, along with the chance level (3.13) and human performance (91.7). Model scores cluster in the 30--60 band, far from both chance and the human near-ceiling.}
    \label{fig:middle-score-anomaly}
\end{figure}

\clearpage

\subsection{Ablation on temperature}

\begin{table*}[h]
    \centering
    \caption{The performance of GPT-4.1-2025-04-14, GPT-4o-2024-11-20, and GPT-4o-Mini-2024-07-18 in {\methodname} using different temperatures of \texttt{\{0.0, 0.5, 1.0\}}.}
    \label{tab:temperature}
    \includegraphics[width=1.0\linewidth]{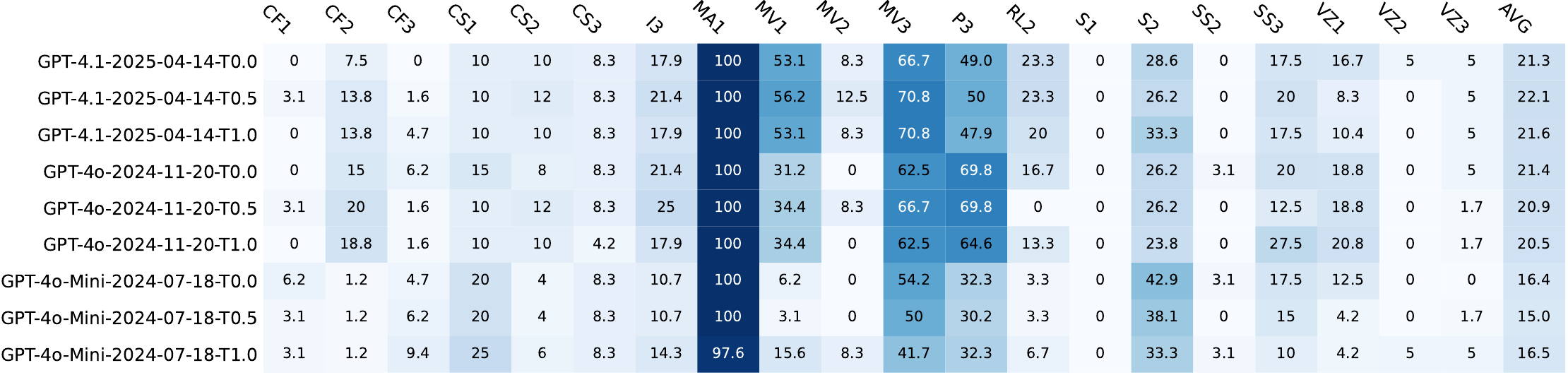}
\end{table*}

\subsection{Ablation on reasoning effort}

\begin{table*}[h]
    \centering
    \caption{The performance of GPT-4.1-2025-04-14, GPT-4o-2024-11-20, and GPT-4o-Mini-2024-07-18 with and without CoT. We also test GPT-5.1-2025-11-13 with different levels of reasoning efforts.}
    \label{tab:reasoning}
    \includegraphics[width=1.0\linewidth]{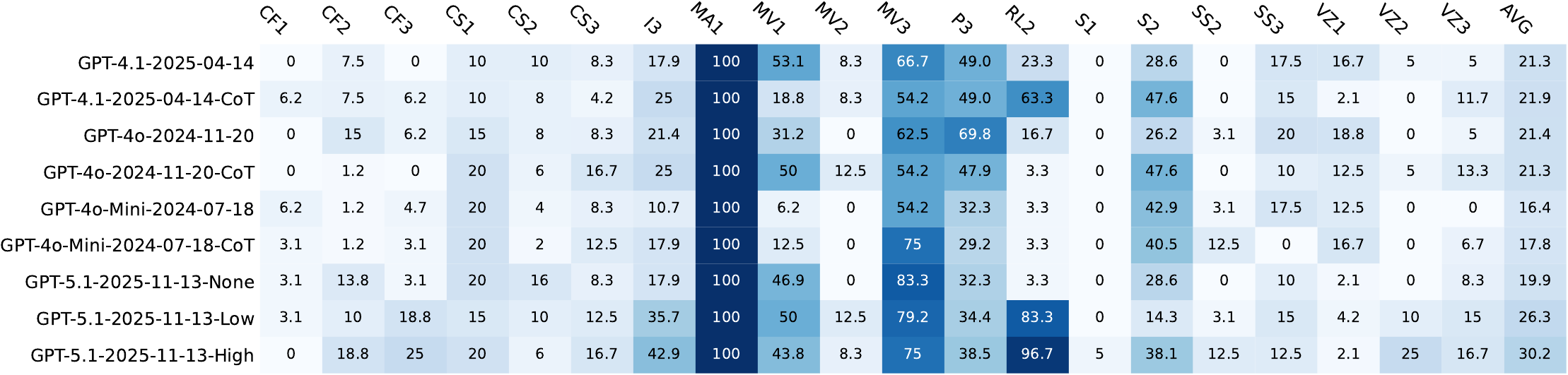}
\end{table*}

\subsection{Construct validity}

\begin{figure*}[h]
    \centering
    \includegraphics[width=1.0\linewidth]{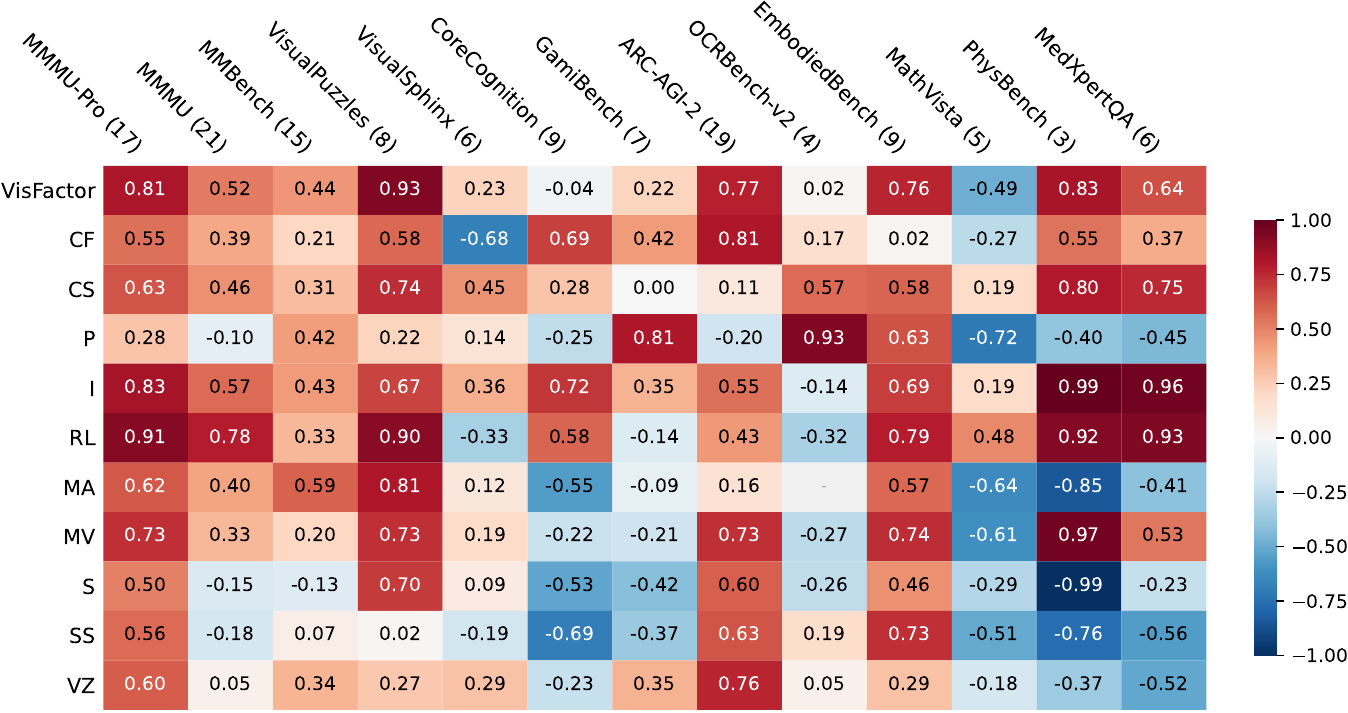}
    \caption{Pearson correlation between {\methodname} (10 factors) and existing benchmarks. The numbers after each benchmark in the parentheses represent the sample size. Note that some of them are relatively small due to availability in the original papers.}
    \label{fig:correlation-factor}
\end{figure*}

\clearpage

\begin{figure*}[h]
    \centering
    \includegraphics[width=1.0\linewidth]{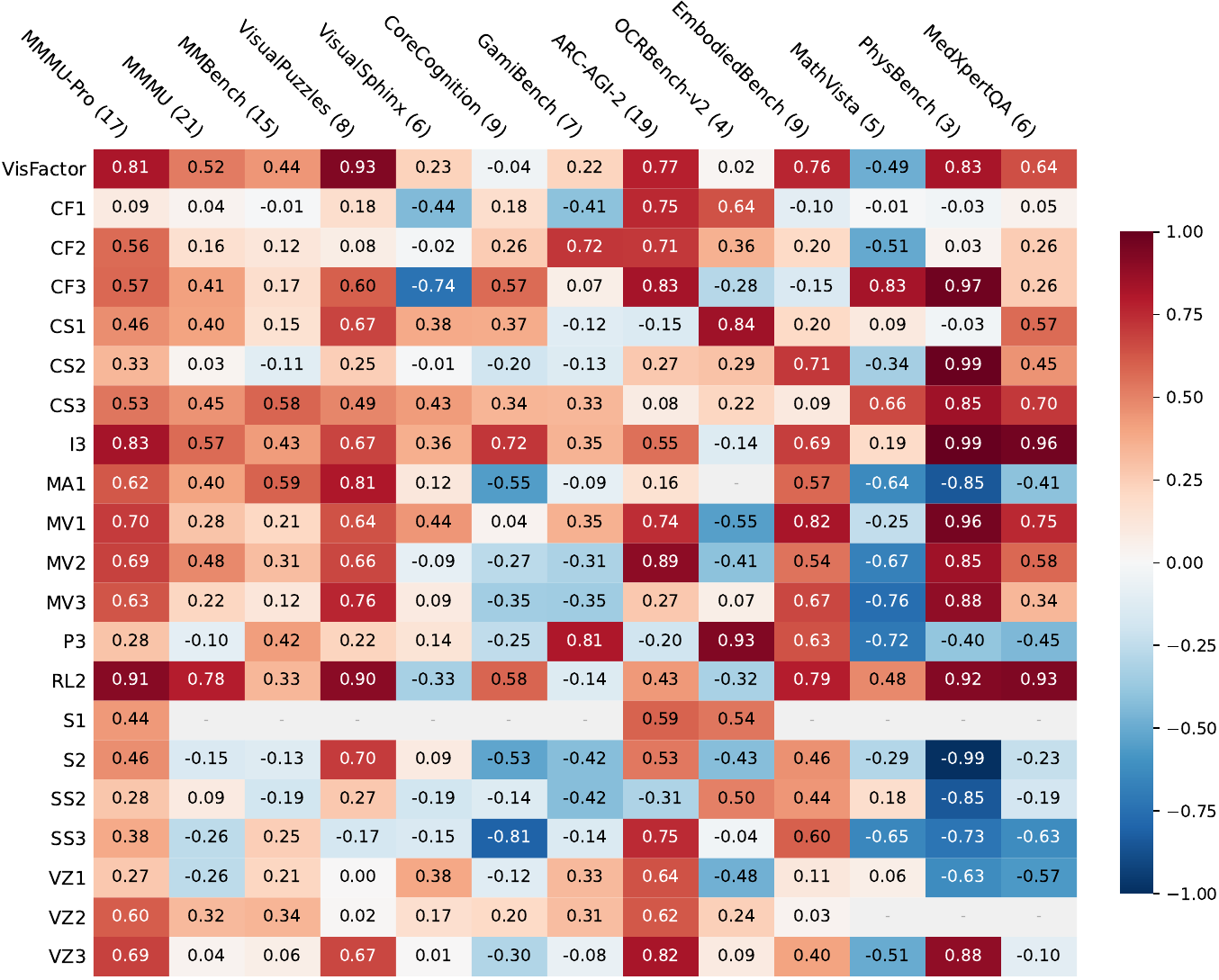}
    \caption{Pearson correlation between {\methodname} (20 subtests) and existing benchmarks. The numbers after each benchmark in the parentheses represent the sample size. Note that some of them are relatively small due to availability in the original papers.}
    \label{fig:correlation-subtest}
\end{figure*}

We conduct a correlation analysis across 13 established vision-language benchmarks, including general vision ability~\cite{yue2024mmmu, yue2025mmmu, liu2024mmbench}, math~\cite{lu2024mathvista}, physics~\cite{chow2025physbench}, medical~\cite{zuo2025medxpertqa}, character recognition~\cite{fu2025ocrbench}, embodied exploration~\cite{yang2025embodiedbench}, and benchmarks similar to ours~\cite{song2026visualpuzzles, feng2025visualsphinx, li2025core, spencer2025gamibench, chollet2025arc}.
{\methodname} demonstrates strong convergent validity, evidenced by its high positive correlation with comprehensive multi-modal benchmarks like MMMU-Pro ($r = 0.81$) and structurally similar visual reasoning tasks such as VisualPuzzles ($r = 0.93$).
This confirms that {\methodname} successfully captures foundational visual intelligence.
Conversely, the benchmark exhibits excellent discriminant validity, yielding near-zero or negative correlations with orthogonal domains such as OCRBench-v2 ($r = 0.02$) and MathVista ($r = -0.49$).
Crucially, {\methodname} maintains modest or low correlations with other matrix-style and Bongard-like reasoning tasks, including VisualSphinx ($r = 0.23$) and GamiBench ($r = 0.22$), highlighting its unique, non-redundant contribution to the evaluation landscape.

At the fine-grained factor level (Figure~\ref{fig:correlation-factor}), our taxonomy successfully isolates distinct cognitive capabilities.
For instance, Factor P strongly aligns with spatial and transformational reasoning benchmarks (GamiBench, $r = 0.81$; OCRBench-v2, $r = 0.93$), despite the overall benchmark's low correlation with them.
Furthermore, a granular sub-task analysis (Figure~\ref{fig:correlation-subtest}) reveals critical nuances in model behaviors and structural biases.
Sub-tasks such as RL2 and I3 exhibit strong, broad correlations with comprehensive benchmarks (e.g., RL2 with MMMU-Pro at $r = 0.91$; I3 with MedXpertQA at $r = 0.96$).
These fine-grained distinctions confirm that {\methodname}'s sub-tasks are not merely homogeneous difficulty variants, but rather distinct diagnostic probes capable of isolating specific architectural or pre-training vulnerabilities in MLLMs.

\clearpage

\subsection{Internal reliability}

\begin{figure*}[h]
    \centering
    \includegraphics[width=1.0\linewidth]{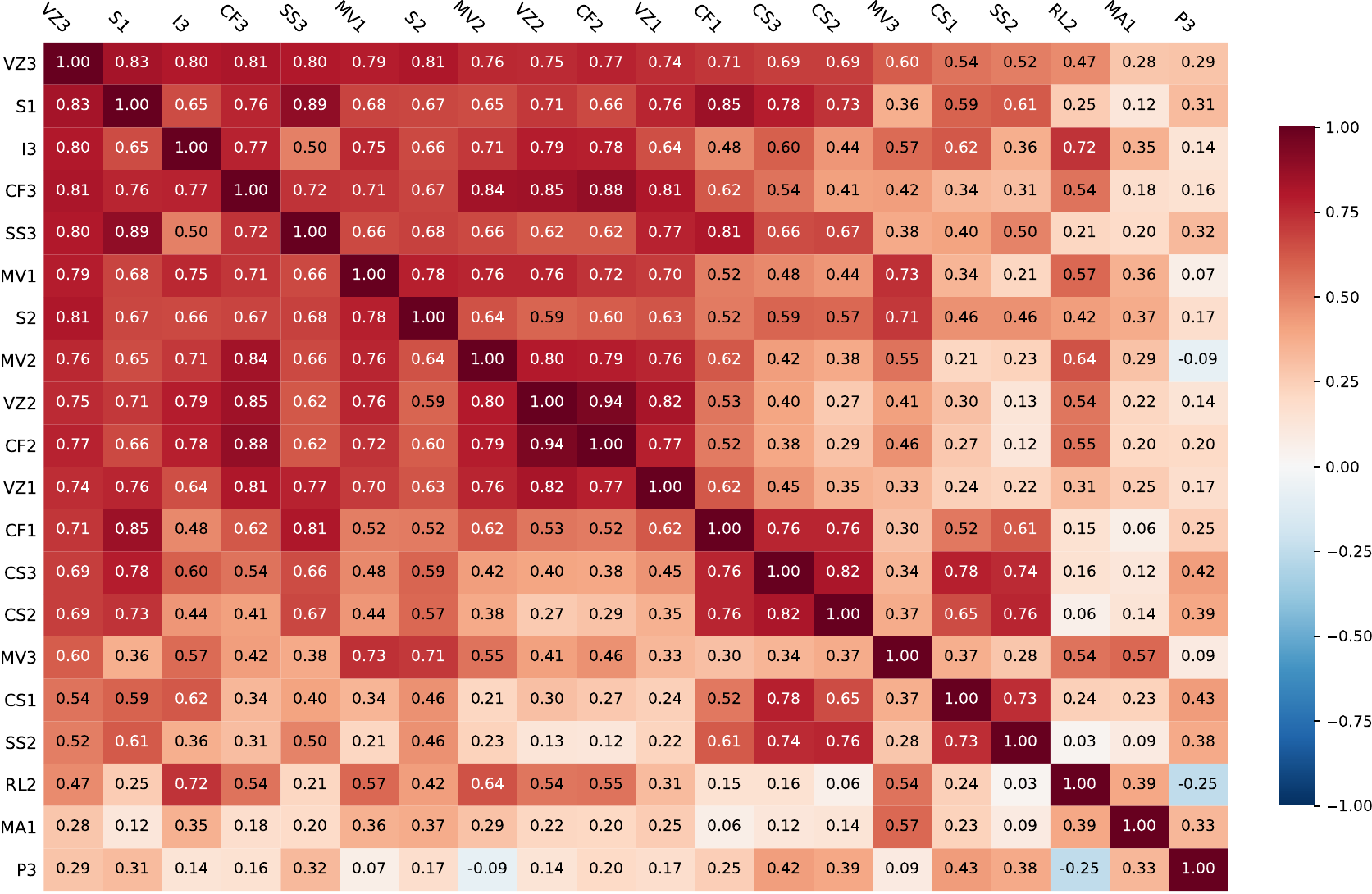}
    \caption{Pearson correlation between all subtests in {\methodname}.}
    \label{fig:correlation}
\end{figure*}

As illustrated in the inter-task Pearson correlation matrix (Figure~\ref{fig:correlation}), the majority of subtests exhibit moderate to high positive correlations ($r \in [0.4, 0.8]$).
This demonstrates strong internal consistency, suggesting that these tasks collectively evaluate a shared, underlying general visual-language capability.
However, subtests MA1, P3, and RL2 exhibit distinct correlational deviations, highlighting the discriminant validity of our benchmark.
Specifically, MA1 demonstrates near-zero correlation with the rest of the suite; this is driven by a strict ceiling effect, as nearly all evaluated models achieved perfect scores, effectively eliminating the statistical variance required to compute meaningful correlations.
Similarly, P3 acts as an outlier due to its fundamentally elementary nature.
As the easiest test in the benchmark (simple image equivalence matching), its cognitive demand diverges sharply from the complex reasoning required elsewhere.
Notably, P3 exhibits a negative correlation ($r = -0.25$) with RL2.
This contrast aligns closely with their architectural demands: whereas P3 relies on basic, low-level visual pattern matching, RL2 is uniquely related to text understanding—a linguistic modality de-emphasized in the remaining vision-centric subtests.
These numbers indicate that a few subtests such as MA1 or RL2 may additionally recruit non-perceptual capabilities instead of the very fundamental visual capabilities.
In contrast, VZ is in fact the most coherent factor (within mean $r = 0.768$, the highest of all).
Even the weakest VZ1--VZ3 pair ($0.74$) ranks 40th of 190 pairs (range $-0.25$ to $0.94$).

Furthermore, the overall reliability of the benchmark is strong.
Across the $N = 40$ evaluated subjects (comprising {\nummodel} distinct models and one human undergraduate baseline), the test battery achieved a Cronbach's $\alpha$ coefficient of $0.9413$, indicating excellent internal consistency ($\alpha > 0.90$).
In addition, within‑factor mean correlation ($0.698$) is substantially higher than the between‑factor one ($0.487$), and this holds for every multi‑subtest factor.
After mean‑centering each model's scores to remove the dominant ``overall capability'' axis, between‑factor correlations drop to $-0.06$ while within‑factor stays positive ($0.232$).
These metrics confirm that the core suite of our subtests functions as a highly cohesive and reliable instrument.
The results collectively establish that our benchmark provides a stable, unified measurement of general visual-language reasoning, while successfully capturing specific architectural vulnerabilities through targeted outliers.

\clearpage

\subsection{Grouped accuracy on I3 and S1}

\begin{table*}[h]
\centering
\caption{Additional results on {\nummodel} models.}
\subfloat[Per-item (ungrouped) accuracy on S1 and I3.]{
\label{tab:ungrouped}
\begin{tabular}{lcc}
\toprule
Model & S1 & I3 \\
\midrule
Baseline & 50.00 & 43.00 \\
\midrule
Claude-Opus-4.5 & 53.12 & 71.88 \\
Claude-Sonnet-3.5 & 44.37 & 62.05 \\
Claude-Sonnet-3.7 & 51.25 & 60.27 \\
Claude-Sonnet-4.5 & 59.38 & 60.27 \\
Claude-Sonnet-4.6 & 56.87 & 73.66 \\
Claude-Sonnet-4 & 56.87 & 63.39 \\
GLM-5V-Turbo & 60.62 & 76.79 \\
GPT-4.1-2025-04-14 & 48.75 & 68.75 \\
GPT-4o-2024-11-20 & 53.75 & 66.96 \\
GPT-4o-Mini-2024-07-18 & 51.88 & 67.41 \\
GPT-5-Mini-2025-08-07 & 50.62 & 72.77 \\
GPT-5.1-2025-11-13 & 55.00 & 76.34 \\
GPT-5.2-2025-12-11 & 61.25 & 80.80 \\
GPT-5.5-2026-04-23 & 88.12 & 82.14 \\
Gemini-2.5-Flash & 45.62 & 46.43 \\
Gemini-2.5-Pro & 59.38 & 75.89 \\
Gemini-3.0-Flash & 76.88 & 87.95 \\
Gemini-3.0-Pro & 73.75 & 83.93 \\
Gemini-3.1-Pro & 81.87 & 92.86 \\
Kimi-K2.5 & 56.25 & 74.11 \\
LLaMA-3.2-11B-Vision & 50.62 & 41.52 \\
LLaMA-3.2-90B-Vision & 42.50 & 39.73 \\
Moonshot-V1-128K-Vision & 46.88 & 54.02 \\
Qwen-2-VL-72B & 46.25 & 57.59 \\
Qwen-2.5-VL-32B & 51.25 & 51.79 \\
Qwen-2.5-VL-72B & 49.38 & 55.36 \\
Qwen-3-VL-235B-A22B & 59.38 & 70.09 \\
Qwen-3.5-Plus & 80.00 & 80.80 \\
Qwen-3.6-Plus & 76.25 & 82.59 \\
Qwen-VL-Max & 53.75 & 65.18 \\
Qwen-VL-Plus & 57.50 & 63.39 \\
Seed-1.5-VL & 53.12 & 60.27 \\
Seed-1.6-Thinking & 51.25 & 66.07 \\
Seed-2.0-Lite & 68.75 & 81.70 \\
Seed-2.0-Mini & 50.62 & 67.86 \\
Seed-2.0-Pro & 64.38 & 80.36 \\
o1-2024-12-17 & 51.88 & 77.23 \\
o3-2025-04-16 & 58.13 & 77.68 \\
o4-Mini-2025-04-16 & 51.25 & 68.30 \\
\bottomrule
\end{tabular}
}
\hfill
\subfloat[Self-contradiction rate on MV1, MV3, S2.]{
\label{tab:contradiction}
\begin{tabular}{cccc}
\toprule
MV1 & MV3 & S2 & Avg \\
\midrule
\\
\midrule
23.44 & 12.50 & 39.29 & 25.08 \\
27.87 & 8.51 & 3.61 & 13.33 \\
21.88 & 2.08 & 17.86 & 13.94 \\
40.62 & 35.42 & 27.38 & 34.47 \\
31.25 & 14.58 & 28.92 & 24.92 \\
41.27 & 33.33 & 29.76 & 34.79 \\
31.25 & 6.25 & 29.76 & 22.42 \\
4.69 & 10.42 & 21.43 & 12.18 \\
25.00 & 18.75 & 39.29 & 27.68 \\
56.25 & 14.58 & 13.10 & 27.98 \\
46.88 & 12.50 & 26.92 & 28.77 \\
28.12 & 8.33 & 13.10 & 16.52 \\
6.25 & 6.25 & 4.76 & 5.75 \\
9.38 & 8.33 & 4.76 & 7.49 \\
36.51 & 9.09 & 36.59 & 27.40 \\
87.50 & 89.58 & 84.52 & 87.20 \\
26.56 & 8.33 & 26.51 & 20.47 \\
28.12 & 4.17 & 16.67 & 16.32 \\
6.25 & 0.00 & 16.67 & 7.64 \\
18.75 & 8.33 & 16.67 & 14.58 \\
57.89 & 50.00 & 43.04 & 50.31 \\
38.18 & 52.27 & 11.11 & 33.85 \\
20.51 & 41.67 & 40.32 & 34.17 \\
48.44 & 31.25 & 40.48 & 40.06 \\
45.31 & 31.25 & 33.33 & 36.63 \\
100.00 & 100.00 & 100.00 & 100.00 \\
45.31 & 18.75 & 34.52 & 32.86 \\
15.62 & 6.25 & 26.19 & 16.02 \\
10.94 & 12.50 & 28.57 & 17.34 \\
26.56 & 25.00 & 2.38 & 17.98 \\
32.81 & 12.50 & 25.00 & 23.44 \\
7.81 & 12.50 & 3.57 & 7.96 \\
33.87 & 14.58 & 42.86 & 30.44 \\
26.56 & 18.75 & 27.38 & 24.23 \\
48.44 & 10.42 & 52.38 & 37.08 \\
18.75 & 4.17 & 19.05 & 13.99 \\
15.62 & 8.33 & 34.52 & 19.49 \\
40.62 & 18.75 & 26.19 & 28.52 \\
29.69 & 10.42 & 34.52 & 24.88 \\
\bottomrule
\end{tabular}
}
\end{table*}

We observed that current models perform poorly under the metric that groups eight questions together.
Therefore, we additionally computed per-item (ungrouped) accuracy for I3 and S1 in Table~\ref{tab:ungrouped}.
First, the two metrics are highly consistent; the models that perform well using the grouped metric are exactly those who dominate at the per-item level: GPT-5.5 (88.1 on S1, 82.1 on I3), Gemini-3.1-Pro (81.9, 92.9), and Qwen-3.5 (80.0, 80.8).
Weaker models sit near their respective chance baselines (50 for S1, 43 for I3), \eg, LLaMA-90B at 42.5/39.7 and Gemini-2.5-Flash at 45.6/46.4.
Second, the near-zero grouped scores of weaker models are the expected consequence of per-item chance behavior, not an artifact of an unfair metric.
GPT-5.5's achieves 88.12\% at the the per-item level on S1.
If the eight questions that we group probe separable questions instead of a single coherent latent ability, the grouped score will be around $0.8812^8=0.3635$.
However, GPT-5.5 gets 50\% on S1 using the grouped metric, exceeding the predicted 36.4\%.

\subsection{Self-consistency evaluation for MV1, MV3, and S2}

To probe whether failures stem from inconsistency or lack of visual grounding rather than mere uncertainty, we added a self-consistency (contradiction) analysis on three Yes-No subtests (MV1, MV3, S2).
Each item is run as four logically linked variants: (1) the original statement, (2) its negation, (3) the original with the two objects swapped, and (4) the negation of (3).
Pairs (1,2) and (3,4) are logical negations, so a grounded model must answer them oppositely.
The contradiction rate is the fraction of negation pairs answered with the same label: 0\% means self-consistent, 50\% indicates random guessing, 100\% shows degenerate constant responder (always the same label regardless of the image).
The 100\% case subsumes the ``answers yes to all'' or ``rejects all'' patterns.

Three findings emerge from Table~\ref{tab:contradiction}.
(1) Gains are genuine, not lucky guesses. The two top-performing models, GPT-5.5 and Gemini-3.1-Pro, are also among the four most self-consistent (all $<10\%$ contradiction).
(2) A degenerate responder is exposed. Qwen-2.5-VL-72B scores 100\% on all three tasks (constant label regardless of image), explaining why it is surpassed by its smaller 32B counterpart in the main results.
(3) Strong general models can still be incoherent. Gemini-2.5-Pro reaches 87\% contradiction despite near-SOTA holistic scores.

\subsection{GPT-4.1 performance on the generated subsets}

\begin{table*}[h]
    \centering
    \caption{The performance of the GPT-4.1 model on the generated subsets in {\methodname}. The ``Original'' row reports performance on the original FRCT questions. The ``Normal'' row uses the same configuration as the original questions. The ``Easy'' and ``Hard'' rows correspond to questions that are modified to be easier and more difficult, respectively.}
    \label{tab:generated-results-4}
    \includegraphics[width=1.0\linewidth]{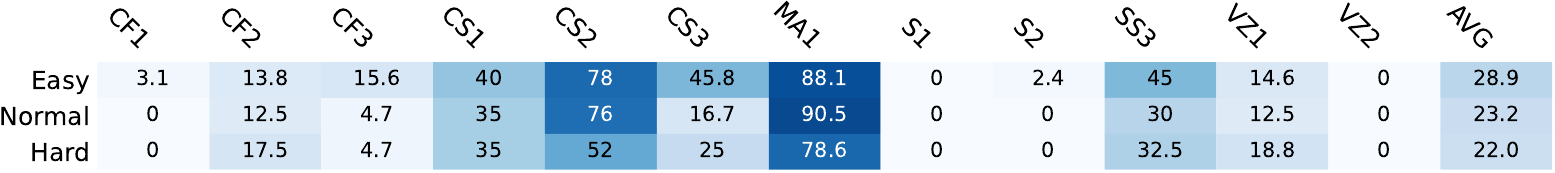}
\end{table*}

Our key findings are as follows:
(1) CS1--3 (object and word recognition under noise): The model achieves higher accuracy on our generated datasets compared to the original ones.
We attribute this to our selection of commonly encountered objects in daily life, which likely reduces recognition difficulty.
Moreover, our framework supports dynamic image updates, allowing the tests to be refreshed as needed in the future.
(2) MA1 (memory test): The original version requires memorizing 21 image-number pairs, a task on which the model achieves 100\% accuracy.
In contrast, our hard version increases the number of pairs to 50, resulting in a substantial performance drop, highlighting the increased challenge.
(3) VZ2 (paper folding test): The original dataset includes questions based on one to three folds.
Our version expands this to include up to five folds, significantly increasing task complexity.
The model fails to answer any of these questions correctly.
These results demonstrate that our generated dataset effectively supports dynamic adjustment of test difficulty, making it suitable for evaluating increasingly capable models.

\clearpage

\section{Failure analysis: a key bottleneck of visual recognition}
\label{sec:failure}

\paragraph{Rely on accurate textual descriptions}
Our evaluation reveals a contrast between models' strong textual reasoning capabilities and their markedly weaker visual perception performance.
This disparity is exemplified by CF3: when models are provided with textual descriptions of line segments (starting coordinates and direction vectors), GPT-4o achieves perfect accuracy (100\%).
In contrast, performance drops sharply when the same information has to be inferred from visual inputs, with accuracy falling to just 6.2\%.
Moreover, for tasks with semantically meaningful objects (\eg, MA1), models accurately describe visual content in their CoT, suggesting that the visual backbone captures the necessary details.
However, several cognitive tasks contain spatial configurations that cannot be faithfully verbalized.
For example, in VZ2, describing the exact relative distances and positions of holes is inherently insufficient, leading to incorrect predictions.

\paragraph{Fail to recognize visual details}
Our generated CF3 cases reveal that start-point identification accuracy decreases systematically with marker size variation: from 92\% with large circular markers to 80\% with medium markers, and ultimately 68\% with small markers.
Moreover, in SS2, models consistently fail to distinguish between intersecting lines with explicit junction markers versus those without visual indicators.
Additionally, models struggle to focus effectively on key regions, resulting in missing information.
Taking the partially erased word ``women'' (CS2) as an example.
Correct identification of the first character requires recognizing the lower left corner that differentiates ``w'' from ``v.''
Similarly, identifying the fifth character as ``n'' relies on detecting a small vertical line in the lower right corner of the letter.
Models misclassify these characters as ``v'' and ``r,'' respectively, indicating their limited ability to prioritize critical local features.
This degradation suggests fundamental constraints in the models' visual attention mechanisms, where reduced visual saliency directly compromises recognition performance.

\paragraph{Low sensitivity to length, angle, and proportion}
Models exhibit notable limitations in processing geometric shapes, particularly in assessing length, angle, and proportion (examples of GPT-4o are shown below).
Similarly, in VZ1, although models correctly identify the need for a rectangle to construct a complex figure, they fail to select sides of the appropriate length.
These results indicate that while models possess some geometric recognition abilities, they struggle with accurately gauging line lengths and proportions, limiting their performance in tasks requiring precise spatial measurements.
Moreover, our analysis reveals \textbf{a bias toward diagonal orientations}: models consistently misclassified various directions as 45-degree angles.
In a controlled test with 20 non-45-degree vectors (\eg, vector $(2,\ 1)$), models achieve zero correct angular identification, consistently defaulting to the nearest 45-degree approximation.
This suggests that models possess only coarse categorical representations of spatial orientation rather than continuous angular perception.

\begin{promptbox}{Failure of CF3: Copying Test (Length)}
{
\includegraphics[width=3cm]{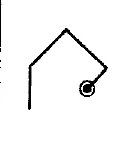} \includegraphics[width=3cm]{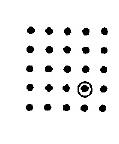} \newline
Step-by-Step Solution: \newline
1. I'm starting at the dot at (4, 4). The first line goes diagonally up and to the right, so I take one step to (3, 5). \newline
2. Then, there's a longer line going back up towards the left. I follow that diagonal path all the way up to (1, 3). \newline
3. From there, the shape turns down and to the left, so I make a small diagonal move to {\color{red} (2, 2)}. \newline
4. Finally, the last part is just a straight vertical line going down, which brings me to {\color{red} (3, 2)}.

The final coordinate is (3, 2).
}
\end{promptbox}

\begin{promptbox}{Failure of CF3: Copying Test (Angle)}
{
\includegraphics[width=3cm]{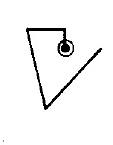} \includegraphics[width=3cm]{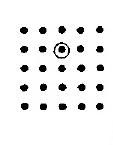} \newline
Step-by-Step Solution: \newline
1. Alright, starting at the circled dot which is at (2, 3). First, I see a short vertical line going straight up, so I'll move one spot to (1, 3). \newline
2. Then the line turns sharply to the left, going straight across to (1, 1). \newline
3. From there, it drops down in a long diagonal. It looks like a typical {\color{red} 45-degree angle}, so if I follow that path down and to the right, {\color{red} I go through (2, 2) and (3, 3) and land on (4, 4)}. \newline
4. Finally, there's a little tail pointing up and right, so I take one diagonal step to finish at (3, 5).

The final coordinate is ```{'answer': (3, 5)}```). 
}
\end{promptbox}

\begin{promptbox}{Failure of VZ2: Paper Folding Test (Proportion)}
{
\includegraphics[width=7cm]{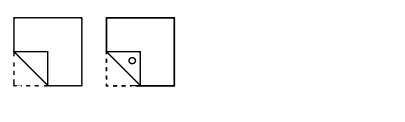} \newline
\includegraphics[width=1.6cm]{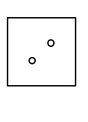} \newline
Step-by-Step Solution: \newline
1. **Initial Setup**: \newline
   - The square sheet starts flat. \newline
   - The first fold is made by folding the bottom-left corner up to the top-right corner, creating a triangular fold. \newline \newline
2. **Punching the Hole**: \newline
   - After the fold, a hole is punched through all layers at the marked position (near the center of the folded triangle). \newline \newline
3. **Unfolding the Sheet**: \newline
   - Reverse the fold to return the sheet to its original flat state. \newline
   - Since the hole was punched through all layers of the folded sheet, it will create multiple holes when unfolded. \newline \newline
4. **Analyzing the Hole Placement**: \newline
   - The fold creates two layers of paper. \newline
   - A single punch through the folded sheet results in two holes when unfolded. \newline
   - The two holes will be symmetrically placed along the diagonal fold line. \newline \newline
5. **Comparing with the Second Image**: \newline
   - {\color{red} The second image shows two holes symmetrically placed along the diagonal of the square sheet.} \newline
   - {\color{red} The positions of the holes in the second image match the expected result from the mental process.} \newline \newline
Final Answer: \newline
The second image matches the mental result exactly. {"answer": "TRUE"}
}
\end{promptbox}

\clearpage

\section{Implementation of parametric generation}
\label{sec:implementation}

\begin{figure*}[h]
    \centering
    \includegraphics[width=1.0\linewidth]{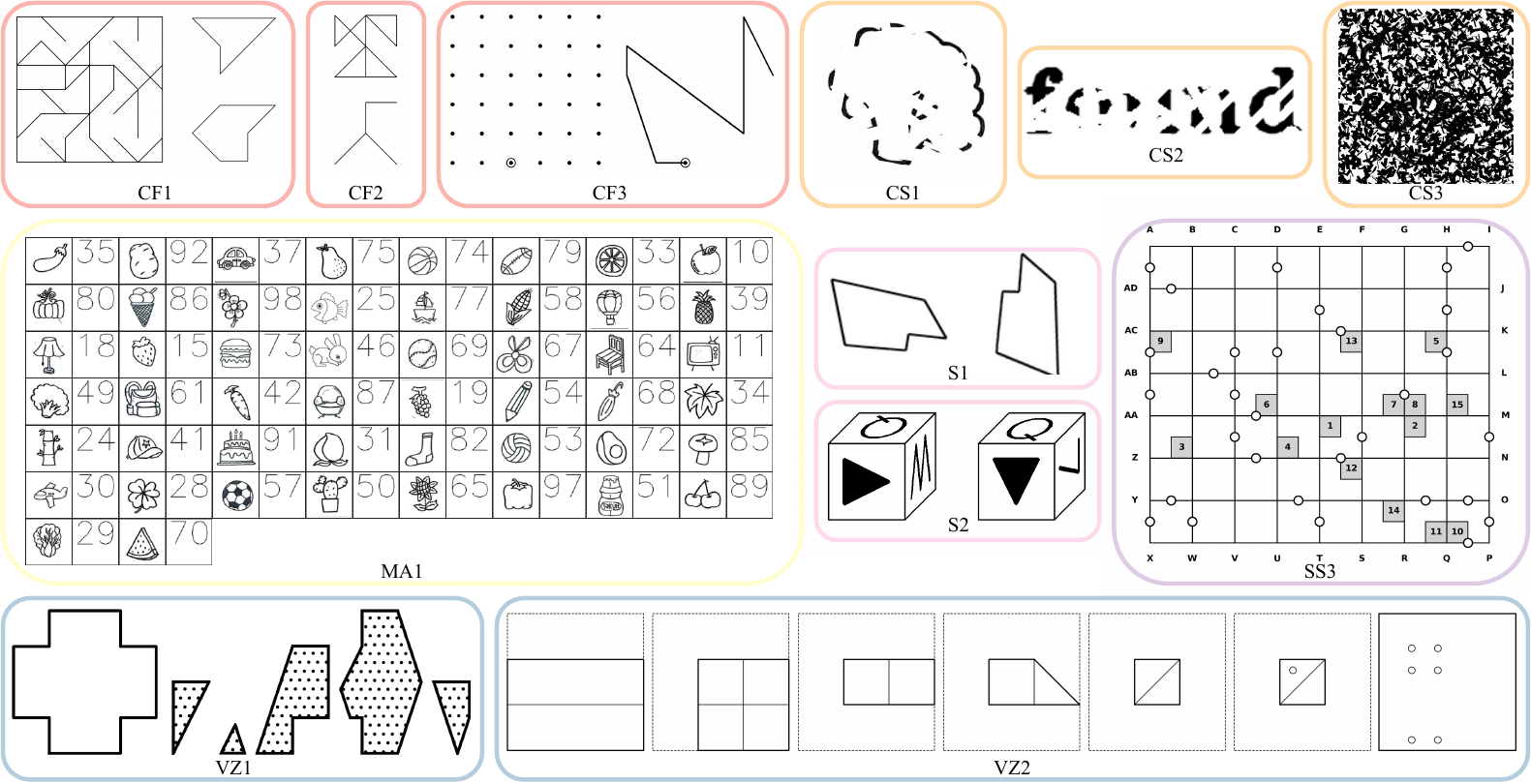}
    \caption{Samples of our generated images. We can dynamically adjust test difficulties in {\methodname}. For example, the grid size of CF3 is changed to $6 \times 6$ instead of $5 \times 5$, $8 \times 9$ instead of $7 \times 8$ for SS3, and more folding steps for VZ2.
    }
    \label{fig:generated-sample}
\end{figure*}

\subsection{CF1: Hidden Figures Test}

We model each pattern as a graph $G=(V,E)$ embedded on an axis-aligned $m \times n$ lattice whose admissible edges join adjacent vertices (4-neighbour plus the two diagonals).
Generation starts by deterministically adding the perimeter edges, thereby fixing a closed bounding rectangle and seeding a single connected component.
The target edge count is then drawn from $k\sim\mathcal{N}(\mu,\sigma^{2})$ with $\mu=\rho \lvert E \rvert$ and $\sigma=\rho_{\text{std}}\lvert E \rvert$ for user-specified density $\rho\in(0,\ 1]$ and $\rho_{\text{std}}$, and clipped to $[0,\ 1]\cdot\lvert E \rvert$.
For sub-pattern detection we represent the user-supplied ``model'' as its own edge set and enumerate all translations obtained by aligning any model vertex with any pattern vertex; containment reduces to a constant-time subset test per translation, which is tractable for the small grids used here and yields exact, translation-invariant matches without recursion or graph isomorphism search.

\subsection{CF2: Hidden Patterns Test}

We introduce a graph-based generator that operates on an $m \times n$ lattice.
We first enumerate the complete set $\mathcal{E}$ of admissible edges—unit horizontal, vertical, and diagonal connections between adjacent lattice nodes—yielding $E=\lvert\mathcal{E}\rvert$ potential segments.
To guarantee global connectivity, we draw a uniformly random spanning tree $T\subset\mathcal{E}$ by performing a depth-first search with randomized successor order; this yields exactly $N-1$ edges, where $N=mn$ is the number of nodes.
Desired edge density is controlled by sampling a target count $k\sim\mathcal{N}(\mu,\sigma^{2})$ with $\mu=\rho E$ and $\sigma=\rho_{\text{std}}E$ for user-specified density $\rho\in(0,\ 1]$ and $\rho_{\text{std}}$; the sample is clipped to $[N-1,\ E]$.
We then augment $T$ with $k-(N-1)$ additional edges drawn without replacement from $\mathcal{E}\setminus T$, producing a connected graph $G=(V,E_{G})$ whose expected density equals $\rho$.

\subsection{CF3: Copying Test}

We develop a procedural grid-walk generator that produces paired images.
Each instance begins by laying out an $m \times n$ lattice whose node coordinates are computed analytically from a single size parameter, ensuring scale-invariance across resolutions.
A start node is selected uniformly at random and a self-avoiding walk is grown whose length is drawn from a user-specified interval $[\mathrm{min\_steps},\mathrm{max\_steps}]$.
At every extension step, the candidate set comprises all yet-unvisited lattice nodes; candidates that would yield a line segment collinear with any existing segment in the path are deterministically excluded via a zero-cross-product test, preventing visual overlap and ensuring topological diversity.
Two images are rendered, a reference grid with the start node circled, and a path image of identical dimensions that shows only the start node and the resulting non-collinear walk.

\subsection{CS1: Gestalt Completion Test}

We begin by curating object silhouettes and their labels from public image repositories.
Each image is partially occluded with randomly oriented white strokes whose number and width scale linearly with a severity coefficient $s \in [0,\ 1]$.

\subsection{CS2: Concealed Words Test}

We synthesize a tunable corpus of occluded word images by sampling from the \texttt{top\_n\_list} in the \texttt{wordfreq} Python library, retaining alphabetic tokens whose lengths fall within a user-defined interval and converting them to lower-case.
Each word is rendered on a white canvas and then obfuscated by superimposing straight white line segments and circular blotches drawn at random positions.
The number, thickness, and radius of these artifacts increase linearly with a continuous severity parameter $s \in [0,\ 1]$, providing precise control over the level of visual concealment.

\subsection{CS3: Snowy Pictures}

Building on the silhouettes and labels introduced in CS1, we corrupt every input image in two successive steps.
First, we overlay $n_r$ white rectangles whose side lengths are sampled uniformly up to a fixed fraction of the image's shorter edge, disrupting local continuity.
Next, we draw $n_\ell$ short, randomly oriented black line segments that imitate dense, edge-like clutter.
Both $n_r$ and $n_\ell$ scale linearly with a severity parameter $s \in [0,\ 1]$.

\subsection{MA1: Picture-Number Test}

Also building on the source from CS1, we first draw $N$ unique items without replacement and an equal-sized set of distinct two-digit integers $\{10, \dots , 99\}$.
The two cells are concatenated horizontally to form an atomic pair, and all pairs are then tiled row-major into an $r\times c$ grid with $rc \ge N$ and $|r-c|$ minimized to approximate isotropy, yielding a visually balanced layout regardless of $N$.
A uniformly random pair is sampled to provide a query image and its label, while the full canvas supplies rich contextual clutter.

\subsection{S1: Card Rotations Test}

We devise a lightweight generator that first samples a simple, non-self-intersecting polygon by drawing \textit{i.i.d.} polar radii and sorted angles, and repeatedly rejecting candidates whose (i) shortest edge falls below a minimum-length threshold and (ii) consecutive edge-length differences are within a tight tolerance—two filters that jointly suppress near-symmetries and visually imperceptible edges.
We optionally apply a horizontal mirror, then rotate it by a uniformly random angle before centrally cropping back to the original spatial extent.
From every base polygon we generate $N$ views and record a binary label indicating whether the transformation involved only rotation (true) or a mirror-plus-rotation (false).

\subsection{S2: Cube Comparisons Test}

To decide whether two partial observations correspond to the same physical cube, we cast the problem as a constrained search over the 24 right-handed orientations of a cube in $\mathbb{Z}^3$.
We first ``pin'' the first view as the reference orientation—its Up, Front and Right faces become the intrinsic Up, Front, Right faces of the cube—which lets us record its three symbols and their rotations in a baseline face–rotation table.
For each of the 24 global orientations we then
(i) map the observer's local axes to intrinsic cube faces via simple cross-product geometry,
(ii) transform the second view's reported rotations into the reference frame by adding a pre-computed $90^\circ$ offset that aligns local ``Up'' vectors, and
(iii) enforce two consistency constraints:
(a) the same intrinsic face observed twice must carry identical symbols whose rotations are equivalent under the symbol's symmetry class (4-fold, 2-fold, or asymmetric), and
(b) a symbol may not appear on two different faces.
Finally, we randomly generate such three-face views and render them as perspective-correct 3-D cube images.

\subsection{SS3: Map Planning Test}

We model the city layout as a rectangular $m \times n$ lattice in an undirected graph, where each vertex represents a street intersection and each edge a unit-length street segment.
From the fully connected lattice we remove a user-specified fraction $r$ of edges, chosen uniformly at random, and tag their mid-points as circular ``road-blocks,'' thereby enforcing non-traversable segments while preserving the geometry for visualization.
$N_B$ quarter-square buildings are sampled without replacement from the $(m-1)(n-1)$ grid cells, along with the two edges each of them touches.
Perimeter intersections are labeled in clockwise order using spreadsheet-style indices (A–Z, AA, AB, $\dots$), after which start–end terminals are selected by random permutation until exactly one shortest path exists between them, which guarantees uniqueness while avoiding exhaustive search.
The final instance thus comprises a sparse planar graph with a provably unique geodesic, alongside metadata for blocked edges, buildings and perimeter labels.

\subsection{VZ1: Form Board Test}

We design an automatic pipeline that transforms an arbitrary lattice-defined polygon into a ``dissect-and-assemble'' puzzle while guaranteeing a unique solution under rotation and translation.
The target shape is first specified on an $n \times n$ integer grid as an ordered list of boundary edges.
A random integer $k\in \{3, 4, 5\}$ determines the number of genuine solution pieces.
Starting from the full polygon, we iteratively bisect the currently largest fragment with straight grid-aligned cuts whose slopes are limited to ${+\infty, 0, \pm1, \pm2, \pm3}$.
Each cut is accepted only if it produces two valid polygons, and the process terminates as soon as $k$ fragments are obtained.
To generate the remaining $5 - k$ distractor pieces, we re-cut one randomly chosen solution fragment, rejecting candidate fragments whose areas coincide with any existing piece, thereby ensuring that no spurious subset of distractors can reconstruct the target.

\subsection{VZ2: Paper Folding Test}

Starting from a unit-square sheet discretized into an $n \times n$ grid, our algorithm iteratively selects a random fold axis—horizontal, vertical, or an arbitrary offset diagonal of the form $y = \pm x + c$.
At each step, the square is partitioned by this axis; the half-plane judged closest to the sheet's geometric center remains stationary, while the opposite half is reflected via an analytic mapping that preserves affine structure.
Crucially, we maintain (i) a ``Polygon'' describing the current outer outline, (ii) an ordered list of internal edges and crease lines, and (iii) the exact set of point holes.
These entities are updated by reflecting only those primitives that lie on the moving half and clipping fold-axis segments to the unfolded outline, guaranteeing topological correctness even for degenerate or off-center folds.
The complete state history enables deterministic reverse unfolding to generate the answer: holes are ``back-propagated'' by conditional reflection.

\clearpage

\section{Descriptions and prompts for all subtests}
\label{sec:description}

FRCT assumes that each sub‑test isolates a single latent visual factor~\cite{ekstrom1974problems, ekstrom1975attempt}.
This section introduces each subtest in detail and provides the prompts we use in {\methodname}.

\subsection{Closure Flexibility (CF)}

\underline{The Factor:}
\begin{quotation}
\textit{``The ability to hold a given visual percept or configuration in mind so as to disembed it from other well defined perceptual material.''}
\end{quotation}

Flexibility of closure, a cognitive factor involving the identification of a configuration within a distracting perceptual field, has been linked to the concept of field independence, though they are not considered identical constructs.
\citet{witkin1971manual} related this factor to both Thurstone's flexibility of closure~\cite{thurstone1938primary} and Guilford's adaptive flexibility~\cite{guilford1967nature}, suggesting similarities to field independence.
\citet{royce1973conceptual} proposed that flexibility of closure may interact with higher-order cognitive factors, while \citet{hettema1968cognitive} posited it as conceptually situated between flexibility and speed of closure.
\citet{wardell1973possible} argued for its identity with figural adaptive flexibility.
\citet{carroll1974psychometric} defined flexibility of closure as involving short-term memory processes that match a figure to its surrounding field, and \citet{cattell1971abilities} framed it as a restructuring ability central to personality and practical intelligence.

\begin{promptbox}{Prompt for CF1: Hidden Figures Test}
{
Look at the two images:\newline \newline Below is the first image, one simple shape:\newline \includegraphics[width=3.0cm]{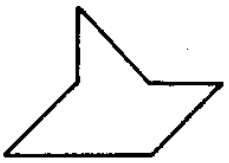} \newline \newline Below is the second image, a larger, complex pattern:\newline \includegraphics[width=3.5cm]{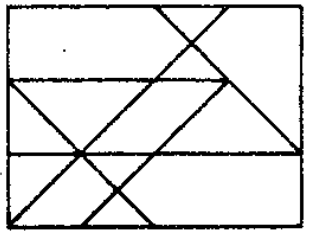} \newline \newline Task: Decide whether the shape in the first image is hidden anywhere inside the second image. The shape will never be rotated, flipped, or resized. The shape will always be right-side-up and exactly the same size as in the first image.\newline \newline Output: Respond with only one word: “TRUE” if it is present, “FALSE” if it is not, in JSON format as follows: \texttt{\{"answer": YOUR\_ANSWER\_HERE\}}.
}
\end{promptbox}

\begin{promptbox}{Prompt for CF2: Hidden Patterns Test}
{
Look at the two images:\newline \newline Below is the first image, a model:\newline \includegraphics[width=1.5cm]{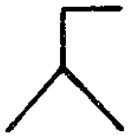} \newline \newline Below is the second image, a pattern:\newline \includegraphics[width=1.5cm]{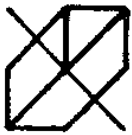} \newline \newline Task: Decide if the model in the first image is hidden anywhere in the pattern in the second image. The model must be in that exact position, no turning or flipping.\newline \newline Output: Respond with only one word: “TRUE” if it is present, “FALSE” if it is not, in JSON format as follows: \texttt{\{"answer": YOUR\_ANSWER\_HERE\}}.
}
\end{promptbox}

\begin{promptbox}{Prompt for CF3: Copying Test}
{
Look at the two images:\newline \newline Below is the first image, a simple line shape:\newline \includegraphics[width=2.5cm]{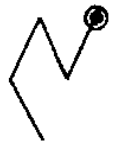} \newline \newline Below is the second image, a 5 times 5 grid of dots; one dot is circled as the starting point:\newline \includegraphics[width=2.5cm]{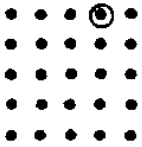} \newline \newline Task: Begin at the circled dot on the second image. Copy the shape shown in the first image onto the grid so that every corner of the line sits exactly on a dot. When you are done, the pattern on the grid must look the same as the shape in the first image.\newline \newline Output: Respond with only a tuple, the dot you finally reach, as a (row, column) pair where the row is counted top-to-bottom and the column left-to-right, in JSON format as follows: \texttt{\{"answer": YOUR\_ANSWER\_HERE\}}.
}
\end{promptbox}

\clearpage

\subsection{Closure Speed (CS)}

\underline{The Factor:}
\begin{quotation}
\textit{``The ability to unite an apparently disparate perceptual field into a single concept.''}
\end{quotation}

The concept of speed of closure refers to the ability to rapidly recognize and organize ambiguous or partially obscured visual stimuli, a process distinct from flexibility of closure, which involves identifying a known configuration within complex figures.
This skill is associated with the early identification of out-of-focus and close-up images~\cite{frederiksen1967cognitive}, and involves long-term memory search strategies~\cite{carroll1974psychometric}.
It has been linked to cognitive factors like restraint-timidity~\cite{cattell1971abilities} and may reflect a broader aptitude for visual scanning and cognitive-affective integration~\cite{thurstone1944factorial, wardell1973possible, roff1952factorial, adcock1971flexibility, messick1975dimensions}.

\begin{promptbox}{Prompt for CS1: Gestalt Completion Test}
{
Look at the incomplete drawing below:\newline \includegraphics[width=5.0cm]{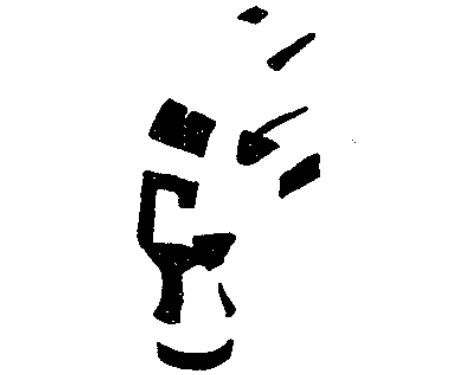} \newline \newline Task: Write the name of the object you think it shows.\newline \newline Output: Respond with only one or two words, in JSON format as follows: \texttt{\{"answer": YOUR\_ANSWER\_HERE\}}.
}
\end{promptbox}

\begin{promptbox}{Prompt for CS2: Concealed Words Test}
{
Look at the image below, which shows one lowercase English word, but parts of the letters are missing:\newline \includegraphics[width=4.5cm]{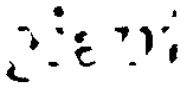} \newline \newline Task: Write the complete word. The word is at least four letters long. Use only lowercase letters.\newline \newline Output: Respond with only the answer word, in JSON format as follows: \texttt{\{"answer": YOUR\_ANSWER\_HERE\}}.
}
\end{promptbox}

\begin{promptbox}{Prompt for CS3: Snowy Pictures}
{
Look at this image below:\newline \includegraphics[width=5.5cm]{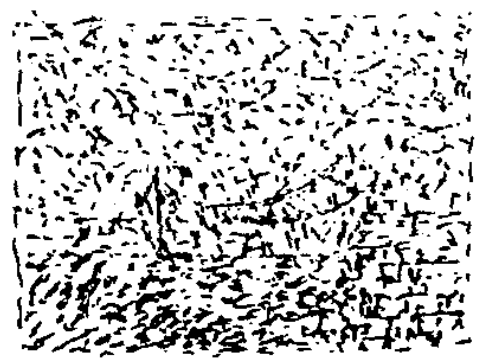} \newline \newline Task: Even if parts are hidden, name the main object you see.\newline \newline Output: Respond with only one or two words, in JSON format as follows: \texttt{\{"answer": YOUR\_ANSWER\_HERE\}}.
}
\end{promptbox}

\clearpage

\subsection{Induction (I)}

\underline{The Factor:}
\begin{quotation}
\textit{``The reasoning abilities involved in forming and trying out hypotheses that will fit a set of data.''}
\end{quotation}

Research on inductive reasoning suggests it involves both concept formation and hypothesis testing, functioning as a synthesizing process~\cite{wardell1973possible}.
Evidence points to several subfactors, with figure classification being particularly distinct~\cite{harris1971factor}.
\citet{guilford1966sixteen} identified 16 types of inductive ability, while \citet{dye1968growth} proposed distinct inductive and symbolic-inductive reasoning factors.
Though \citet{pawlick1966concepts} argued that induction and general reasoning are not separate, \citet{cattell1971abilities} allowed for a possible figural reasoning factor.
\citet{carroll1974psychometric} emphasized the role of long-term memory search in induction, noting that success depends on the content of a ``general logic store'' and the ability to construct new hypotheses through serial operations.

\begin{promptbox}{Prompt for I3: Figure Classification}
{
Look at the four images:\newline \newline Below is the first image, three figures in the Group 1:\newline \includegraphics[width=4.5cm]{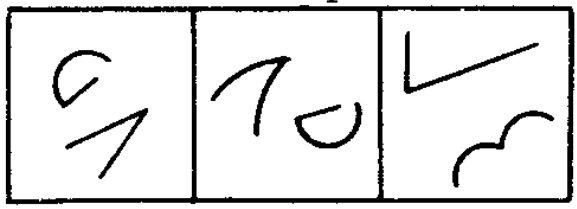} \newline \newline Below is the second image, three figures in the Group 2:\newline \includegraphics[width=4.5cm]{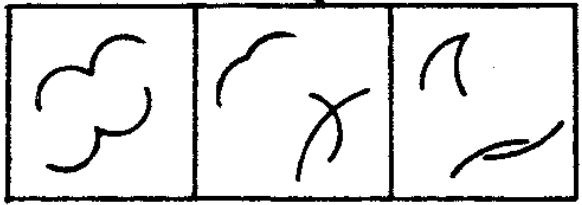} \newline \newline Below is the second image, three figures in the Group 3:\newline \includegraphics[width=4.5cm]{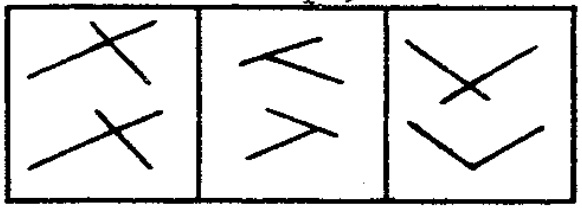} \newline \newline Below is the fourth image, the figure to classify: \newline\includegraphics[width=1.5cm]{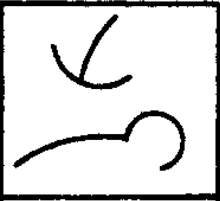} \newline \newline Task: Inside a group, all three figures share one rule. Different groups follow different rules. Find the rule and decide whether the figure in the fourth image belongs to Group 1, 2, or 3.\newline \newline Output: Respond with only the group number (1, 2, or 3), in JSON format as follows: \texttt{\{"answer": YOUR\_ANSWER\_HERE\}}.
}
\end{promptbox}

\newpage

\subsection{Associative Memory (MA)}

\underline{The Factor:}
\begin{quotation}
\textit{``The ability to recall one part of a previously learned but otherwise unrelated pair of items when the other part of the pair is presented.''}
\end{quotation}

Tasks assessing this factor are similar to those used in paired-associates learning and may involve memory for non-meaningful material.
This factor reflects intermediate-term memory processes, where individual differences arise from the use of strategies such as short-term rehearsal and the identification of mnemonic mediators in long-term memory~\cite{carroll1974psychometric}.

\begin{promptbox}{Prompt for MA1: Picture-Number Test}
{
Look at the two images:\newline \newline Below is the first image, the 21 picture-number pairs to memorize:\newline \includegraphics[width=7.5cm]{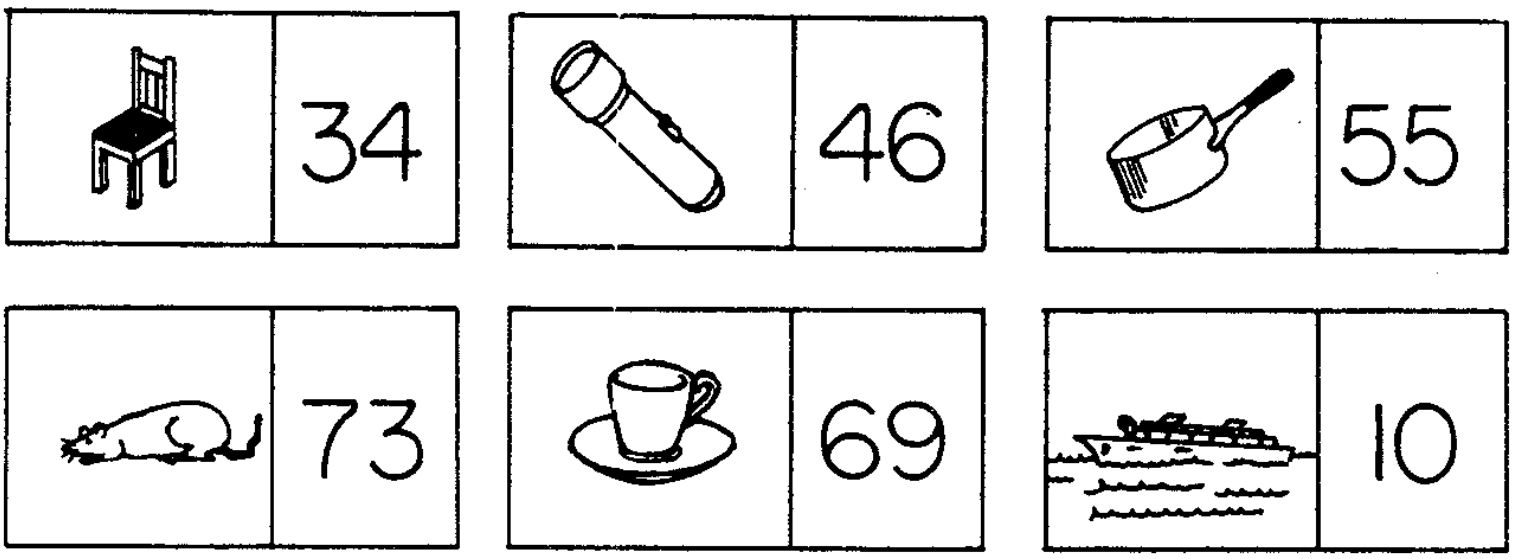} \newline \newline Below is the second image, a picture:\newline \includegraphics[width=1.5cm]{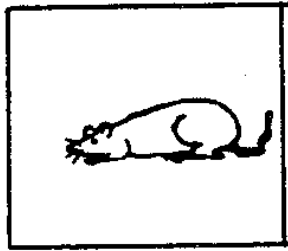} \newline \newline Task: Write down the number that the picture in the second image belongs to, as shown in the first image.\newline \newline Output: Respond with only a number, in JSON format as follows: \texttt{\{"answer": YOUR\_ANSWER\_HERE\}}.
}
\end{promptbox}

\clearpage

\subsection{Visual Memory (MV)}

\underline{The Factor:}
\begin{quotation}
\textit{``The ability to remember the configuration, location, and orientation of figural material.''}
\end{quotation}

Visual memory involves distinct cognitive processes beyond mere test content, as suggested by research on iconic memory, which stores visual impressions~\cite{thurstone1946theories}.
While \citet{thurstone1946theories} argued that ``the memorizing factor transcends the nature of the content,'' later studies demonstrated that visual memory is a multifaceted construct.
\citet{guilford1967nature} identified six figural memory abilities, and \citet{petrov1970memory} distinguished between factors for iconic memory and short-term visual retention, indicating the presence of sub-factors within visual memory.

\begin{promptbox}{Prompt for MV1: Shape Memory Test}
{
Look at the two images:\newline \newline Below is the first image, memorize each shape and the way it is turned:\newline \includegraphics[width=7.5cm]{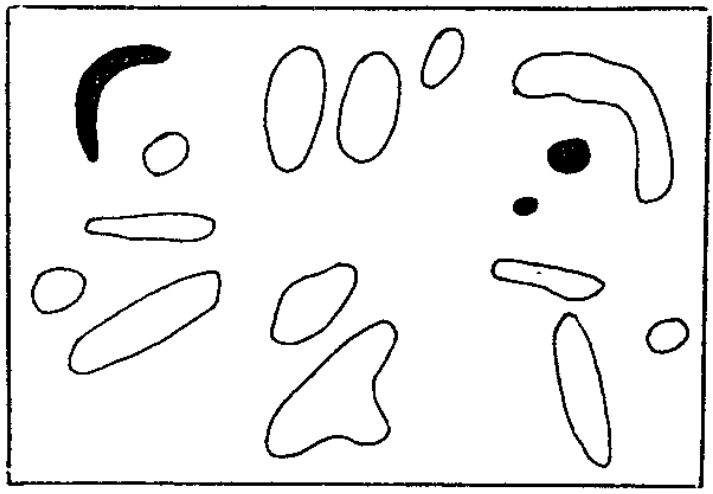} \newline \newline Below is the second image:\newline \includegraphics[width=2.5cm]{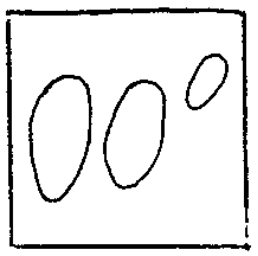} \newline \newline Task: Decide whether the following statement is true or false: the second image does not show any part of the first image with the same shapes in the same orientation.\newline (!!!) Three other prompts are: (1) the second image does not show any part of the first image with the same shapes in the same orientation (2) some part of the first image contains the second image with the same shapes in the same orientation (3) some part of the first image does not contain the second image with the same shapes in the same orientation\newline \newline Output: Respond with only one word: “TRUE” or “FALSE”, in JSON format as follows: \texttt{\{"answer": YOUR\_ANSWER\_HERE\}}.
}
\end{promptbox}

\begin{promptbox}{Prompt for MV2: Building Memory}
{
Look at the two images:\newline \newline Below is the first image, memorize where every building sits on this street map:\newline \includegraphics[width=6.5cm]{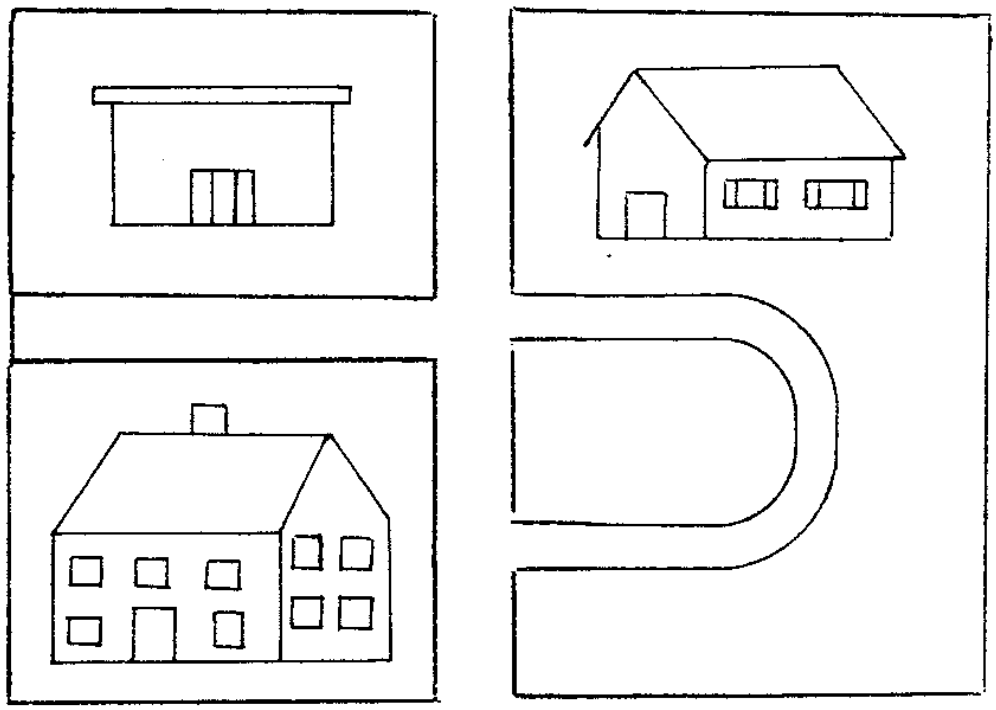} \newline \newline Below is the second image, the streets are the same, but each block is labeled A, B, C, D, E:\newline \includegraphics[width=6.5cm]{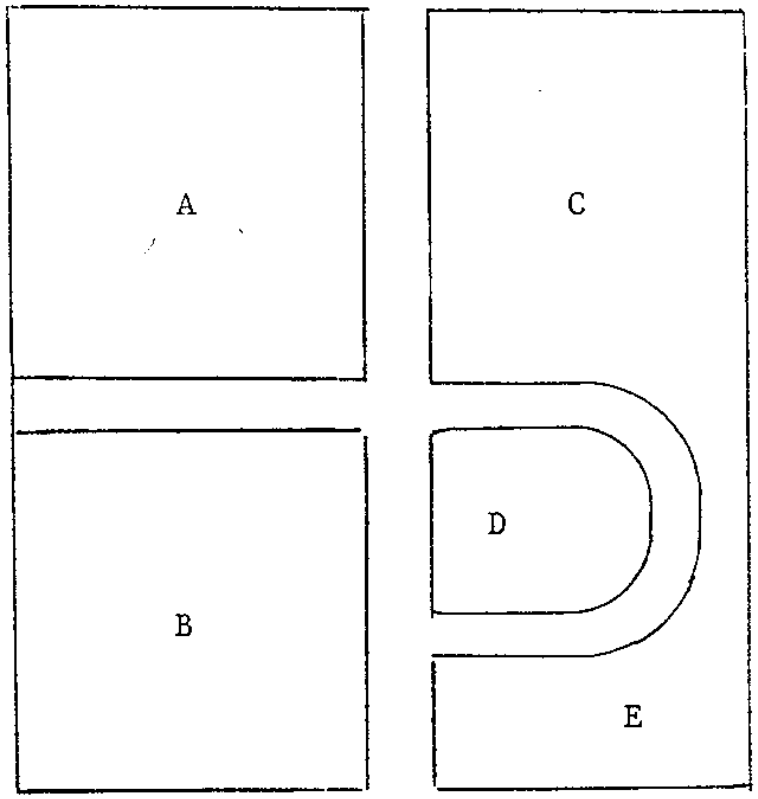} \newline \newline Below is the third image, a building:\newline \includegraphics[width=3.0cm]{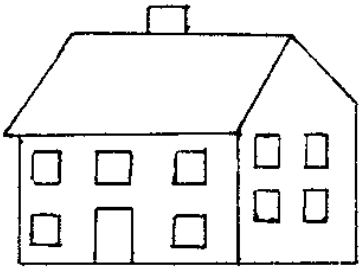} \newline \newline Task: Decide whether the building in the third image is in block E.\newline \newline Output: Respond with only one word: “TRUE” if it is, “FALSE” if it is not, in JSON format as follows: \texttt{\{"answer": YOUR\_ANSWER\_HERE\}}.
}
\end{promptbox}

\onecolumn

\begin{promptbox}{Prompt for MV3: Map Memory}
{
Look at the two images:\newline \newline Below is the first image, memorize each map:\newline \includegraphics[width=13.5cm]{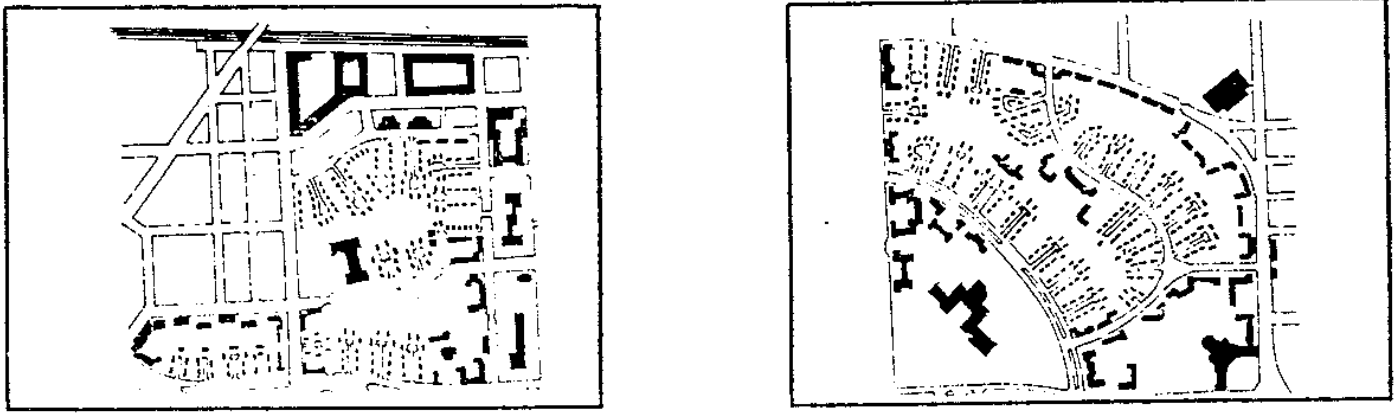} \newline \newline Below is the second image, a single map:\newline \includegraphics[width=6cm]{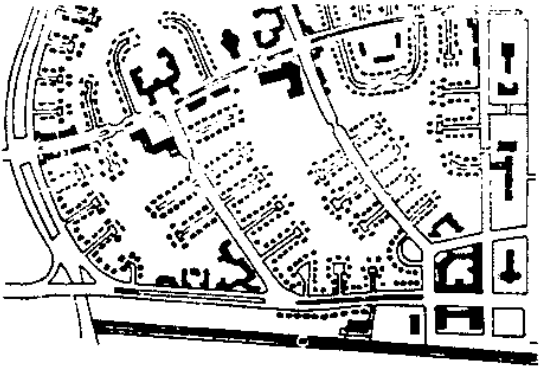} \newline \newline Task: Decide whether the following statement is true or false: the map in the second image appears in the first image.\newline (!!!) Three other prompts are: (1) the map in the second image does not appear in the first image (2) the maps in the first image contain the map in the second image (3) the maps in the first image do not contain the map in the second image\newline \newline Output: Respond with only one word: “TRUE” or “FALSE”, in JSON format as follows: \texttt{\{"answer": YOUR\_ANSWER\_HERE\}}.
}
\end{promptbox}

\clearpage

\subsection{Perceptual Speed (P)}

\underline{The Factor:}
\begin{quotation}
\textit{``Speed in comparing figures or symbols, scanning to find figures or symbols, or carrying out other very simple tasks involving visual perception.''}
\end{quotation}

Perceptual speed has been described as comprising three components:
(1) perceptual fluency, or the readiness with which individuals switch between alternating percepts;
(2) decision speed, or the readiness of choice when the response is not fully driven by sensory input~\cite{thurstone1938primary, kunnapas1969figural};
and (3) immediate perceptual memory.
\citet{carroll1974psychometric} defines perceptual speed as involving the temporal aspects of visual search through a field of specified elements by accessing sensory buffers.
It may be related to flexibility of closure~\cite{pawlick1966concepts, ekstrom1973cognitive} or to an ``automatic process'' factor.
Additionally, \citet{royce1973conceptual} suggested it may be a subfactor of the scanning cognitive style and possibly linked to the automatization cognitive style.
It may be the centroid of several subfactors (including form discrimination and symbol discrimination) which can be separated but are more usefully treated as a single concept for research purposes.

\begin{promptbox}{Prompt for P3: Identical Pictures Test}
{
Look at the two images:\newline \newline Below is the first image, the target object:\newline \includegraphics[width=1.2cm]{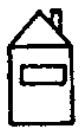} \newline \newline Below is the second image, the test object:\newline \includegraphics[width=1.2cm]{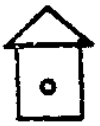} \newline \newline Task: Decide whether the two objects are exactly the same.\newline \newline Output: Respond with only one word: “TRUE” if they are, “FALSE” if they are not, in JSON format as follows: \texttt{\{"answer": YOUR\_ANSWER\_HERE\}}.
}
\end{promptbox}

\newpage

\subsection{Logical Reasoning (RL)}

\underline{The Factor:}
\begin{quotation}
\textit{``The ability to reason from premise to conclusion, or to evaluate the correctness of a conclusion.''}
\end{quotation}

The cognitive factor historically referred to as ``Deduction''~\cite{thurstone1938primary}, later termed ``Syllogistic Reasoning,'' and also known as ``Logical Evaluation'', involves evaluating the correctness of presented answers rather than pure deductive reasoning~\cite{guilford1967nature}.
\citet{carroll1974psychometric} emphasized its complexity, highlighting the need for retrieving meanings and algorithms from long-term memory and applying serial operations, with individual differences influenced by content, timing, and attentional focus on stimuli.

\begin{promptbox}{Prompt for RL2: Diagramming Relationships}
{
Look at the image below:\newline \includegraphics[width=2.0cm]{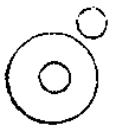} \newline \newline Each circle stands for one group of things. Simple rules:\newline 1. A circle inside another: all things in the inner group belong to the outer group.\newline 2. Circles that overlap partly: the two groups share some, but not all, things.\newline 3. Circles that do not touch: the two groups share nothing.\newline \newline Task: Decide whether the image follows these rules for the three groups: Desks, furniture, pencils.\newline \newline Output: Respond with only one word: “TRUE” if it shows the relationships for the three groups, “FALSE” if it does not, in JSON format as follows: \texttt{\{"answer": YOUR\_ANSWER\_HERE\}}.
}
\end{promptbox}

\clearpage

\subsection{Spatial Relations (S)}

\underline{The Factor:}
\begin{quotation}
\textit{``The ability to perceive spatial patterns or to maintain orientation with respect to objects in space.''}
\end{quotation}

Research has differentiated between spatial orientation and visualization, suggesting that while spatial orientation involves perceiving figures as wholes and performing mental rotation~\cite{zimmerman1954influence, werdelin1971relationship}, visualization requires more complex restructuring and serial operations~\cite{carroll1974psychometric, shepard1971mental}.
Although some distinguished between spatial relations and orientation (with the latter involving the observer's body), \citet{guilford1971analysis} treated them as a single cognitive factor linked to egocentrism.

\begin{promptbox}{Prompt for S1: Card Rotations Test}
{
Look at the two images:\newline \newline Below is the first image, the target shape:\newline \includegraphics[width=1.5cm]{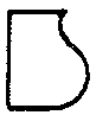} \newline \newline Below is the second image, the test shape:\newline \includegraphics[width=2.0cm]{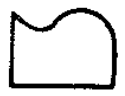} \newline \newline Task: The test shapes may be rotated, but they are not allowed to be flipped (mirrored). Decide whether test shape is the same shape as the target.\newline \newline Output: Respond with only one word: “TRUE” if it is, “FALSE” if it is not, in JSON format as follows: \texttt{\{"answer": YOUR\_ANSWER\_HERE\}}.
}
\end{promptbox}

\begin{promptbox}{Prompt for S2: Cube Comparisons Test}
{
Look at the two images:\newline \newline Below is the first image, the first cube:\newline \includegraphics[width=2.5cm]{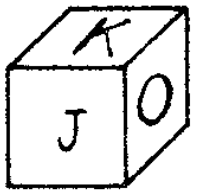} \newline \newline Below is the second image, the second cube:\newline \includegraphics[width=2.5cm]{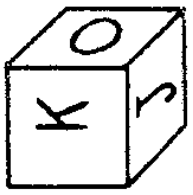} \newline \newline Rules:\newline 1. Each cube has six faces. Every face shows a different letter, number, or symbol.\newline 2. Hidden faces may show any symbols, but no symbol appears on more than one face of the same cube.\newline \newline Task: Decide whether the following statement is true or false: the first cube is a certain view of the second cube after it is turned.\newline (!!!) Three other prompts are: (1) the first cube is not any view of the second cube no matter how it is turned (2) the second cube is a certain view of the first cube after it is turned (3) the second cube is not any view of the first cube no matter how it is turned\newline \newline Output: Respond with only one word: “TRUE” or “FALSE”, in JSON format as follows: \texttt{\{"answer": YOUR\_ANSWER\_HERE\}}.
}
\end{promptbox}

\clearpage

\subsection{Spatial Scanning (SS)}

\underline{The Factor:}
\begin{quotation}
\textit{``Speed in exploring visually a wide or complicated spatial field.''}
\end{quotation}

The ability to navigate a paper maze relies on quickly scanning for viable paths and rejecting false leads, engaging a visual search process somewhat akin to scanning text for comprehension.
While sometimes associated with ``planning,'' the process primarily reflects a willingness to visually evaluate options before committing.
\citet{carroll1974psychometric} noted that this skill involves managing sensory input and that individuals may adopt strategies such as working backward from the goal to simplify the task.

\begin{promptbox}{Prompt for SS2: Choosing A Path}
{
Look at the diagram shown in the image below:\newline \includegraphics[width=7.5cm]{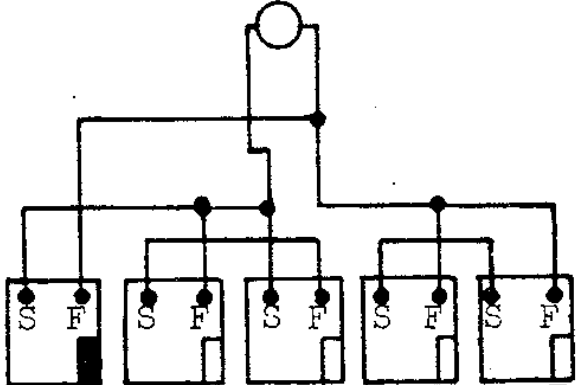} \newline \newline Rules:\newline 1. You may switch lines only where a black dot is drawn.\newline 2. Lines that cross or touch without a dot are not connected.\newline 3. The path must stay inside the chosen box and must not stop at a dead-end.\newline \newline Task: For box E, decide if there is one continuous line that:\newline 1. Starts at S inside that box.\newline 2. Reaches the single circle at the top.\newline 3. Comes back to F inside the same box without entering any other box.\newline \newline Output: Respond with only one word: “TRUE” if box E meets all the rules, “FALSE” if it does not, in JSON format as follows: \texttt{\{"answer": YOUR\_ANSWER\_HERE\}}.
}
\end{promptbox}

\begin{promptbox}{Prompt for SS3: Map Planning Test}
{
Look at the city map shown in the image below:\newline \includegraphics[width=7.5cm]{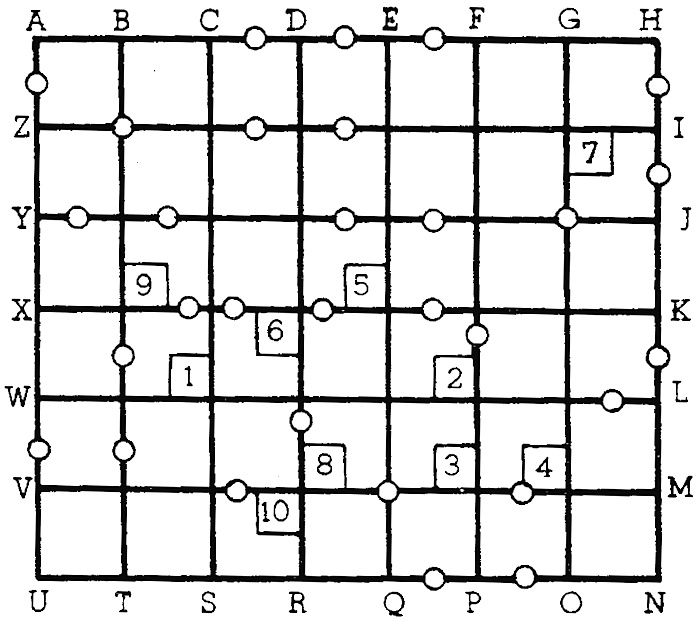} \newline \newline In the map:\newline 1. Streets = black lines.\newline 2. Circles = road-blocks (you cannot cross there).\newline 3. Numbered squares = buildings.\newline \newline Task: Find the shortest street route from F to T. Rules:\newline 1. The route will always touch the side of one and only one numbered building.\newline 2. Touching only a corner does not count.\newline 3. Move only along streets (horizontal or vertical), never through circles.\newline \newline Output: Respond with only one number: the number on the building your shortest route touches, in JSON format as follows: \texttt{\{"answer": YOUR\_ANSWER\_HERE\}}.
}
\end{promptbox}

\clearpage

\subsection{Visualization (VZ)}

\underline{The Factor:}
\begin{quotation}
\textit{``The ability to manipulate or transform the image of spatial patterns into other arrangements.''}
\end{quotation}

Visualization and spatial orientation are related cognitive factors, yet visualization involves mentally restructuring figures into components for manipulation, making it more complex than spatial orientation, which deals with rotating entire figures.
While some researchers view visualization as a higher-order or secondary factor encompassing various spatial abilities~\cite{cattell1971abilities, royce1973conceptual}, others emphasize its reliance on short-term visual memory and serial processing~\cite{carroll1974psychometric}.
Analytic strategies, such as identifying symmetry and reflection planes, are often used in visualization tasks, as illustrated by \citet{shepard1972chronometric}'s work on paper-folding tests.

\begin{promptbox}{Prompt for VZ1: Form Board Test}
{
Look at the two images:\newline \newline Below is the first image, which is the figure you must make:\newline \includegraphics[width=2.5cm]{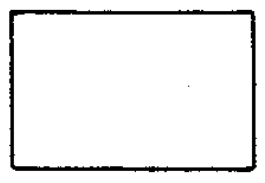} \newline \newline Below is the second image, which are the five pieces you can use:\newline \includegraphics[width=7.5cm]{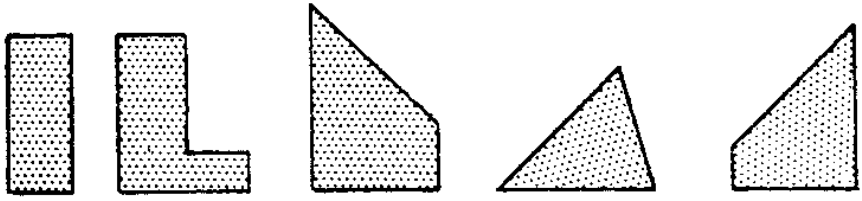} \newline \newline Rules:\newline 1. Use 2–5 of the pieces to fill the figure exactly.\newline 2. You may rotate pieces but do not flip them.\newline \newline Task: Decide whether the Fifth piece is in the set of pieces that makes the figure.\newline \newline Output: Respond with only one word: “TRUE” if it is or “FALSE” if it is not, in JSON format as follows: \texttt{\{"answer": YOUR\_ANSWER\_HERE\}}.
}
\end{promptbox}

\begin{promptbox}{Prompt for VZ2: Paper Folding Test}
{
Look at the two images:\newline \newline Below is the first image, a step-by-step drawing of a square sheet being folded (solid lines) and then punched (small circle marks):\newline \includegraphics[width=4.0cm]{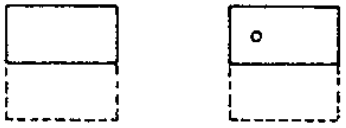} \newline \newline Below is the second image, the same sheet shown completely unfolded, with any holes that appear:\newline \includegraphics[width=1.4cm]{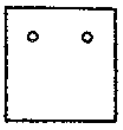} \newline \newline Task:\newline 1. Mentally follow every fold in the first image exactly as drawn. Do not flip or rotate the paper except for the folds shown.\newline 2. Imagine a hole being punched through all layers where each circle is drawn.\newline 3. Unfold the paper, step by step, in reverse order of the folds, keeping the sheet's original orientation.\newline 4. After it is flat, note where every hole should appear on the sheet.\newline 5. Compare this mental result with the pattern of holes in the second image.\newline \newline Output: Respond with only one word: “TRUE” if every hole (number and position) in the second image matches your mental result exactly, otherwise “FALSE”, in JSON format as follows: \texttt{\{"answer": YOUR\_ANSWER\_HERE\}}.
}
\end{promptbox}

\begin{promptbox}{Prompt for VZ3: Surface Development Test}
{
Look at the two images:\newline \newline Below is the first image, the flat paper:\newline \includegraphics[width=3.3cm]{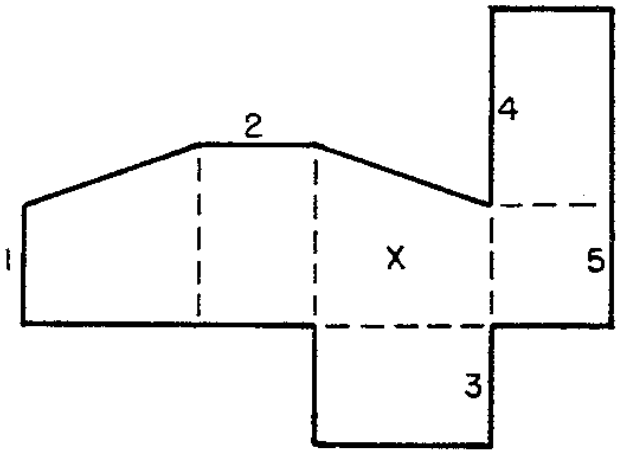} \newline \newline Below is the second image, the 3-D object:\newline \includegraphics[width=2.5cm]{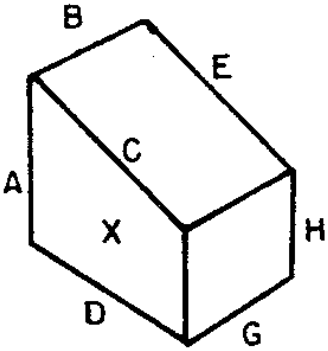} \newline \newline Task: Fold the flat paper in the first image on every dashed line so that the face marked X ends up on the outside of the 3-D object in the second image. Decide edge 5 on the flat paper in the first image touches which lettered edge on the 3-D object in the second image after folding.\newline (!!!) Decide whether the pair of one letter on the 3-D object in the second image and one number on the flat paper in the first image: (5, H) are two edges that touch each other after folding.\newline \newline Output: Respond with only one letter, in JSON format as follows: \texttt{\{"answer": YOUR\_ANSWER\_HERE\}}.\newline (!!!) Output: Respond with only one word: “TRUE” if they touch, “FALSE” if they do not, in JSON format as follows: \texttt{\{"answer": YOUR\_ANSWER\_HERE\}}.
}
\end{promptbox}

\clearpage

\section{Solutions from GPT-4o}

We test GPT-4o without image inputs, aiming to evaluate its ability to infer the correct solution to each subtest.
We can see that GPT-4o already knows the feasible solutions, indicating that its failure mainly comes from the vision recognition and vision-language alignment part.

\begin{promptbox}{Solution to CF1: Hidden Figures Test}
{
1. **Identify the Pattern**: Examine the small shape in the first image and record its exact pixel or cell configuration (e.g., a 2D grid of colors or pixels). \newline
2. **Scan the Larger Image**: Systematically slide a window of the same size as the first image over the second image, checking each possible sub-region. \newline
3. **Compare**: For each sub-region, check if it matches the pattern from the first image exactly—no rotation, flip, or size change allowed. \newline
4. **Decision**: If an exact match is found, output `{"answer": "TRUE"}`. If no match is found after scanning the entire larger image, output `{"answer": "FALSE"}`.
}
\end{promptbox}

\begin{promptbox}{Solution to CF2: Hidden Patterns Test}
{
1. **Identify Model Dimensions**: Note the size (rows x columns) of the model in the first image. \newline
2. **Scan Pattern Image**: Slide a window of the same dimensions across the second image (top-left to bottom-right). \newline
3. **Check for Exact Match**: At each position, compare the sub-section of the pattern with the model. \newline
4. **No Rotation or Flip**: Ensure the comparison uses the model as-is, without any transformations. \newline
5. **Return Result**: If an exact match is found, return `{"answer": "TRUE"}`. Otherwise, return `{"answer": "FALSE"}`.
}
\end{promptbox}

\begin{promptbox}{Solution to CF3: Copying Test}
{
1. **Observe the shape** in the first image and break it into straight line segments along the grid. \newline
2. **Start at the circled dot** in the second image. \newline
3. **Trace the same movements** (up/down/left/right/diagonal) from the start point, replicating the shape exactly by placing corners on the grid dots. \newline
4. **Count steps carefully** to ensure each corner aligns with a grid dot as in the original shape. \newline
5. **Record the final dot** reached after completing the entire shape.
}
\end{promptbox}

\begin{promptbox}{Solution to CS1: Gestalt Completion Test}
{
1. **Observe the drawing**: Look closely at the curved and linear segments to infer what object is being sketched. \newline
2. **Look for familiar outlines**: Identify key features—shapes, proportions, and positioning—that suggest a common object (e.g., wheels, body, handles). \newline
3. **Mentally complete the figure**: Use the partial lines to visualize what the full object would look like. \newline
4. **Identify the object**: Based on the partial sketch, determine the most likely object.
}
\end{promptbox}

\begin{promptbox}{Solution to CS2: Concealed Words Test}
{
1. **Analyze the visible fragments**: Identify parts of letters that are still visible and match them to possible lowercase letters. \newline
2. **Visualize missing parts**: Mentally fill in the gaps based on typical letter structures. \newline
3. **Look for patterns**: Combine identified letters into a coherent word, considering common English words. \newline
4. **Verify length**: Ensure the word is at least four letters long and uses only lowercase letters.
}
\end{promptbox}

\begin{promptbox}{Solution to CS3: Snowy Pictures}
{
1. **Identify visible features**: Focus on the parts that are not hidden—shape, color, structure, or details that hint at the object. \newline
2. **Infer the whole object**: Use context and symmetry to mentally complete the object, even if part is obscured. \newline
3. **Choose the most likely object**: Based on the visible portion and common objects with that appearance.
}
\end{promptbox}

\begin{promptbox}{Solution to I3: Figure Classification}
{
1. **Examine Group 1 and Group 2 figures**: Look for common traits shared within each group (e.g., shape count, orientation, fill patterns, symmetry). \newline
2. **Identify the rule per group**: Determine what consistent rule applies to all three figures in each group (e.g., all shapes have a diagonal line, or all contain a specific number of elements). \newline
3. **Compare rules between groups**: Make sure the rule is not shared across groups—each group must have a distinct rule. \newline
4. **Analyze the figure to classify**: Determine which group’s rule the new figure follows. \newline
5. **Assign it to the correct group**: Match the figure to the group with the corresponding visual rule.
}
\end{promptbox}

\begin{promptbox}{Solution to MA1: Picture-Number Test}
{
1. **Study the 21 picture-number pairs** in the first image: Memorize or note the associations between each unique picture and its corresponding number. \newline
2. **Examine the picture in the second image**: Identify the object or scene shown. \newline
3. **Match the second image** to one of the 21 pictures from the first image by comparing visual features. \newline
4. **Retrieve the associated number** from the first image that corresponds to the matched picture. \newline
5. **Return the number** in the required JSON format.
}
\end{promptbox}

\begin{promptbox}{Solution to MV1: Shape Memory Test}
{
1. **Memorize the shapes and orientations** in the first image: Focus on each shape's design and the direction it's facing (rotation or reflection). \newline
2. **Examine the second image**: Identify the specific shape and its orientation shown here. \newline
3. **Compare it to the memorized shapes**: Look for an exact match in both shape and orientation from the first image. \newline
4. **Evaluate the statement**: Determine if the given claim (e.g., “This shape matches one from the first image”) accurately reflects what is shown. \newline
5. **Decide if the statement is TRUE or FALSE** based on your comparison.
}
\end{promptbox}

\begin{promptbox}{Solution to MV2: Building Memory}
{
1. **Memorize the street map** in the first image: Note the position of each unique building relative to the street layout. \newline
2. **Study the block labels** in the second image: Understand how the blocks (A–E) correspond to the same street layout from the first image. \newline
3. **Identify the building** in the third image: Match its shape, size, and features to one on the original street map. \newline
4. **Locate that building** on the labeled block map from the second image. \newline
5. **Determine if it is in the specified block**: Compare its actual position to the named block in the question.
}
\end{promptbox}

\begin{promptbox}{Solution to MV3: Map Memory}
{
1. **Memorize the maps** in the first image: Focus on the layout of walls, paths, and any unique structures in each map. \newline
2. **Examine the single map** in the second image: Pay attention to the same features—structure, layout, and orientation. \newline
3. **Compare the second map** to the ones memorized: Check for exact matches or close similarities, including possible rotations or reflections. \newline
4. **Evaluate the statement**: Determine whether it correctly asserts a match (or lack thereof) between the second map and any from the first image. \newline
5. **Answer TRUE or FALSE** depending on whether the claim aligns with your comparison.
}
\end{promptbox}

\begin{promptbox}{Solution to P3: Identical Pictures Test}
{
1. **Study the target object** in the first image: Note its overall shape, orientation, components, and details. \newline
2. **Examine the test object** in the second image: Observe the same features—shape, structure, and orientation. \newline
3. **Compare both objects** precisely: Check for any differences in angles, positioning, parts, or missing elements. \newline
4. **Determine exact match**: Decide if the test object is an identical copy of the target object in all aspects.
}
\end{promptbox}

\begin{promptbox}{Solution to RL2: Diagramming Relationships}
{
1. **Understand the group relationships described** in the statement (e.g., one group is a subset of another, or groups partially overlap or are completely separate). \newline
2. **Interpret the image**: Analyze how the circles are arranged—nested, overlapping, or separate. \newline
3. **Match visual arrangement to rules**: \newline
   - Inner circle → subset. \newline
   - Partial overlap → some shared elements. \newline
   - No contact → no shared elements. \newline
4. **Compare stated relationships to the diagram**: Check if the image visually represents the described group relationships.
}
\end{promptbox}

\begin{promptbox}{Solution to S1: Card Rotations Test}
{
1. **Analyze the target shape**: Note the structure, angles, and direction of each segment. \newline
2. **Examine the test shape**: Look closely at its geometry and orientation. \newline
3. **Mentally rotate the test shape**: Try 90°, 180°, and 270° rotations to see if it matches the target shape. \newline
4. **Avoid flipping**: Ensure no mirroring is involved—only rotation is allowed. \newline
5. **Compare for an exact match**: If any rotation results in a match with the target shape, it's the same.
}
\end{promptbox}

\begin{promptbox}{Solution to S2: Cube Comparisons Test}
{
1. **Observe the visible faces** on both cubes: Note which letters, numbers, or symbols are shown. \newline
2. **Apply the rule**: Each symbol can appear only once per cube. \newline
3. **Check the statement**: Determine if it claims, for example, that both cubes can show the same symbol on different faces. \newline
4. **Assess the possibility**: If a symbol appears on both cubes, that’s allowed. But if the statement implies a repeat **on the same cube**, and that symbol is already visible, then it’s **FALSE**. \newline
5. **Use logic for hidden faces**: Hidden symbols must be unique on the same cube, so validate the claim accordingly.
}
\end{promptbox}

\begin{promptbox}{Solution to SS2: Choosing A Path}
{
1. **Identify box [X]**: Locate the specific box labeled with the given identifier (e.g., A, B, etc.). \newline
2. **Trace the path from S**: Begin at the "S" (Start) within that box and follow valid line connections. \newline
3. **Follow only allowed connections**: Switch lines only at black dots; ignore line crossings without dots. \newline
4. **Reach the single circle**: Confirm there’s a path to the circle at the top within the same box. \newline
5. **Return to F**: Ensure a path continues from the circle back to "F" (Finish) without dead ends or exiting the box. \newline
6. **Check for continuity**: Confirm the entire path is unbroken, valid, and obeys all rules.
}
\end{promptbox}

\begin{promptbox}{Solution to SS3: Map Planning Test}
{
1. **Identify the starting and ending points**: Determine the origin location mentioned (e.g., an address or landmark) from the input. \newline
2. **Trace valid street paths**: Follow black lines only, avoiding any paths blocked by circles (road-blocks). \newline
3. **Evaluate shortest routes**: Try different paths from start to end, counting steps while obeying street and block rules. \newline
4. **Check building contact**: Ensure each route touches the side (not just corner) of only **one** numbered building. \newline
5. **Choose the shortest valid path**: Among all such routes, pick the one with the fewest steps and identify the building it touches. \newline
6. **Output the building number**: Return only the number of the building that the shortest valid path touches.
}
\end{promptbox}

\begin{promptbox}{Solution to VZ1: Form Board Test}
{
1. **Examine the target figure and puzzle pieces**: Analyze the target shape's structure and the five given pieces, noting their shapes and how they might fit together. \newline
2. **Identify usable pieces**: Consider each piece's potential rotations (no flipping), and determine which combinations (using 2–5 pieces) can fill the target figure exactly. \newline
3. **Test combinations logically**: Mentally or visually try valid combinations of pieces that can recreate the target shape without gaps or overlaps. \newline
4. **Check for inclusion of the highlighted piece**: For each valid solution, confirm whether the highlighted (`<ADDITIONAL\_0>`) piece is included. \newline
5. **Decide TRUE or FALSE**: If the highlighted piece appears in at least one valid combination that forms the figure, answer TRUE; otherwise, answer FALSE.
}
\end{promptbox}

\begin{promptbox}{Solution to VZ2: Paper Folding Test}
{
1. **Analyze the folding sequence**: Follow each fold step-by-step in the order shown, ensuring the paper orientation is preserved. \newline
2. **Track fold layers**: At each fold, visualize how layers stack on top of each other. \newline
3. **Simulate hole punches**: When a hole is punched through all layers, determine where it would pass through based on the stacked layers. \newline
4. **Unfold in reverse order**: Carefully reverse each fold, mentally tracking where the hole marks should appear on the unfolded paper. \newline
5. **Compare with the second image**: Check that the number, position, and symmetry of all holes match exactly with your mental unfolding.
}
\end{promptbox}

\begin{promptbox}{Solution to VZ3: Surface Development Test}
{
1. **Identify face X** on the flat paper: This face must end up on the outside of the folded 3D object. \newline
2. **Visualize folding**: Mentally fold the paper along the dashed lines to form the 3D shape shown in the second image, ensuring X is on the correct face. \newline
3. **Track edge positions**: Focus on the specified edge (e.g., "edge A") and determine how it moves and which edge it connects to after folding. \newline
4. **Compare with the 3D object**: Match the final position of the folded edge to the labeled edges on the 3D object. \newline
5. **Identify the corresponding edge**: Find the lettered edge that directly touches the given flat paper edge after all folds are complete.
}
\end{promptbox}

\clearpage

\section{Copyright issues}
\label{sec:copyright}

According to the licensing information provided by ETS, the PDF version of the Kit of Factor-Referenced Cognitive Tests is made available to researchers for academic use (BY-NC).
The licensing agreement specifies royalty requirements for researchers who reproduce physical copies of the specimen test booklets for test administration.
Our work does not involve administering the tests to human participants, printing, or reproducing test forms.
We only use AI to process the visual contents of the scanned items directly from the publicly accessible PDF.

\section{Broader impacts}
\label{sec:impacts}

\begin{enumerate}[leftmargin=*]
    \item \textbf{Shifting the Development Paradigm:} This research serves as a critical call to action for AI developers to move toward a bottom-up hierarchical training approach. It demonstrates that relying solely on large-scale pretraining for complex tasks can result in ``castles in the air''—models that appear intelligent but lack mastery of basic visual primitives like mental rotation, spatial relation inference, and figure-ground discrimination.
    \item \textbf{Scalable Infrastructure for Curriculum Learning}: Our parametric generator provides a framework for creating an unlimited supply of difficulty-controlled test cases. This is a vital resource for the community to implement curriculum-style training, allowing models to gradually master simple visual factors (\eg, edge detection, 2D-to-3D folding) before tackling complex semantic reasoning.
    \item \textbf{Calibrating User Expectations and Safety:} For AI users and stakeholders, {\methodname} provides a clear map of the boundaries of MLLM capabilities. Even frontier models like {\bestmodel} achieve only {\best} accuracy on these foundational tests, compared to 78.8 for humans. This highlights significant risks in deploying current MLLMs for safety-critical applications—such as medical imaging, autonomous robotics, or clinical decision support—where brittle spatial reasoning could lead to catastrophic hallucinations.
\end{enumerate}

\section{Declaration of AI usage}
\label{sec:llm-usage}

We use AI for revising draft, understanding relevant concepts, and writing the code of our framework.

\end{document}